%% file: main.tex
\documentclass[10pt,logo,copyright]{giga-report}
\input{packages}

\input{common.tex}

\definecolor{darkred}{rgb}{0.7, 0.0, 0.0}

\usepackage{makecell}
\usepackage{tabularx}
\usepackage{graphicx}
\usepackage{booktabs}
\usepackage{multirow}
\usepackage{svg}
\usepackage{listings}
\usepackage{wrapfig}
\usepackage{array}
\usepackage{enumitem}
\usepackage{needspace}
\usepackage[table]{xcolor}
\definecolor{mygray}{gray}{0.9}

\usepackage{amsmath}
\usepackage{pifont}

\usepackage[nameinlink]{cleveref}
\crefname{equation}{Eq.}{Eqs.}
\crefname{figure}{Fig.}{Figs.}
\crefname{section}{Sec.}{Secs.}
\crefname{appendix}{App.}{Apps.}
\crefname{table}{Tab.}{Tabs.}
\crefname{algorithm}{Algo.}{Algos.}
\crefname{thm}{Thm.}{Thms.}
\Crefname{thm}{Thm.}{Thms.}
\crefname{prop}{Prop.}{Props.}

\newcommand{\crefnames}[3]{%
  \@for\next:=#1\do{%
    \expandafter\crefname\expandafter{\next}{#2}{#3}%
  }%
}

\title{GigaBrain-0.7: Scaling Embodied Foundation Models to Emergent Capabilities with a Three-System Architecture}

\author{
\vspace{-0.1in}
\centerline{GigaBrain Team}
\centerline{{Project: \href{https://gigaai.cc/blog/gigabrain07}{https://gigaai.cc/blog/gigabrain07}}}
\centerline{Code: \href{https://github.com/open-gigaai/giga-brain-0}
{https://github.com/open-gigaai/giga-brain-0}}
\textbf{GigaBrain Team (alphabetical order)}:
\normalfont
Angen Ye, Axiang Sun, Can Jin, Chenxi Cheng, Chong Shi, Dengke Shang, Dingqian Zhang, Guan Huang, Guangqiang Wang, Guangqing Ding, Guo Li, Hangcong Li, Hengyu Zhong, Hongtao Lu, Jianbo Qin, Jiming Mao, Jing Zhu, Jindi Lv, Jingzhi Cui, Junjie Xie, Junyi Bao, Kai Liu, Lei Yuan, Limin Long, Lv Feng, Mingming Yu, Peng Li, Pengfei Yi, Qi Li, Qianli Zhang, Qingfang Li, Qitang Hu, Rui Zhang, Shaoyan Sun, Shibo Sun, Shiying Duan, Tenghui Chen, Tianze Liu, Weijie Ke, Wenyao Xue, Xiaofeng Wang, Xiaoyu Tian, Xinyu Liu, Xinze Chen, Yang Wang, Yankai Wang, Yejun Zeng, Yifan Li, Yifei Nie, Yilong Li, Yilong Liu, Yongchao Feng, Yumeng Wang, Yun Ye, Zhichao Liu, Ziheng He, Zonghai Yang, Zheng Zhu
\vspace{-1em}
}

\usepackage{svg}

\usepackage{etoolbox}

\makeatletter
\patchcmd{\abscontent}
  {\parbox{\dimexpr\linewidth}{\absfont \theabstract}}
  {{\absfont \theabstract\par}}
  {}
  {\PackageError{GigaBrain}{Failed to patch abscontent}{}}
\makeatother

\begin{document}
\maketitle

\begin{figure*}[htbp]
\centering
\captionsetup{type=figure, justification=justified, singlelinecheck=false}
\includegraphics[width=0.99\textwidth]{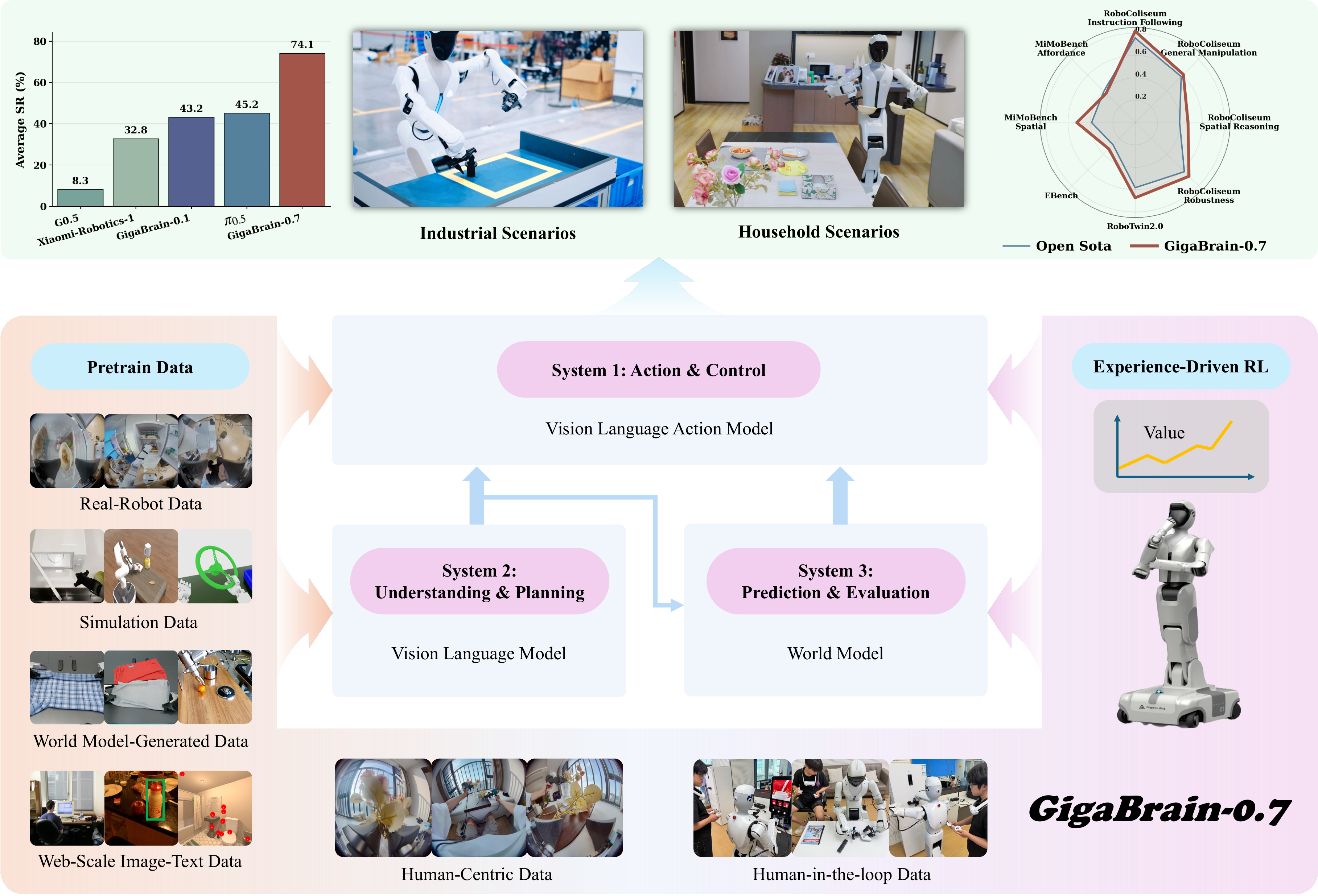}
\caption{\textbf{Overview of GigaBrain-0.7}. GigaBrain-0.7 is an embodied foundation-model system that coordinates understanding and planning, prediction and evaluation, and action and control through a three-system architecture. It learns from heterogeneous pretraining data and on-robot experience to support generalization and continual improvement across diverse robot embodiments and real-world scenarios.}
\label{fig:teaser}
\end{figure*}

\begin{abstract}
\input{sections/abstract}
\end{abstract}
\abscontent

\input{sections/introduction}

\input{sections/related_work}

\input{sections/data}

\input{sections/Model_Architecture}
\input{sections/model_pretrain}
\input{sections/experiment}
\input{sections/conclusion}
\input{sections/Appendix}

\clearpage

\setcitestyle{numbers}
\bibliographystyle{plainnat}
\bibliography{main}

\end{document}

%% file: packages.tex
\usepackage[numbers,sort&compress,square]{natbib}

\usepackage{iftex}
\ifPDFTeX
    \usepackage[utf8]{inputenc} %
    \usepackage[T1]{fontenc}    %
\fi

\usepackage{parskip}        %
\usepackage{url}            %
\usepackage{booktabs}       %
\usepackage{amsfonts}       %
\usepackage{nicefrac}       %
\usepackage{microtype}      %
\usepackage[dvipsnames]{xcolor} %
\usepackage{graphicx}
\usepackage{animate}        %
\usepackage{subcaption}
\usepackage{tabularx}
\usepackage{makecell}
\usepackage{adjustbox}
\usepackage{setspace}
\usepackage{todonotes}
\usepackage{colortbl}
\usepackage{wrapfig}
\usepackage{svg} 

\definecolor{rankbest}{RGB}{144,238,144}
\definecolor{rankmid}{RGB}{255,255,180}
\definecolor{rankworst}{RGB}{248,150,150}
\definecolor{deltadarkpos}{RGB}{50,180,50}
\definecolor{deltalightpos}{RGB}{180,235,180}
\definecolor{deltayellow}{RGB}{255,255,200}
\definecolor{deltalightneg}{RGB}{250,180,180}
\definecolor{deltadarkneg}{RGB}{220,80,80}

\newcolumntype{M}[1]{>{\centering\arraybackslash}m{#1}}
\usepackage{float}
\usepackage{tikz}
\usetikzlibrary{positioning,shapes,arrows,arrows.meta,fit,tikzmark}
\usepackage{amsmath,amsfonts,bm, bbm,leftindex}
\usepackage{multirow}
\usepackage{comment}
\usepackage{lipsum}
\usepackage[para]{threeparttable}

%% file: common.tex
\def\eqref#1{equation~\ref{#1}}

\def\1{\bm{1}}

\DeclareMathAlphabet{\mathsfit}{\encodingdefault}{\sfdefault}{m}{sl}
\SetMathAlphabet{\mathsfit}{bold}{\encodingdefault}{\sfdefault}{bx}{n}

\makeatletter
\let\save@mathaccent\mathaccent
\newcommand*\if@single[3]{%
  \setbox0\hbox{${\mathaccent"0362{#1}}^H$}%
  \setbox2\hbox{${\mathaccent"0362{\kern0pt#1}}^H$}%
  \ifdim\ht0=\ht2 #3\else #2\fi
  }
\newcommand*\rel@kern[1]{\kern#1\dimexpr\macc@kerna}
\newcommand*\widebar[1]{\@ifnextchar^{{\wide@bar{#1}{0}}}{\wide@bar{#1}{1}}}
\newcommand*\wide@bar[2]{\if@single{#1}{\wide@bar@{#1}{#2}{1}}{\wide@bar@{#1}{#2}{2}}}
\newcommand*\wide@bar@[3]{%
  \begingroup
  \def\mathaccent##1##2{%
    \let\mathaccent\save@mathaccent
    \if#32 \let\macc@nucleus\first@char \fi
    \setbox\z@\hbox{$\macc@style{\macc@nucleus}_{}$}%
    \setbox\tw@\hbox{$\macc@style{\macc@nucleus}{}_{}$}%
    \dimen@\wd\tw@
    \advance\dimen@-\wd\z@
    \divide\dimen@ 3
    \@tempdima\wd\tw@
    \advance\@tempdima-\scriptspace
    \divide\@tempdima 10
    \advance\dimen@-\@tempdima
    \ifdim\dimen@>\z@ \dimen@0pt\fi
    \rel@kern{0.6}\kern-\dimen@
    \if#31
      \overline{\rel@kern{-0.6}\kern\dimen@\macc@nucleus\rel@kern{0.4}\kern\dimen@}%
      \advance\dimen@0.4\dimexpr\macc@kerna
      \let\final@kern#2%
      \ifdim\dimen@<\z@ \let\final@kern1\fi
      \if\final@kern1 \kern-\dimen@\fi
    \else
      \overline{\rel@kern{-0.6}\kern\dimen@#1}%
    \fi
  }%
  \macc@depth\@ne
  \let\math@bgroup\@empty \let\math@egroup\macc@set@skewchar
  \mathsurround\z@ \frozen@everymath{\mathgroup\macc@group\relax}%
  \macc@set@skewchar\relax
  \let\mathaccentV\macc@nested@a
  \if#31
    \macc@nested@a\relax111{#1}%
  \else
    \def\gobble@till@marker##1\endmarker{}%
    \futurelet\first@char\gobble@till@marker#1\endmarker
    \ifcat\noexpand\first@char A\else
      \def\first@char{}%
    \fi
    \macc@nested@a\relax111{\first@char}%
  \fi
  \endgroup
}
\makeatother

%% file: sections/abstract.tex
Vision-language-action (VLA) models have become a dominant paradigm for generalist embodied agents, demonstrating strong complex and long-horizon task completion in structured settings. Yet it remains an open question whether current VLA systems can benefit from more effective architectural design, scale to substantially larger and more heterogeneous data regimes, and achieve broader generalization across tasks and embodiments. To this end, we present GigaBrain-0.7, an embodied foundation model with substantially improved generalization across diverse robot embodiments. Specifically, GigaBrain-0.7 unifies understanding, prediction, and action through a three-system architecture, scales pretraining to over 37,000 hours of heterogeneous embodied data, and introduces one-stage alignment training that jointly optimizes vision-language understanding and multi-embodiment action generation. Compared with the preceding GigaBrain-0 series and prior state-of-the-art models including $\pi_{0.5}$, GigaBrain-0.7 achieves substantial improvements in foundation zero-shot capabilities, language-conditioned instruction following, and post-training task success rates. In particular, on our in-house Maker H01 platform and mainstream robot embodiments, GigaBrain-0.7 demonstrates strong task adaptability and completion ability across both home and industrial scenarios. 
All training code and pretrained model weights will be released.

%% file: sections/introduction.tex
\section{Introduction}
 
Vision-Language-Action (VLA) models have emerged as a compelling paradigm for general-purpose robot control by adapting pretrained Vision-Language Models (VLMs) to translate visual observations and language instructions into executable actions~\citep{pi0,gigabrain0,openvla,rt2}.
As pretraining data and model capacity scale, recent VLAs have demonstrated increasingly strong out-of-the-box real-robot execution, downstream adaptation, and generalization across tasks, environments, and embodiments~\citep{walloss05,xiaomi_robotics_1,qwen_robotmanip,pi05,pi07,lingbot_vla2}.
However, embodied scaling presents challenges that extend well beyond collecting more trajectories or enlarging the policy network.
Robot data vary substantially across embodiments, action spaces, and execution conditions; without appropriate alignment and contextualization, greater data diversity may introduce interference rather than transferable knowledge~\citep{qwen_robotmanip,pi07}.
At the model level, most VLAs remain centered on reactive observation-to-action prediction, with limited mechanisms for anticipating future states or evaluating behavioral progress.
Recent systems have begun to address this through generated visual subgoals, world-model-conditioned policy learning, and predictive training objectives~\citep{pi07,gigabrain05m,lingbot_vla2,cosmos_policy}, yet understanding and planning, predictive judgment, action execution, and experience-driven improvement remain only partially coordinated across the full learning lifecycle.
Building capable embodied foundation models therefore requires more than scaling isolated policies---it calls for a unified learning system that integrates heterogeneous experience and multi-embodiment pretraining with understanding, prediction, action, and continual improvement from rollout and corrective feedback.

To this end, we present \textbf{GigaBrain-0.7}, an embodied foundation-model system designed to scale heterogeneous experience and coordinate understanding, prediction, action, and experience-driven improvement.
At its foundation is a \textbf{one-stage, multi-embodiment VLA pretraining framework} that jointly trains a pretrained VLM backbone, a continuous-action expert, and multimodal understanding objectives in a single pass, eliminating the optimization fragmentation typical of multi-stage training pipelines.
The model is organized around a three-system architecture.
\textbf{System~1} (Action and Control) generates continuous action chunks via flow matching~\citep{flowmatching} from multimodal observations, language instructions, proprioceptive states, and higher-level contextual signals, with an embodiment-aware action expert that accommodates different degrees of freedom and control modes across robot morphologies.
\textbf{System~2} (Understanding and Planning) interprets current and historical visual observations through temporal context modeling, tracks task progress, and decomposes long-horizon instructions into executable subtasks via hierarchical prompts.
\textbf{System~3} (Prediction and Evaluation) employs a world model to predict future visual states (subgoal images) and estimate state values, providing two complementary conditioning interfaces for System~1.
The three systems communicate through structured semantic, visual, and value-based interfaces while retaining specialized objectives.
On the data side, GigaBrain-0.7 is pretrained on over 37,000~hours of curated embodied trajectory data spanning 16 robot morphologies and approximately 270~million vision-language samples.
All trajectory data is standardized through a unified processing pipeline including format conversion to LeRobot~v3.0~\citep{cadene2026lerobot}, cross-embodiment state-action normalization, LLM-assisted language instruction rewriting and subtask annotation, and multi-stage quality control.
Building on Knowledge Insulation~\citep{knowledge_insulation}, we employ \textbf{Soft Knowledge Insulation (Soft KI)}, which attenuates rather than completely blocks action gradients entering the VLM backbone, allowing controlled adaptation to embodied control while preserving general vision-language capabilities.
Building on GigaBrain-0~\citep{gigabrain0}, which introduced world models as scalable data engines and embodied reasoning supervision, and GigaBrain-0.5M*~\citep{gigabrain05m}, which developed world-model-conditioned policy learning via RAMP, GigaBrain-0.7 extends the series toward a coordinated learning system spanning data curation, pretraining, planning, prediction, execution, and continual improvement.

In our experiments, we first conduct systematic scaling studies to evaluate the effects of robot-data scale, data-source composition, VLM backbone capacity, and VLA coupling architecture.
Results show a clear positive trend with increasing data scale: larger-scale pretraining consistently reduces validation loss and makes challenging real-robot behaviors increasingly reliable.
Together with the System~3 ablations and out-of-distribution evaluations, these results show signs of emergent embodied capabilities arising from large-scale heterogeneous pretraining.
We then assess the foundation capabilities of GigaBrain-0.7 on multimodal understanding benchmarks and on simulation environments spanning tabletop, mobile, and household manipulation.
Furthermore, we conduct extensive real-world evaluations on the AgileX PiPER/PiPER-X and our in-house Maker~H01 platforms, covering zero-shot and multi-task execution, language following, 
out-of-distribution generalization, and long-horizon manipulation.
We additionally perform controlled ablation studies to analyze the individual and joint contributions of System~3's future-state predictions and value-based evaluation signals.
Finally, we compare GigaBrain-0.7 against prior GigaBrain models~\citep{gigabrain0,gigabrain05m} and representative VLA baselines including $\pi_{0.5}$~\citep{pi05} after task-specific post-training, and evaluate experience-driven policy improvement using human-in-the-loop rollout data and corrective feedback.
GigaBrain-0.7 achieves substantial improvements over the preceding GigaBrain-0 series in foundation zero-shot capabilities, language-conditioned instruction following, and post-training task success rates.
All training code and pretrained model weights will be publicly released.

%% file: sections/related_work.tex
\section{Related Work}
\label{sec:related_work}

\subsection{Vision-Language-Action Architectures}
\label{sec:rw_vla_arch}

Recent progress in pretrained multimodal models has catalyzed the
development of vision-language-action (VLA) models for general-purpose
robot control~\citep{rt2, openvla, pi0, pi05, groot_n1, gigabrain0, qwen_robotmanip, xiaomi_robotics_0, xiaomi_robotics_1, walloss05, hyvla05, galaxea_g05, openvla_oft, lingbot_vla2}.
These models extend large-scale vision-language representations to
physical action, but differ substantially in how actions are represented
and how action generation is coupled to the pretrained VLM.
From this perspective, existing architectures can be broadly organized
into three families:
autoregressive VLM-as-Actor models, serial (cascaded)
VLM--action-expert models, and parallel Mixture-of-Transformers (MoT)
models.

\paragraph{Autoregressive action generation.}
The first family extends the VLM's native autoregressive interface to
robot control by representing continuous actions as discrete token
sequences.
RT-2~\citep{rt2} and OpenVLA~\citep{openvla} predict robot-action tokens
through next-token prediction, keeping the VLM itself as the action
generator.
Subsequent work improves the efficiency and expressiveness of action
tokenization.
FAST~\citep{fast} exploits temporal redundancy through frequency-domain
compression, while learned tokenization methods
~\citep{vqbet,actioncodec} learn compact discrete representations of
continuous action sequences.
Galaxea~G0.5~\citep{galaxea_g05} further develops this direction with a
learned cross-embodiment action tokenizer, allowing reasoning and action
tokens to be generated within a single autoregressive Transformer stream.
These approaches preserve a direct interface between language modeling and
physical action, but sequential token decoding becomes increasingly costly
as control frequency, prediction horizon, and action dimensionality grow.
OpenVLA-OFT~\citep{openvla_oft} consequently explores parallel decoding,
action chunking, and continuous action representations for more efficient
downstream adaptation of autoregressive VLA backbones.
\paragraph{Serial (cascaded) VLM--action-expert coupling.}

A second family separates multimodal reasoning from continuous action
generation.
Rather than requiring the VLM itself to decode low-level control, these
architectures use the pretrained VLM as a multimodal context encoder and
attach a separately parameterized action decoder or Action Expert.
Representative cascaded or decoupled designs include
GR00T~N1~\citep{groot_n1}, Gemini Robotics~\citep{gemini_robotics}, and
Qwen-RobotManip~\citep{qwen_robotmanip}.
In this paradigm, vision-language representations are first constructed
by the VLM and subsequently consumed by the motor generator through an
explicit feature interface.
Qwen-RobotManip, for example, employs a flow-matching DiT whose blocks
cross-attend to visual and language representations extracted from the
final VLM layer.
Such modular designs allow the action module to specialize in continuous
control, while their effectiveness depends strongly on the information
exposed through the VLM--action interface.

\paragraph{Parallel Mixture-of-Transformers.}

A third and increasingly common direction retains a dedicated continuous
Action Expert while coupling it more deeply with the vision-language
backbone through Mixture-of-Transformers architectures
~\citep{pi0, pi05,pi07, gigabrain0,xiaomi_robotics_0,xiaomi_robotics_1,walloss05,hyvla05}.
$\pi_0$~\citep{pi0} established the influential VLM--Action Expert
formulation with block-wise causal interaction and flow-matching action
generation.
The same architectural lineage is extended in
$\pi_{0.5}$~\citep{pi05} and $\pi_{0.7}$~\citep{pi07} with heterogeneous
pretraining, hierarchical task conditioning, observation history, and
richer context conditioning.
GigaBrain-0~\citep{gigabrain0} similarly combines a pretrained VLM with a
continuous Action Expert in an MoT formulation, together with RGB-D
perception, Embodied Chain-of-Thought supervision, and
Knowledge Insulation~\citep{knowledge_insulation}.
Wall-OSS-0.5~\citep{walloss05} introduces layer-wise VL and Action Experts
with joint attention and end-to-end gradient flow, while
HyVLA-0.5~\citep{hyvla05} combines an embodied-native MoT backbone with a
flow-matching Action Expert and compact temporal memory.
Xiaomi-Robotics-0~\citep{xiaomi_robotics_0} and
Xiaomi-Robotics-1~\citep{xiaomi_robotics_1} also adopt VLM--DiT MoT
architectures, conditioning flow-matching action generation on VLM
representations and KV caches.

Although these models differ in attention routing and gradient coupling,
they share a common design principle: a pretrained VLM is paired with a
dedicated continuous Action Expert, with different architectures exposing
different degrees of interaction between the two components.
Several recent systems additionally incorporate visual history into the
policy context.
HyVLA-0.5~\citep{hyvla05} introduces a compact memory encoder for
spatiotemporal context, while $\pi_{0.7}$~\citep{pi07} incorporates
multi-frame observation history into context-conditioned policy learning.
GigaBrain-0.7 follows the parallel VLM--Action Expert direction for
continuous action generation, while introducing separate functional
interfaces for understanding and planning, and for future-state prediction
and evaluation.

\subsection{VLA Training Paradigms}
\label{sec:rw_vla_training}

Beyond architecture, VLA systems differ substantially in how
vision-language understanding and robot action generation are introduced
and optimized over the model lifecycle.
Training recipes range from staged adaptation, where representation
learning and continuous action optimization are separated, to increasingly
joint formulations that optimize vision-language and action objectives
within the same VLA pretraining stage.

\paragraph{Multi-stage vs.\ one-stage pretraining.}

Most VLAs initialize from pretrained vision-language models and
subsequently introduce embodied and action supervision, but differ in how
strongly these objectives are separated during optimization.
Early autoregressive approaches such as RT-2~\citep{rt2} and
OpenVLA~\citep{openvla} adapt pretrained multimodal representations to
robot control through action-token prediction.
$\pi_{0.5}$~\citep{pi05} adopts a staged training recipe in which robot
actions are represented with discrete tokens during broad pretraining,
while a flow-matching Action Expert is introduced during post-training for
continuous action generation.
Xiaomi-Robotics-0~\citep{xiaomi_robotics_0} further decomposes VLA
pretraining into two optimization steps: it first trains the VLM jointly
on vision-language and robot-trajectory supervision and subsequently
freezes the VLM while training the flow-matching DiT.
Xiaomi-Robotics-1~\citep{xiaomi_robotics_1} similarly adopts an explicit
pretraining--post-training recipe, pretraining on over 100,000 hours of
real-world UMI trajectories with automatically generated state-transition
language and subsequently aligning the learned capabilities to robot
embodiments and imperative task instructions.

A complementary direction increasingly unifies vision-language and action
objectives within the same VLA pretraining stage.
GigaBrain-0~\citep{gigabrain0} jointly optimizes embodied reasoning,
discrete-action prediction, and continuous flow-matching action generation
under a unified objective.
Wall-OSS-0.5~\citep{walloss05} develops a single-stage
gradient-bridged co-training recipe that jointly optimizes multimodal
cross-entropy, action-token prediction, and continuous flow matching.
Qwen-RobotManip~\citep{qwen_robotmanip} emphasizes cross-source alignment
as a prerequisite for scaling heterogeneous manipulation data across
different representations, motions, and behaviors.
LingBot-VLA~2.0~\citep{lingbot_vla2} further scales pretraining to
approximately 60,000 hours of embodied data spanning diverse robot
configurations and egocentric human experience.
GigaBrain-0.7 follows this joint-training direction and scales it to
heterogeneous multi-embodiment data, jointly optimizing general
vision-language supervision, hierarchical task prediction, discrete action
representations, and continuous action generation within the same
\emph{VLA pretraining stage}.

\paragraph{Task-specific post-training.}

Large-scale pretraining establishes broadly transferable representations
and manipulation priors, while post-training specializes these capabilities
to target embodiments, tasks, and deployment conditions.
$\pi_{0.5}$~\citep{pi05} introduces continuous flow-matching action
generation during post-training and specializes the pretrained model on
high-quality task-relevant robot data.
OpenVLA-OFT~\citep{openvla_oft} studies efficient downstream fine-tuning of
autoregressive VLA backbones through parallel decoding, action chunking,
and continuous action prediction.
Xiaomi-Robotics-1~\citep{xiaomi_robotics_1} uses cross-embodiment robot
data during post-training to align UMI-pretrained capabilities with robot
embodiments and imperative instructions.
HyVLA-0.5~\citep{hyvla05} similarly specializes its pretrained model to
target robot embodiments through supervised fine-tuning before subsequent
reinforcement-learning refinement.
Together, these approaches establish post-training as an important bridge
between general-purpose pretraining and high-precision downstream robot
behavior.

\paragraph{Reinforcement learning.}

Supervised post-training remains bounded by the state distribution and
behavioral quality represented in demonstration data, motivating the use
of policy-generated experience for further improvement.
Recent approaches therefore extend VLA learning with reinforcement
learning over simulated or real-robot rollouts.
VLA-RL~\citep{vla_rl} and SimpleVLA-RL~\citep{simplevla_rl} investigate
direct reinforcement-learning updates of pretrained VLAs from environment
feedback, while Z-1~\citep{z1} develops a GRPO-based post-training
framework for flow-based VLAs with prefix-based rollout construction and
success-aware reward design.
RL~Token~\citep{rltoken} instead exposes a compact representation from a
pretrained VLA as an interface for lightweight online actor--critic
optimization.

Other approaches make greater use of offline experience, advantage
information, or preference supervision.
AWR~\citep{awr} performs advantage-weighted regression over off-policy
experience, while $\pi_{0.6}^{*}$~\citep{recap} introduces RECAP
(\emph{RL with Experience and Corrections via Advantage-conditioned
Policies}), which learns from demonstrations, on-policy experience, and
expert corrections through advantage conditioning.
GigaBrain-0.5M*~\citep{gigabrain05m} introduces RAMP, in which
world-model-predicted future states and values condition policy
fine-tuning and are iteratively combined with human-in-the-loop rollout
experience.
HyVLA-0.5~\citep{hyvla05} introduces FlowPRO, a critic-free and
reward-free PRO-based offline RL method that learns from paired failure
and corrective trajectories.
Together, these approaches move VLA adaptation beyond static imitation
toward closed-loop learning from policy-generated experience and
corrective feedback.

\subsection{World Model-Powered VLA}
\label{sec:rw_wm}

World models provide a complementary route for improving embodied
foundation models by learning predictive structure from large-scale visual
experience.
Existing work connects world modeling with robot learning through several
distinct roles, ranging from scalable data generation to unified
world-action modeling and predictive or evaluative signals for policy
learning.

\paragraph{Data generation and augmentation.}

One line of work treats world models as scalable data engines that expand
the distribution available for policy training.
GigaBrain-0~\citep{gigabrain0} uses world models to synthesize diverse
embodied experience through video generation, Real2Real and Sim2Real
transfer, viewpoint variation, and human-to-robot transfer.
DreamGen~\citep{dreamgen} similarly uses video world models to generate
synthetic interaction trajectories and recovers corresponding robot actions
through latent-action or inverse-dynamics models.
These approaches primarily use predictive models to increase the diversity
and coverage of training experience beyond what can be collected
efficiently on physical robots.

\paragraph{Unified World-Action Models.}

A second direction integrates visual dynamics and action prediction within
a common generative model.
GR-2~\citep{gr2} combines future video generation with robot-action
prediction.
DreamZero~\citep{dreamzero} builds a World-Action Model on a pretrained
video diffusion backbone that jointly models future world evolution and
robot actions.
Unified World Models (UWM)~\citep{uwm} couples video and action diffusion
within a unified architecture that supports policy, forward-dynamics,
inverse-dynamics, and video-generation modes.
GigaWorld-Policy~\citep{gigaworld_policy} develops an
\emph{action-centered World-Action Model}, using future visual dynamics
to provide additional supervision for action learning while allowing
explicit future-video generation to be omitted during deployment.
Cosmos Policy~\citep{cosmos_policy} fine-tunes pretrained video models for
visuomotor control and supports robot-action prediction together with
future-state and value prediction for planning.
Cosmos~3~\citep{cosmos3} further extends world modeling toward an
omnimodal foundation model for physical AI.

Beyond unified World-Action Models,
World-VLA-Loop~\citep{world_vla_loop} studies closed-loop refinement
between a video world model and a VLA policy.
Policy-generated failure rollouts are used to improve the world model,
which subsequently provides an updated learned environment for further
VLA reinforcement learning.
This formulation allows the world model and policy to improve iteratively
rather than treating either component as fixed.

\paragraph{Policy conditioning and value estimation.}

A complementary line retains a VLA-oriented policy while using predictive
representations, visual futures, or task-progress estimates to improve
policy learning and execution.
JEPA-VLA~\citep{vla_jepa} integrates video-predictive visual embeddings
into existing VLA models to strengthen representations of temporal
dynamics and action-relevant visual information.
$\pi_{0.7}$~\citep{pi07} conditions policy behavior on generated visual
subgoals together with language, observation history, and episode-level
context.
GigaBrain-0.5M*~\citep{gigabrain05m} introduces RAMP, where a separately
pretrained world model predicts future states and task values that
condition VLA policy learning and support iterative refinement from
human-in-the-loop rollout experience.
Feat2Go~\citep{feat2go} derives continuous task-progress targets from a
pretrained visual world model, trains a value model to estimate this
progress, and uses the resulting values for reward shaping during policy
optimization.

Together, these works establish complementary roles for predictive models
in robot learning, including data generation, dynamics representation,
World-Action modeling, policy conditioning, value estimation, and
closed-loop policy improvement.
GigaBrain-0.7 builds on these directions through a dedicated
\textbf{System~3} for prediction and evaluation.
System~3 is pretrained separately for robot-centric future-state prediction
and task-progress estimation.
Its predicted subgoal images and value-derived progress signals are
subsequently provided as additional conditions during task-specific VLA
post-training.
At inference time, the policy is conditioned on positive progress to guide
continuous action generation, without explicit candidate-action scoring or
selection.
The resulting design coordinates predictive modeling, conditional action
generation, and subsequent experience-driven refinement while retaining
specialized objectives for understanding, prediction, and action.

%% file: sections/data.tex
\section{Data Curation}
\label{sec:data_curation}

To support instruction understanding, visual perception, and continuous control of VLA models in open-world environments, we construct a multi-source training corpus composed of embodied trajectory data and vision-language data.
Embodied trajectory data provide state--action sequences, interaction processes, and robot-control supervision, while vision-language data preserve the model's general visual-semantic capabilities and further strengthen spatial-relation reasoning, object localization, affordance prediction, and task understanding.
These two data families provide complementary supervision during training, mitigating the degradation of vision-language capabilities that can arise from training solely on action data.

\subsection{Dataset Composition}
\label{sec:dataset_composition}

\begin{figure}[htbp]
    \centering
    \captionsetup{
        type=figure,
        justification=justified,
        singlelinecheck=false
    }
    \includegraphics[width=0.99\linewidth]
    {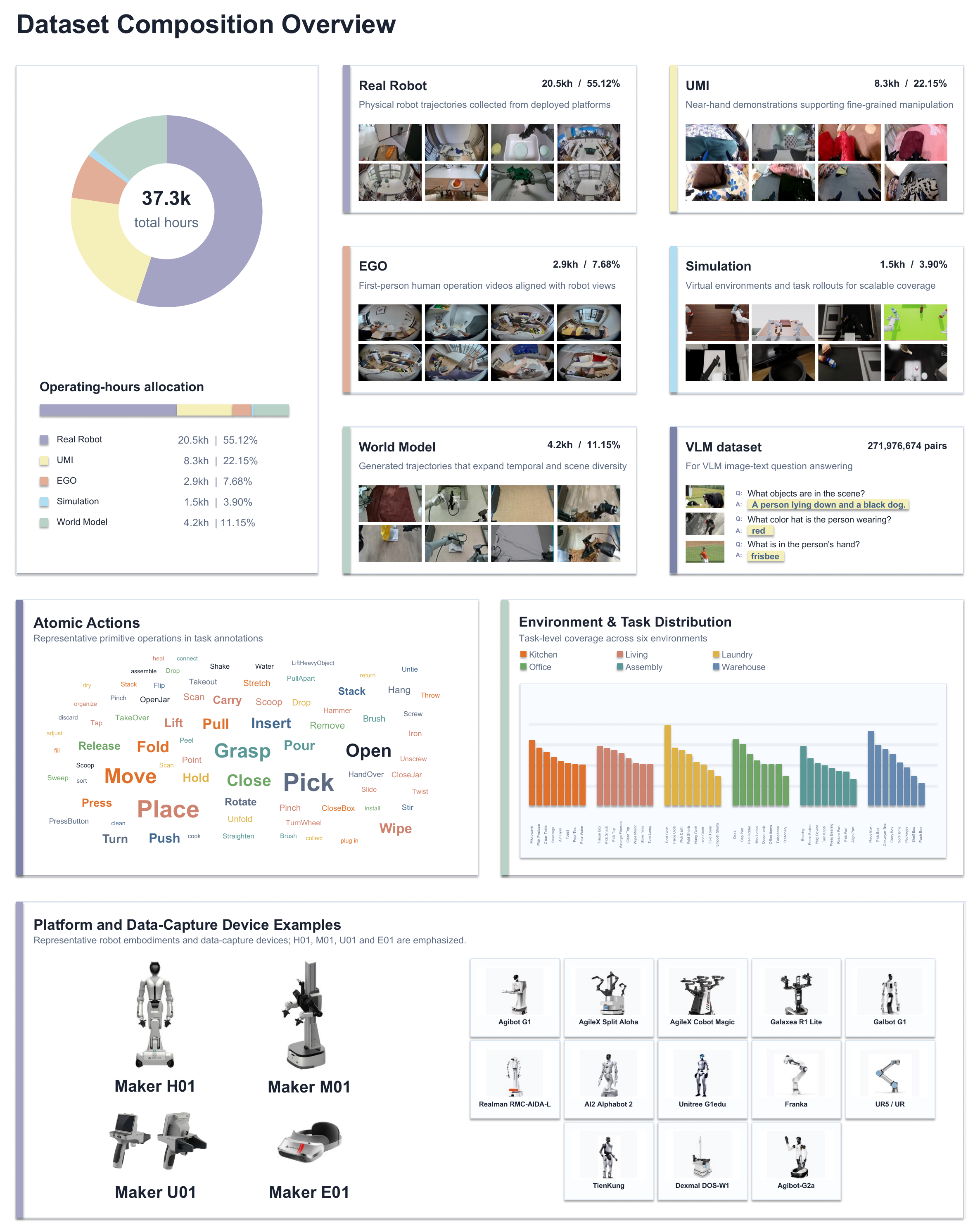}
    \caption{
    \textbf{Composition of the GigaBrain-0.7 training corpus.}
    The embodied trajectory corpus contains 37.3K hours of real-robot,
    UMI, EGO, simulation, and world-model-generated data, while the VLM
    corpus contains 271,976,674 image--text/question-answering samples.
    The figure further summarizes representative robot embodiments, 
    data-capture devices, atomic actions, and task/environment distributions.
    }
    \label{fig:data_composition}
\end{figure}

The training corpus consists of two major components: embodied trajectory data and VLM image--text/question-answering data.
After cleaning, the embodied trajectory corpus contains \textbf{37,256.98~hours} of data from real robots, Universal Manipulation Interface (UMI) demonstrations~\citep{umi}, egocentric (EGO) human demonstrations, simulation, and world-model-generated (WM) data.
These trajectories primarily provide state--action sequences, interaction processes, and robot-control supervision.
In parallel, we construct \textbf{271,976,674} VLM image--text/question-answering samples covering image captioning, general visual question answering, spatial-relation reasoning, region grounding, point prediction, affordance understanding, and robotic task understanding.
This multimodal corpus preserves broad vision-language capabilities while providing semantic and spatial supervision for embodied tasks.
Real-robot trajectories constitute the primary source of continuous-action supervision; UMI and EGO data supplement near-manipulation viewpoints and first-person human-interaction priors; and simulation and WM data provide complementary samples for controlled environments, long-horizon tasks, and specific target distributions.
Tab.~\ref{tab:trajectory_data} summarizes the embodied trajectory corpus, and Fig.~\ref{fig:data_composition} provides an overview of the complete training-data composition.

\begin{table}[t]
\centering
\caption{Composition of the embodied trajectory corpus after cleaning.}
\label{tab:trajectory_data}
\small
\begin{tabular}{lrr}
\toprule
\textbf{Data Type} & \textbf{Duration (h)} & \textbf{Proportion} \\
\midrule
Real Robot & 20,535.65 & 55.12\% \\
UMI        &  8,251.83 & 22.15\% \\
EGO        &  2,862.36 &  7.68\% \\
Simulation &  1,453.92 &  3.90\% \\
WM         &  4,153.22 & 11.15\% \\
\midrule
\textbf{Total} & \textbf{37,256.98} & \textbf{100.00\%} \\
\bottomrule
\end{tabular}
\end{table}

\paragraph{Real-robot data.}
The real-robot corpus is assembled from in-house robot deployments and large-scale real-world robot datasets, including~\citep{oxe,agibotworld2025colosseo,agibotworld2026,galaxea2025openworld,wu2025robocoin,hou2025robomind2}. After cleaning and standardization, the real-robot corpus spans 16 robot types, 1,810,101 episodes, 2,077,837,071 frames, and a total of 20,535.65 hours. Maker H01, Maker M01, Agibot-G1, AgileX, Ark, and Galaxea R1 Lite constitute the major data sources, while Franka, UR5, Realman RMC-AIDA-L, and other platforms further extend cross-embodiment and cross-hardware coverage. This distribution provides the basis for learning action patterns across different degrees of freedom, end-effectors, and locomotion modalities. Tab.~\ref{tab:real_robot_data} reports the detailed statistics by robot type.

\begin{table}[t]
\centering
\caption{Statistics of the real-robot data by robot type.}
\label{tab:real_robot_data}
\small
\setlength{\tabcolsep}{4pt}
\begin{tabular}{lrrr}
\toprule
\textbf{Robot Type} &
\textbf{Episodes} &
\textbf{Frames} &
\textbf{Frame Prop.} \\
\midrule
Agibot-G1          & 226,675 & 380,282,116 & 18.30\% \\
AgileX             & 133,101 & 179,727,199 &  8.65\% \\
Maker H01          & 866,864 & 910,496,160 & 43.82\% \\
Ark                &  30,559 &  35,960,476 &  1.73\% \\
Galaxea R1 Lite    &  27,793 &  35,058,414 &  1.69\% \\
Franka             &  87,148 &  25,149,233 &  1.21\% \\
UR5                &  57,210 &  25,007,639 &  1.20\% \\
Maker M01          & 319,964 & 418,534,560 & 20.14\% \\
Realman RMC-AIDA-L &  19,419 &  17,087,665 &  0.82\% \\
Dexmal DOS-W1      &  12,043 &  16,122,952 &  0.78\% \\
Agibot-G2          &   6,092 &  15,641,535 &  0.75\% \\
TienKung           &  11,218 &   7,598,429 &  0.37\% \\
Galbot G1          &   5,452 &   5,175,924 &  0.25\% \\
Ark-mobile         &   4,295 &   4,318,879 &  0.21\% \\
Unitree G1edu      &   1,411 &     910,983 &  0.04\% \\
AI2 Alphabot 2     &     857 &     764,907 &  0.04\% \\
\bottomrule
\end{tabular}
\end{table}

\paragraph{UMI and EGO data.}
The UMI corpus is composed of in-house Maker~U01 demonstrations and the Jianzhi~10K dataset~\citep{jianzhi10k2026}.
After cleaning, it contains \textbf{8,251.83~hours}, including \textbf{6,876.25~hours} of in-house data and \textbf{1,375.58~hours} from Jianzhi~10K.
This data source provides large-scale human manipulation demonstrations from near-manipulation viewpoints that visually resemble local observations from wrist-mounted or end-effector-proximal robot cameras.
It therefore complements real-robot trajectories with supervision for hand--eye coordination, local spatial geometry, and object-contact relationships.

The EGO corpus contains \textbf{2,862.36~hours} after cleaning and combines in-house Maker~E01 data with EgoDex~\citep{egodex}, EgoVerse~\citep{egoverse}, and World in Your Hands (WiYH)~\citep{wiyh}.
Because first-person human videos share similar observation viewpoints with robot-mounted cameras, they provide broader object categories, manipulation scenes, and human-interaction priors, strengthening semantic understanding and temporal modeling of real manipulation processes.
Tab.~\ref{tab:umi_ego_data} summarizes the UMI and EGO data composition.

\begin{table}[t]
\centering
\caption{Composition of the UMI and EGO datasets after cleaning.}
\label{tab:umi_ego_data}
\small
\begin{tabular}{llr}
\toprule
\textbf{Data Type} & \textbf{Dataset} & \textbf{Duration (h)} \\
\midrule
UMI & In-house Maker U01 & 6,876.25 \\
UMI & Jianzhi 10K        & 1,375.58 \\
\midrule
\textbf{UMI Total} & -- & \textbf{8,251.83} \\
\midrule
EGO & EgoDex      &   724.84 \\
EGO & EgoVerse    &   835.77 \\
EGO & WiYH        &   165.38 \\
EGO & In-house Maker E01 & 1,136.37 \\
\midrule
\textbf{EGO Total} & -- & \textbf{2,862.36} \\
\bottomrule
\end{tabular}
\end{table}

\paragraph{Simulation and generated data.}
To complement physically collected trajectories, we further construct
\textbf{5,607.14~hours} of simulation and generated data through two
complementary pipelines.
The simulation subset contains \textbf{1453.92~hours} of trajectories
collected from configurable physics-based environments, including
high-quality trajectories curated from~\citep{ebench,robotwin2,gu2023maniskill2,matthews2024kinetix,han2025robocerebra},
while the world-model-generated subset contributes
\textbf{4,153.22~hours} of data derived from real interaction samples.
The two sources target complementary regimes: simulation provides
controllable coverage of task and scene configurations that are costly
or difficult to reproduce on physical robots, whereas generation expands
the visual and contextual diversity of existing real-world interaction
data.
Together, they increase coverage of long-tail interaction conditions
while reducing the need to repeatedly collect rare or safety-sensitive
scenarios on physical hardware.

\paragraph{VLM image--text/question-answering data.}
In addition to embodied trajectory data, we construct \textbf{271,976,674} VLM image--text/question-answering samples, including \textbf{15,234,327} self-collected samples, to preserve foundational vision-language capabilities and strengthen the spatial understanding, object localization, and affordance reasoning required by robotic tasks.
The corpus covers image captioning, general visual question answering, multi-image reasoning, region grounding, point prediction, spatial-relation reasoning, robotic task understanding, and affordance learning.

The captioning data include~\citep{capsfusion,coco,flickr30k,cc12m,laion400m,danqing}, and image--text samples converted from~\citep{webvid10m}, providing supervision for open-vocabulary recognition and image-level semantic description.
The general VQA data include~\citep{cambrian,pixmo,vqav2,nlvr2,openimages,localized_narratives,llava665k}.
Embodied-specific data include~\citep{robo2vlm,robopoint,robointer_vqa,ai2thor,roboafford,refspatial,umi_vqa}, strengthening spatial referring, target localization, manipulation-feasibility judgment, and trajectory-related reasoning in robotic scenes.

\subsection{Embodied Trajectory Data Processing Pipeline}
\label{sec:trajectory_pipeline}

Robot trajectories and human manipulation data from heterogeneous sources differ substantially in storage format, robot degrees of freedom, coordinate definitions, action frequency, camera naming, language-annotation granularity, and quality distribution.
To enable stable joint training across these sources, we develop a unified embodied-trajectory processing pipeline that converts raw data into standardized representations and applies multi-stage quality control before export.

\begin{figure}[htbp]
    \centering
    \captionsetup{
        type=figure,
        justification=justified,
        singlelinecheck=false
    }
    \includegraphics[width=0.99\linewidth]
    {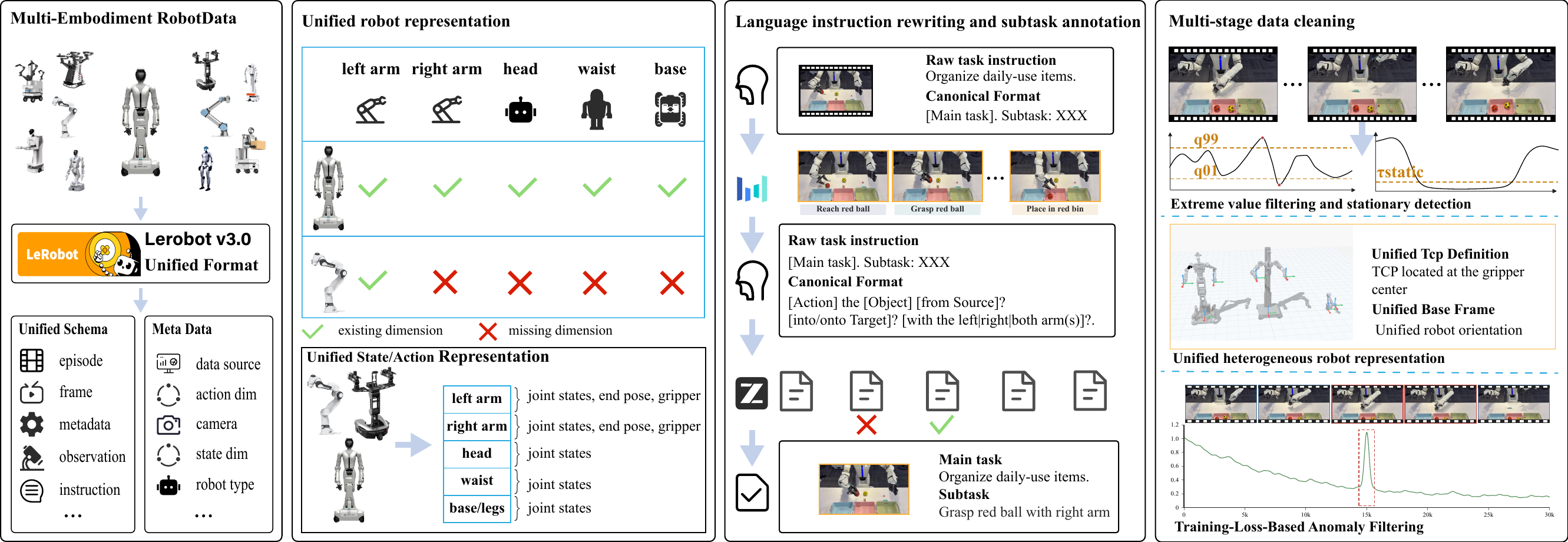}
    \caption{
    \textbf{Real-robot data processing pipeline.}
    Heterogeneous robot data are converted into the LeRobot~v3.0 format,
    mapped to a unified robot representation, standardized through
    language-instruction rewriting and subtask annotation, and filtered
    through multi-stage quality control before training.
    }
    \label{fig:real_robot_data_pipeline}
\end{figure}

\paragraph{Data format conversion and metadata organization.}
We first convert raw data into a unified training-storage format.
For LeRobot datasets, all samples are standardized to the LeRobot~v3.0 format, improving data-loading efficiency and unifying the organization of \texttt{episode}, \texttt{frame}, \texttt{observation}, \texttt{action}, and \texttt{language instruction} fields.
During conversion, we also organize dataset provenance, robot type, task description, camera list, sampling frequency, state dimensionality, action dimensionality, and cleaning flags, enabling downstream training to perform precise filtering and sampling by data source, robot morphology, and task type.

\paragraph{Unified robot representation.}
To handle mismatched degrees of freedom and sensor configurations across robot embodiments, we construct a unified robot state and action representation.
The dimension ordering follows left arm, right arm, head, waist, and base or legs, while both joint-space and end-effector-space fields are retained to support different supervision signals.
Under this representation, joint states, end-effector poses, gripper states, base velocities, and camera observations are organized into a reusable standard format.
Missing motion components are represented with masks or placeholder dimensions, avoiding embodiment-specific data formats for individual robot morphologies.
Because camera naming and mounting conventions also vary across platforms, preprocessing maps camera fields from each source to standardized camera definitions while retaining the original camera names for traceability.

\paragraph{Language instruction rewriting and subtask annotation.}
Raw task instructions often contain inconsistent naming, mixed languages, heterogeneous description granularity, or weak correspondence with the actual task.
For example, the same manipulator or gripper may be referred to as \texttt{left/right/both gripper} or \texttt{hand} across different data sources; some Chinese instructions must be aligned with English training corpora; and some samples contain only scene-level words without executable task descriptions.
To address these inconsistencies, we adopt an LLM-assisted instruction-standardization pipeline.
We first extract instructions using rule-based procedures, rewrite them into canonical task instructions with GLM-5.1~\citep{glm5team2026glm5}, and write the normalized instructions back to the dataset, followed by human spot checks.
To improve modeling of long-horizon task structure, we further construct subtask-level annotations.
For robot manipulation videos, a multimodal large model temporally segments the manipulation process and generates atomic subtask descriptions; samples with ambiguous task boundaries or unstable model annotations are further corrected through human annotation and review.

\paragraph{Multi-stage data cleaning.}
The cleaning stage filters or corrects numerical anomalies, prolonged stationary segments, kinematic inconsistencies, and training-anomalous samples.

First, we perform \emph{q01/q99 extreme-value filtering}.
Because quantile normalization is used during training, extreme outliers can substantially stretch the normalization interval and compress normal action signals into a narrow range, thereby destabilizing optimization.
We therefore compute the 1st and 99th percentiles independently for each robot type and action dimension and remove anomalous frames or trajectories outside the valid range.

Second, we remove prolonged stationary segments.
We compute inter-frame state differences, normalize each dimension by the corresponding joint range of motion, and then calculate the overall $L_2$ change magnitude.
When the magnitude falls below a predefined threshold, the corresponding frame pair is classified as stationary; consecutive stationary pairs are merged into stationary segments.
This procedure reduces the proportion of long, uninformative idle periods and focuses training on effective interaction behaviors.

Third, following Qwen-RobotManip~\citep{qwen_robotmanip}, we perform end-effector-pose consistency correction.
For each robot type in the training corpus, we collect the corresponding URDF files, align the base coordinate systems, and manually configure the end-effector reference point, thereby standardizing end-effector pose definitions across embodiments.

Finally, we use anomalous training losses as an additional filtering signal.
During early pretraining, we record sample-level and data-source-level loss distributions.
If particular trajectories persistently produce abnormally high losses or exhibit large loss spikes, we trace them back to the original video, action sequence, and language annotation to identify temporal misalignment, action discontinuities, incorrect instructions, or video corruption.
Confirmed anomalous samples are then added to the filtering list.

\paragraph{UMI data processing.}
For UMI demonstrations, we first synchronize multi-camera images, handle poses, gripper states, and host-side timestamps to ensure temporal consistency between visual observations and manipulation trajectories.
The acquisition system relies on synchronized multi-camera exposure and triggering, while handle-pose and gripper-opening signals are recorded through wired connections, providing a common temporal basis for subsequent cross-modal alignment, trajectory retargeting, and executability validation.

\begin{figure}[htbp]
    \centering
    \captionsetup{
        type=figure,
        justification=justified,
        singlelinecheck=false
    }
    \includegraphics[width=0.99\linewidth]
    {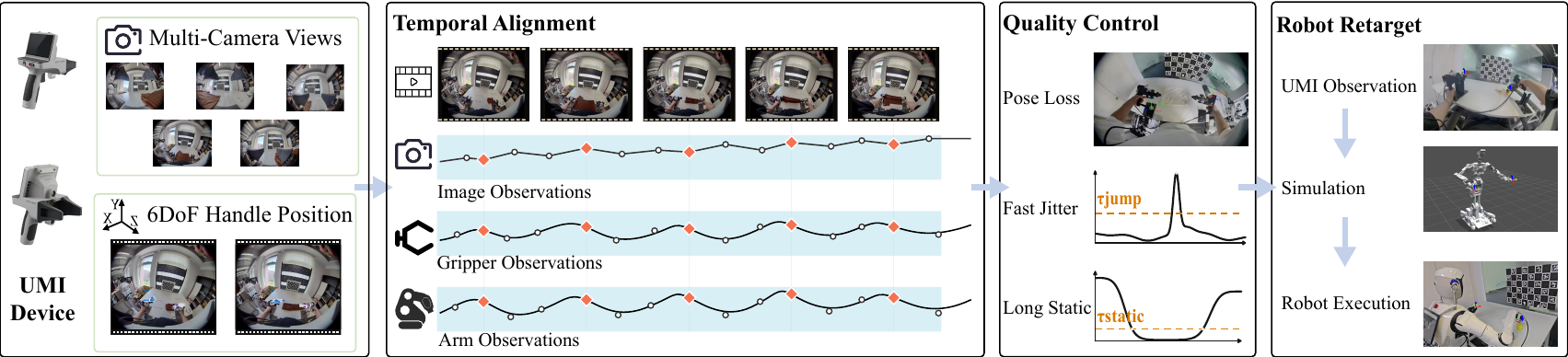}
    \caption{
    \textbf{UMI data processing pipeline.}
    Multi-camera observations, 6-DoF handle poses, and gripper states are
    temporally aligned, filtered for trajectory and modality quality, and
    retargeted to the Maker~H01 end-effector frame before simulation and
    sampled real-robot replay validation.
    }
    \label{fig:umi_data_pipeline}
\end{figure}

For quality control, we screen raw clips according to trajectory smoothness, information density, cross-modal consistency, and embodiment executability.
We remove samples with pose loss, rapid jitter, prolonged stationary periods, or clear inconsistencies between projected handle motion and visual observations.
The retained trajectories are then mapped to the Maker~H01 end-effector Tool Center Point (TCP) frame and validated through inverse-kinematics solving, simulation-based executability checks, and sampled real-robot replay.
After processing, UMI data are organized into unified training samples containing visual observations, end-effector actions, timestamps, and task-semantic information, and are jointly used with real-robot data for policy learning.

\paragraph{EGO data processing.}
EGO data supplement first-person examples of human manipulation, allowing the model to observe how human hands approach target objects, establish contact, and complete actions such as moving, placing, or opening objects.

\begin{figure}[t]
    \centering
    \captionsetup{
        type=figure,
        justification=justified,
        singlelinecheck=false
    }
    \includegraphics[width=0.99\linewidth]
    {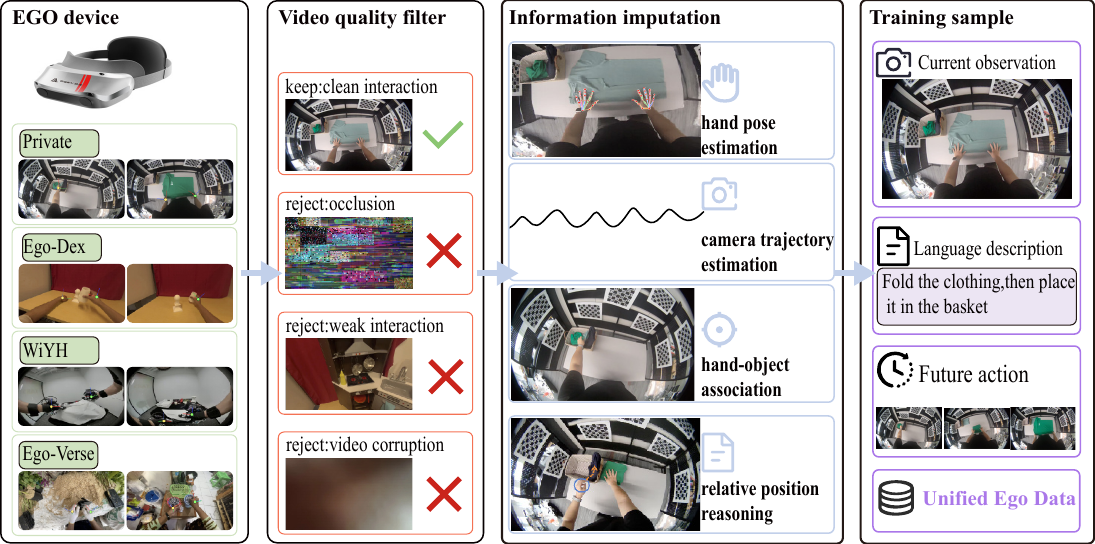}
    \caption{
    \textbf{EGO data processing pipeline.}
    First-person manipulation videos are filtered for interaction and video
    quality, supplemented with hand-pose and hand--object information when
    necessary, and standardized across EgoDex, EgoVerse, WiYH, and in-house
    data sources.
    }
    \label{fig:ego_data_pipeline}
\end{figure}

We first apply video-level quality filtering and remove clips without clear hand--object interaction, with corrupted frames, prolonged stationary periods, severe occlusion, or viewpoints unsuitable for interpreting the manipulation process.
This ensures that retained clips clearly expose the hand, target object, and resulting manipulation outcome.

We then standardize video frames, camera parameters, hand trajectories, and existing action annotations.
Timestamps, coordinate systems, and field formats are normalized across data sources so that video observations, hand motion, and action labels can be chronologically organized into training samples consistent with robot trajectories.
For data with hand keypoints or hand trajectories, we organize the temporal evolution of hand positions and associate them with contacted or manipulated target objects.
For samples without hand- or object-position annotations, we use camera-trajectory estimation, hand-pose estimation, or existing annotations to supplement interaction locations, target objects, and their relative spatial relationships.
After processing, EGO samples enter training as current observations paired with language descriptions and future actions or hand-motion trajectories, strengthening the model's understanding of first-person manipulation processes, hand--object contact relationships, and near-field spatial interactions.

\subsection{Simulation and Generated Data}
\label{sec:simulation_generated_data}

Real-robot collection provides physically grounded interaction data, but
scaling coverage to rare, long-tail, or safety-sensitive conditions
through physical collection alone remains costly.
We therefore complement real-world data with two scalable production
pipelines: world-model-based video generalization and physics-based
simulation.
Together, these pipelines contribute \textbf{5,607.14~hours} to the
embodied pretraining corpus.

\paragraph{World-model-based data generation.}
The generation pipeline starts from curated real-world interaction
samples and expands their scene and visual conditions through
prompt-conditioned video generation.
Generation tasks are constructed from the source interaction together
with configurable scene and generation conditions, producing additional
samples that preserve the underlying interaction context while broadening
the visual and environmental coverage of the training distribution.
After generation, the resulting samples are organized under the same data
management and quality-control framework as the remaining embodied corpus.

\paragraph{Physics-based simulation.}
In parallel, we construct simulation tasks from configurable robot, object,
and scene assets.
Each task is instantiated in a physics-based environment and executed to
produce temporally consistent simulated interaction data.
Compared with physical collection, this pipeline provides a controllable
and repeatable source of interaction experience for configurations that
are difficult to reproduce frequently or safely on real hardware.

\paragraph{Complementary long-tail coverage.}
The two pipelines serve complementary purposes.
World-model-based generation expands the diversity of real interaction
samples while retaining their visual and behavioral context, whereas
physics-based simulation provides explicit control over task and scene
configurations.
Their combination broadens the coverage of long-tail interaction
conditions and provides additional experience beyond what can be collected
efficiently from physical robots alone.

\subsection{VLM Data Processing and Annotation}
\label{sec:vlm_data_pipeline}

\paragraph{VLM data processing pipeline.}
The VLM corpus combines public image--text/question-answering datasets, converted video--text resources, embodied-specific VLM datasets, and self-collected embodied annotations derived from our trajectory data. We process all sources into a unified, training-ready multimodal dialogue format. The pipeline consists of annotation-format unification, image quality control, and sample-structure validation, ensuring consistency in text fields, image files, visual placeholders, dialogue-role ordering, and task metadata before training.

During annotation-format unification, image descriptions, single-turn question answering, multi-turn question answering, and embodied-specific supervision are mapped into a common dialogue representation according to task type, with appropriate image placeholders retained for samples that require visual input.
This step reduces formatting discrepancies across data sources and enables direct integration with the downstream training framework.
During image quality control, we check image readability, resolution, aspect ratio, color mode, and content validity, standardizing repairable samples and removing samples that are missing, corrupted, or insufficiently informative.
Before final export, we further validate field completeness, image--text path consistency, dialogue-role ordering, and non-empty text constraints, producing structurally consistent and quality-controlled VLM training data.

\paragraph{Embodied VLM annotation pipeline.}
In addition to aggregating public VLM datasets, we generate self-collected embodied VLM annotations from the cleaned real-robot, UMI, and EGO data. We extract keyframes, short video clips, historical observation windows, and the corresponding task instructions or trajectory context, and organize them into annotation units that can be processed by multimodal models. We then use Qwen3.6-27B~\citep{qwen3.6-27b} for large-scale batch annotation.
Following the annotation strategy in~\citep{xiaomi_robotics_0}, the model extracts intermediate structured states including object instances, interactable parts, spatial relations, viewpoint information, task stages, and action intent, from which VLM supervision samples are subsequently derived.

\paragraph{Affordance prediction.}
For affordance-prediction tasks, the annotation pipeline focuses on identifying the interactable properties of objects and object parts, including graspable regions, pressable or pullable components, placeable regions, and manipulable objects relevant to the task objective.
The multimodal model generates candidate interaction regions and corresponding language descriptions from object states in images or videos.
These annotations are further transformed into supervision for target localization, component identification, affordance judgment, and placement-region selection, enabling the model to learn correspondences among object function, manipulation constraints, and locally reachable space.

\paragraph{Spatial understanding.}
For spatial-understanding tasks, the annotation pipeline constructs spatial relationships among objects and between objects and the observer.
For data with depth, multi-view observations, or camera parameters, geometric information is preferentially used to generate supervision for relative position, proximity, occlusion, containment, and free space.
For RGB-only images or first-person videos, candidate spatial relationships are generated by combining visual cues with model reasoning and are subsequently filtered through rule-based consistency checks.
This data is primarily used to construct spatial-relation judgment, referring-expression localization, region-grounding, and spatial-feasibility question-answering samples.

\paragraph{High-level task planning.}
For high-level task-planning tasks, annotations are built from real-robot trajectories, UMI near-manipulation demonstrations, and EGO first-person human-manipulation clips, with emphasis on task stages, current subtasks, next-step actions, and task-completion states.
The multimodal model performs temporal understanding over continuous clips and generates current-state summaries, subtask labels, action pre/post relationships, and plausible next operations.
These structured annotations are further converted into next-step action prediction, task-progress judgment, action-consequence prediction, and goal-consistency question-answering samples, providing supervision for long-horizon task understanding and high-level planning.

\paragraph{Quality control.}
To reduce noise from automatic generation, we apply multi-level quality control after data production.
Samples are first automatically filtered for geometric consistency, target uniqueness, temporal consistency, and answer-format compliance.
Low-confidence samples, non-unique answers, and semantically ambiguous cases are further selected for human spot-check review.
Retained samples are then converted into a unified multimodal dialogue format, with data provenance, task type, generation rules, and quality status recorded for downstream sampling, error tracing, and dataset version management.

%% file: sections/Model_Architecture.tex
\section{Model Architecture}
\label{sec:model_architecture}

\begin{figure*}[t]
    \centering
    \captionsetup{
        type=figure,
        justification=justified,
        singlelinecheck=false
    }
    \includegraphics[width=\textwidth]
    {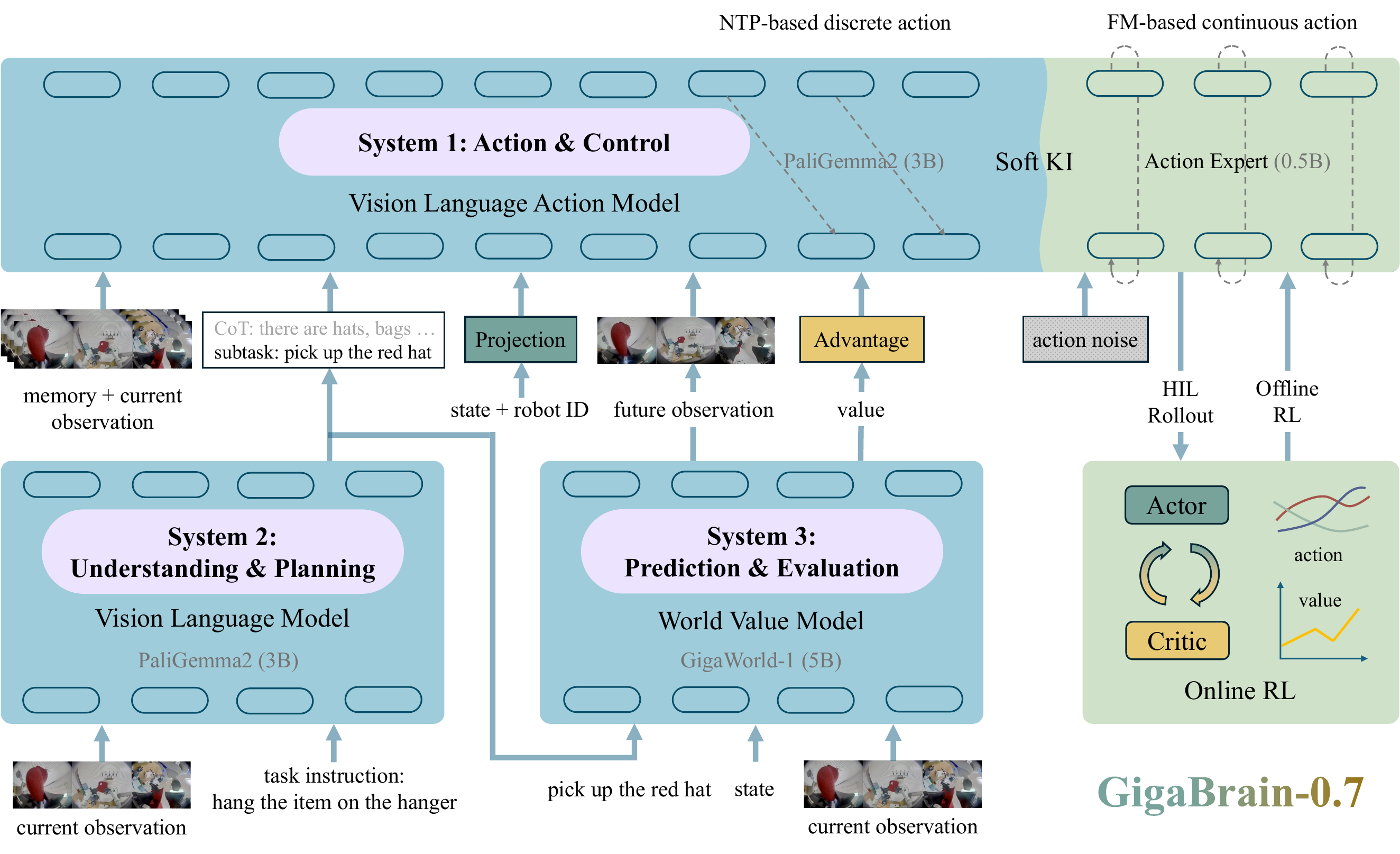}
    \caption{
    \textbf{Architecture of GigaBrain-0.7.}
    GigaBrain-0.7 organizes embodied intelligence into three interacting
    systems. System~2 performs visual-language understanding and task
    planning. System~3 introduces explicit prediction and evaluation by
    predicting future observations and estimating task-related values.
    System~1 integrates the current observation and short-term memory,
    subtask instructions, robot state, and predictive signals to generate
    executable actions through parallel discrete and continuous action
    pathways. The architecture further exposes interfaces for offline and
    online experience reinforcement, connecting perception, prediction,
    action, and subsequent policy improvement.
    }
    \label{fig:architecture}
\end{figure*}

Vision-language-action models provide a scalable interface between
visual-language understanding and continuous robot control.
A common design couples a pretrained vision-language model with a dedicated
action generator, allowing semantic representations to condition low-level
motor commands.
For long-horizon physical interaction, however, understanding the current
scene and predicting the next action are not sufficient on their own.
The policy must also maintain temporal context, anticipate how the scene may
evolve, and assess whether its current behavior is making useful progress.

GigaBrain-0.7 addresses these requirements through a three-system
architecture, as illustrated in Fig.~\ref{fig:architecture}.
\emph{System~2} is responsible for understanding and planning,
\emph{System~3} for prediction and evaluation, and
\emph{System~1} for action and control.
System~2 is a PaliGemma2 (3B)~\citep{paligemma2} vision-language model that interprets the
scene and decomposes tasks.
System~1 extends a PaliGemma2 (3B) backbone with a dedicated Action Expert
(0.5B) for continuous action generation.
System~3 is an independent GigaWorld-1~\citep{gigaworld1} based world value model (5B) that
provides System~1 with predictive signals about future task evolution.

We organize the architecture into three progressively coupled layers.
\emph{World simulation} provides semantic understanding, task planning,
future prediction, and evaluation.
\emph{Action alignment} maps these multimodal signals to continuous control
across heterogeneous robot embodiments.
\emph{Experience reinforcement} connects policy execution back to offline
and online policy improvement.

\subsection{World Simulation: Understanding, Prediction, and Evaluation}
\label{sec:world_simulation}

The world-simulation layer provides System~1 with context beyond the current
visual observation.
It combines System~2, which interprets the current scene and decomposes the
task, with System~3, which explicitly models future observations and
task-related values.
System~2 describes \emph{what should be done next}, while System~3 provides
prospective information about \emph{what may happen next} and
\emph{how the current execution state is evaluated}.

\paragraph{System~2: Understanding and Planning.}

System~2 receives the current visual observation together with the task
instruction.
It produces chain-of-thought reasoning that describes the current visual
context---identifying relevant objects and their spatial
arrangement---followed by a subtask instruction for the immediate
manipulation objective.
For example, given the instruction ``hang the item on the hanger,''
System~2 may observe ``there are hats, bags\ldots'' and decompose the task
into the subtask ``pick up the red hat''
(Fig.~\ref{fig:architecture}).

This decomposition provides an intermediate interface between high-level
task understanding and low-level action generation.
The resulting subtask information is passed to System~1 and also provides
task context for System~3, allowing prediction and control to operate on a
more explicit representation of the current task stage.

\paragraph{System~3: Prediction and Evaluation.}

System~3 is the predictive component of GigaBrain-0.7.
Built upon GigaWorld-1~\citep{gigaworld1}, it receives the current
embodied context---including the current observation, proprioceptive state,
and subtask description from System~2
(Fig.~\ref{fig:architecture}).
Unlike System~2, which reasons over observations that are already
available, System~3 explicitly models information about future task
evolution.
System~3 exposes two complementary signals to System~1.

\textbf{Future observation.}
Conditioned on the current embodied context and subtask instruction, the
world model generates a short video whose temporal extent is aligned with
the subtask horizon of System~1.
The last frame of this predicted sequence is extracted as the
\emph{subgoal image}~$g_t$---a compact visual representation of the
physical state expected after short-horizon task progress.
It is encoded and provided to System~1 as an additional visual condition,
allowing action generation to incorporate an explicit representation of the
anticipated scene evolution.

\textbf{Value-based evaluation.}
System~3 additionally estimates a scalar value~$V_t$ representing the
current task progress---how far the current execution state has advanced
toward subtask completion.
While sharing the same world-model backbone as the video pathway, the value
pathway attends primarily to visual observations.
The value estimate is converted into a binary advantage condition
$A_t \in \{0,\,1\}$, indicating whether the predicted task progress is
increasing or not over the current action segment.
As shown in Fig.~\ref{fig:architecture}, this value-derived condition is
injected into System~1's prompt, providing a compact evaluation of the
current execution state.
The derivation of~$A_t$ from~$V_t$ is described in
Section~\ref{sec:wm_vla_posttraining}.

Together, the subgoal image captures anticipated physical outcomes in
visual space, while the advantage provides a scalar evaluation signal for
action generation and subsequent experience-based learning.

\subsection{Action Alignment: Multi-Embodiment Action Generation}
\label{sec:action_alignment}

The action-alignment layer converts multimodal context into executable
robot control.
This role is performed by System~1, a vision-language-action model that
integrates semantic planning, predictive information, temporal context, and
embodiment-specific robot states within a shared action-generation framework.

\paragraph{System~1: Action and Control.}

System~1 receives the current observation and short-term visual memory,
the chain-of-thought output and subtask from System~2, proprioceptive state
and Robot~ID, and predictive signals from System~3
(Fig.~\ref{fig:architecture}).
These inputs are fused by the VLA to produce action chunks for closed-loop
robot execution.

System~1 combines its PaliGemma2 (3B) vision-language backbone with the
Action Expert (0.5B) in a Mixture-of-Transformers (MoT) design.
Let $\mathbf{H}^{\mathrm{VL}}_l$ and $\mathbf{H}^{\mathrm{AE}}_l$ denote
the hidden states of the vision-language and action-expert streams at
layer~$l$.
The two streams are coupled through joint attention but maintain separate
feed-forward parameters:
\begin{align}
    \mathbf{O}^{\mathrm{VL}}_l
    &=
    \mathrm{CausalAttn}_l
    \!\left(
    \mathbf{H}^{\mathrm{VL}}_l;\;
    \mathbf{H}^{\mathrm{VL}}_l
    \right),
    \label{eq:mot_vl_attn}
    \\[3pt]
    \mathbf{O}^{\mathrm{AE}}_l
    &=
    \mathrm{Attn}_l
    \!\left(
    \mathbf{H}^{\mathrm{AE}}_l;\;
    \bigl[\mathbf{H}^{\mathrm{VL}}_l,\,
    \mathbf{H}^{\mathrm{AE}}_l\bigr]
    \right),
    \label{eq:mot_ae_attn}
    \\[3pt]
    \mathbf{H}^{\mathrm{VL}}_{l+1}
    &=
    \mathrm{FFN}^{\mathrm{VL}}_l
    \!\left(\mathbf{O}^{\mathrm{VL}}_l\right),
    \quad
    \mathbf{H}^{\mathrm{AE}}_{l+1}
    =
    \mathrm{FFN}^{\mathrm{AE}}_l
    \!\left(\mathbf{O}^{\mathrm{AE}}_l\right),
    \label{eq:mot_ffn}
\end{align}
where $\mathrm{Attn}(\mathbf{Q};\,\mathbf{C})$ denotes attention with
queries from~$\mathbf{Q}$ and keys/values from context~$\mathbf{C}$,
$[\cdot,\cdot]$ denotes sequence concatenation, and residual connections
and layer normalizations are omitted for clarity.
The vision-language stream uses causal self-attention over its own tokens
(Eq.~\ref{eq:mot_vl_attn}), preserving the pretrained VLM's autoregressive
interface.
The action expert attends bidirectionally to the union of both streams
(Eq.~\ref{eq:mot_ae_attn}), enabling grounded action generation informed by
the full semantic context without injecting action-specific representations
into the VLM attention computation.

\paragraph{Short-term visual memory.}

Physical interaction is inherently temporal.
The same instantaneous observation may correspond to different task states
depending on preceding interactions, while temporary occlusions and repeated
visual configurations make single-frame perception insufficient for many
long-horizon tasks.

GigaBrain-0.7 therefore incorporates short-term visual memory through
\textbf{Temporal-Spatial Blocks} inserted into the visual encoder.
Given the current observation together with a short history of preceding
frames, each block factorizes visual aggregation into temporal and spatial
processing.
The temporal operation causally aggregates information across recent frames,
allowing the current representation to capture how task-relevant regions
evolve over time.
The subsequent spatial operation models interactions among visual regions
within the current observation, integrating the temporally enriched features
into a coherent scene representation.

Rather than forwarding all historical visual tokens to the VLA backbone,
the Temporal-Spatial Blocks progressively fuse historical information into
the current-frame representation.
After temporal-spatial processing, past-frame tokens are discarded and only
the enriched current-frame tokens are retained.
The number of visual tokens entering the VLA backbone therefore remains
comparable to the single-frame setting, providing temporal context without
naively increasing the high-level context length with the number of history
frames.


\paragraph{Discrete and continuous action pathways.}

System~1 supports two complementary action representations.
The NTP-based discrete pathway generates discrete action
tokens autoregressively through the vision-language head, providing a
symbolic action representation compatible with next-token prediction.
The FM-based continuous pathway generates action chunks through the Action
Expert via conditional flow matching, taking noise-corrupted action
representations as input and predicting continuous trajectories for precise
robot control.

Both pathways share the same visual-language context through the MoT joint
attention but serve different roles: the discrete pathway bridges language
understanding with motor control during training, while the continuous
pathway provides the executable actions used at deployment.

\paragraph{Soft Knowledge Insulation.}

A central difficulty in VLA learning is balancing pretrained vision-language
representations with robot-specific action learning.
The VLM must remain sensitive to language, objects, spatial relations, and
task semantics, while adapting sufficiently to support physical control.

GigaBrain-0.7 introduces \textbf{Soft Knowledge Insulation} (Soft KI)
between the vision-language and continuous-action pathways.
Rather than completely isolating semantic representations from action
learning~\citep{pi0}, Soft~KI provides a controlled interface that allows
the backbone to develop embodied-aware visual representations while
preserving its general-purpose vision-language capabilities.

\paragraph{Embodiment-aware state interface.}

Scaling a single VLA across heterogeneous robots requires alignment between
different state spaces and kinematic configurations.
Following the unified robot representation in
Section~\ref{sec:data_curation}, GigaBrain-0.7 provides System~1 with both
the proprioceptive state and a Robot~ID through a dedicated projection
module.
The proprioceptive state is element-wise masked by a per-embodiment
validity vector---distinguishing absent dimensions from zero-valued
states---and projected together with the Robot~ID into the shared VLA
representation.
This corresponds to the ``Projection'' module in
Fig.~\ref{fig:architecture}.

The main Action Expert parameters are shared across embodiments, while
per-embodiment input and output projections handle differences in native
action definitions.
This allows different robot platforms to share the core manipulation
knowledge while preserving embodiment-specific precision.
Rotations are represented using the continuous 6D format~\citep{6drot}
throughout the action space.

\subsection{Experience Reinforcement}
\label{sec:experience_rl}
\begin{figure*}[t]
    \centering
    \captionsetup{
        type=figure,
        justification=justified,
        singlelinecheck=false
    }
    \includegraphics[width=\textwidth]
    {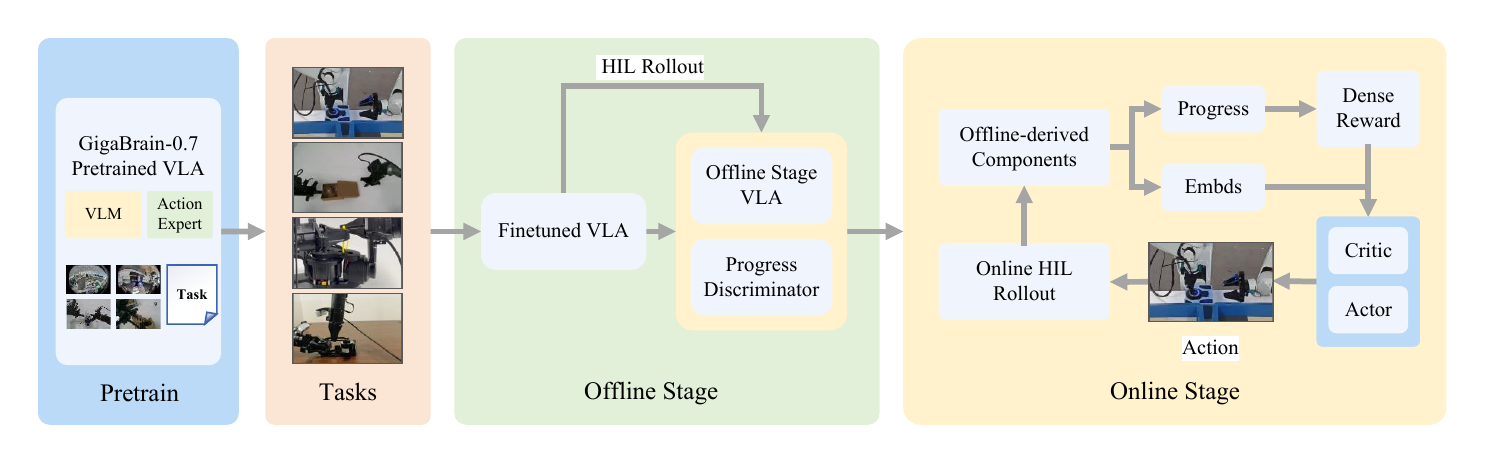}
    \caption{
    \textbf{Experience Reinforcement Pipeline.}
    The experience reinforcement pipeline consists of three stages: supervised fine-tuning, offline reinforcement, and online reinforcement. Supervised fine-tuning establishes basic competence on complex, fine-grained out-of-domain tasks. Offline reinforcement develops progress awareness and error recovery, while online reinforcement further improves fine-grained manipulation and task success.
    }
    \label{fig:rl_arch}
\end{figure*}
As illustrated on the right side of Fig.~\ref{fig:architecture}, GigaBrain-0.7 builds an experience-reinforcement interface around System 1 that connects the post-trained policy to both offline and online learning stages. First, human-in-the-loop (HIL) intervention data generated by System 1 are used for offline learning, yielding an error-aware progress discriminator and a policy with a certain degree of error-recovery capability. During online interaction, subtask success labels, together with the progress discriminator, are used to construct dense rewards, while the policy obtained from offline learning is connected to a lightweight actor--critic loop.
The same System 1 policy is retained throughout pretraining, post-training, and deployment, allowing supervised learning, offline experience learning, and online interaction to form a continuous model lifecycle without introducing a separate controller.

%% file: sections/model_pretrain.tex
\section{Model Training}
\label{sec:model_training}

Training proceeds in four stages.
We first pre-train Systems~1 and~2 jointly as a generalist
vision-language-action policy on the heterogeneous corpus described in
Section~\ref{sec:data_curation}.
We then extend GigaWorld-1 into a World Value Model through robot-centric
video pre-training followed by value learning, producing System~3.
During task-specific VLA post-training, System~3 is frozen and its
outputs condition System~1 on downstream demonstrations while Systems~1
and~2 are jointly optimized.
Finally, the post-trained policy is refined through offline and online
reinforcement learning on its own deployment experience.

\subsection{VLA Pre-Training}
\label{sec:vla_pretraining}

We jointly train Systems~1 and~2 on the full heterogeneous corpus---real
robot, UMI, EGO, simulation, world-model-generated trajectories, and
vision-language data---using a unified pre-training recipe that jointly
optimizes semantic understanding, hierarchical task prediction, and
continuous action generation.

\paragraph{Joint vision-language-action learning.}

GigaBrain-0.7 supports two complementary
prediction pathways (Section~\ref{sec:action_alignment}):
an autoregressive next-token prediction (NTP) pathway for language,
subtask, and discrete-action outputs, and a flow-matching (FM) pathway
for continuous action chunks.
Unlike staged training recipes that introduce continuous action generation
only in a later training phase, both pathways are optimized throughout
GigaBrain-0.7 pre-training.

For an autoregressive target sequence $y_{1:M}$ conditioned on multimodal
context $c$, the NTP objective is
\begin{equation}
\mathcal{L}_{\mathrm{NTP}}
=
-
\mathbb{E}
\left[
\sum_{j=1}^{M}
m_j
\log
p_{\theta}
\left(
y_j
\mid
c, y_{<j}
\right)
\right],
\label{eq:ntp_loss}
\end{equation}
where $m_j$ masks positions without autoregressive supervision.
Depending on the data source, the targets include subtask descriptions,
discrete action representations, and general vision-language outputs such
as captioning, VQA, spatial reasoning, grounding, and affordance
understanding.

For continuous action generation, let $a_{t:t+H}$ denote a ground-truth
action chunk and let $\epsilon \sim \mathcal{N}(0,I)$ denote Gaussian
noise of the same dimensionality.
Following the flow-matching formulation used in prior GigaBrain
models~\citep{gigabrain0}, we construct
\begin{equation}
a_{t:t+H}^{\tau}
=
\tau a_{t:t+H}
+
(1-\tau)\epsilon,
\qquad
\tau\in[0,1].
\label{eq:fm_path}
\end{equation}
The Action Expert predicts the corresponding flow field and is optimized
with
\begin{equation}
\mathcal{L}_{\mathrm{FM}}
=
\mathbb{E}_{a,\epsilon,\tau}
\left[
\left\|
v_\theta
\left(
a_{t:t+H}^{\tau}, c, \tau
\right)
-
\left(
a_{t:t+H}-\epsilon
\right)
\right\|_2^2
\right].
\label{eq:fm_loss}
\end{equation}

The two pathways are jointly optimized during pre-training:
\begin{equation}
\mathcal{L}_{\mathrm{VLA}}
=
\mathcal{L}_{\mathrm{NTP}}
+
\mathcal{L}_{\mathrm{FM}}.
\label{eq:vla_pretraining_objective}
\end{equation}
To reduce interference between continuous-action learning and the
pretrained vision-language representation, we apply Soft Knowledge
Insulation~\citep{knowledge_insulation}.
Specifically, FM gradients are propagated fully within the Action Expert
but attenuated by a coefficient $\alpha_{\mathrm{KI}}$ before entering the
vision-language backbone:
\begin{equation}
\nabla_{\theta_{\mathrm{VL}}}
\mathcal{L}_{\mathrm{VLA}}
=
\nabla_{\theta_{\mathrm{VL}}}
\mathcal{L}_{\mathrm{NTP}}
+
\alpha_{\mathrm{KI}}
\nabla_{\theta_{\mathrm{VL}}}
\mathcal{L}_{\mathrm{FM}},
\qquad
0 < \alpha_{\mathrm{KI}} < 1.
\label{eq:soft_ki}
\end{equation}
This preserves full autoregressive supervision of the vision-language
backbone while still allowing continuous-action learning to adapt its
representations toward embodied control.

\paragraph{Temporal context supervision.}

To train the Temporal-Spatial Blocks described in
Section~\ref{sec:action_alignment}, we provide historical observations
alongside the current frame.
Historical observations are randomly dropped during training, exposing
the model to varying amounts of temporal context---including
single-frame inputs---and reducing dependence on a fixed history length.

\paragraph{Hierarchical task supervision.}

Each trajectory carries both a task-level instruction $\ell$ and a
temporally aligned subtask instruction $\hat{\ell}_t$.
We mix two supervision modes.
In the first, both $\ell$ and $\hat{\ell}_t$ are provided and the model
directly predicts the corresponding action.
In the second, only the task-level instruction is provided, requiring
System~2 to autoregressively infer the current subtask before System~1
generates the corresponding action.
This couples high-level task reasoning and low-level control within the
same pre-training framework.
Causal attention masks prevent subtask targets and future actions from
leaking into the representations used to predict them.

\paragraph{Multi-embodiment unified optimization.}

The pre-training corpus spans embodiments with substantially different
morphologies, degrees of freedom, and control conventions.
Following the unified representation in
Section~\ref{sec:data_curation}, heterogeneous states and actions are
aligned to a common semantic structure while retaining
embodiment-specific validity information.

Each embodiment's proprioceptive state is projected into the shared model
space through the embodiment-aware interface described in
Section~\ref{sec:action_alignment}.
Validity masks distinguish unavailable dimensions from valid zero-valued
states, and invalid action dimensions are excluded from both
flow-matching supervision and noise injection.
The Action Expert shares its main Transformer parameters across
embodiments, while embodiment-specific input and output projections
accommodate differences in action dimensionality and control convention.

For egocentric data, future actions are re-expressed in the coordinate
frame of the current observation to provide a stable reference throughout
the action chunk.
Rotations use the continuous 6D representation defined in
Section~\ref{sec:action_alignment}.

\subsection{World Model Pre-Training}
\label{sec:world_model_pretraining}

System~3 provides two signals to System~1
(Section~\ref{sec:world_simulation}): a predicted future observation and a
task-progress value estimate.
We obtain both capabilities by extending GigaWorld-1~\citep{gigaworld1}
through two successive training stages.

\paragraph{Stage~I: Robot-centric video pre-training.}

Starting from GigaWorld-1, we continue video pre-training on robot
manipulation data drawn from the real-robot and simulation tiers of the
data pyramid (Section~\ref{sec:data_curation}).
The objective is to adapt general visual dynamics toward the state
transitions and contact interactions that characterize robotic manipulation.

Conditioned on the current embodied context and subtask instruction, the
model generates a future video whose temporal extent matches the
action-chunk horizon of System~1.
The final predicted frame serves as the subgoal image~$g_t$---a compact
visual representation of the physical state expected after short-horizon
task progress.

\paragraph{Stage~II: Value learning.}

Future-state prediction tells the policy what may happen, but not whether
the current state is favorable for task completion.
In a second stage, we equip the model with task-progress estimation.

We extend the video-pre-trained checkpoint into a Mixture-of-Transformers
World Value Model that jointly supports future prediction and value
estimation.
Similar to Viva~\citep{viva2026}, the value pathway is supervised to estimate task progress---a scalar
reflecting how far the current execution state has advanced toward subtask
completion, derived from trajectory-level completion annotations in the
training data.
The value pathway attends primarily to current and historical visual
observations when producing the progress estimate~$V_t$.

System~3 parameters are frozen after this stage.
During subsequent VLA post-training, System~3 generates fresh
outputs~$g_t$ and~$V_t$ for each training sample but receives no gradient
updates.

\subsection{World-Model-Powered VLA Post-Training}
\label{sec:wm_vla_posttraining}

We combine the pre-trained VLA with the frozen System~3 during task-specific
post-training.
System~3's outputs are provided directly to System~1 as additional input
conditions; only Systems~1 and~2 are optimized on downstream
demonstrations.
Freezing the World Value Model preserves its predictive representations
while allowing the VLA to learn how these signals should influence
task-specific action generation.

\paragraph{Subgoal-image conditioning.}

For each post-training sample, System~3 generates future video predictions spanning the same time horizon as the target subtask.
The subgoal image~$g_t$ is encoded through the visual pathway and appended
to System~1's multimodal context, providing the policy with an explicit
visual prediction of the near-future state alongside the current
observation.

\paragraph{Value-derived conditioning.}

The task-progress estimate from System~3 is converted into a binary
advantage condition.
For two temporally separated states within the action-chunk interval, we
compute the progress difference and discretize it:
\begin{equation}
A_t
=
\mathbb{1}\!\left[
V_{t+\delta_t} - V_t > 0
\right]
\;\in\; \{0,\,1\},
\label{eq:advantage}
\end{equation}
where $A_t = 1$ marks an action segment associated with increasing task
progress and $A_t = 0$ marks non-improving or regressing segments.
The advantage is embedded as a conditioning token in the VLA prompt,
allowing System~1 to distinguish productive from unproductive behaviors
in the training data.
At inference time, the policy is conditioned on positive progress,
biasing action generation toward behaviors associated with advancing task
state---analogous to the advantage-conditioned approach in
RAMP~\citep{gigabrain05m}, but with the advantage derived from a dedicated
world value model.

\paragraph{Condition dropout.}

We apply stochastic condition dropout to prevent over-dependence on
System~3 outputs.
The subgoal image and the value-derived condition are dropped
\emph{independently}, with probabilities~0.50 and~0.15 respectively.
The VLA is consequently exposed to four conditioning regimes: neither
signal, subgoal only, value only, and both.

\paragraph{Post-training optimization.}

Task-specific post-training optimizes the policy through continuous
action generation under the augmented context
\begin{equation}
c_t^{\mathrm{post}}
=
\left(
c_t^{\mathrm{base}},
g_t,
A_t
\right),
\label{eq:post_context}
\end{equation}
where $c_t^{\mathrm{base}}$ contains the current observation, temporal
context, language instruction, proprioceptive state, and embodiment
information; $g_t$ denotes the predicted subgoal image; and $A_t$
denotes the value-derived progress condition.

We optimize the same flow-matching objective defined in
Eq.~\ref{eq:fm_loss}, using $c_t^{\mathrm{post}}$ in place of the
pre-training context $c$.
System~3 remains frozen throughout this stage, while Systems~1 and~2
are optimized on task-specific demonstrations.

\subsection{Experience-Driven Reinforcement Learning}
\label{sec:rl_training}

Demonstration learning is bounded by the states present in the offline
dataset.
Deploying the policy on real robots reveals failure states, recovery
opportunities, and corrective behaviors that arise from the policy's own
execution distribution.
Starting online RL directly at this point is sample-inefficient---the
policy may repeatedly visit the same failure states---so we adopt a staged
offline-to-online pipeline.

\paragraph{Offline experience reinforcement.}

We deploy the world-model-powered post-trained VLA to collect rollout
trajectories and organize them by execution outcome.
The resulting dataset contains both successful and failed behavior,
providing direct evidence about which actions lead to desirable and
undesirable outcomes.

We apply an Advantage-Weighted Regression (AWR)~\citep{awr} style objective
for offline policy refinement.
In this part, reward $r_t$ is calculated from the progress discriminator and predefined rules, advantage $\tilde{A}_t$ is defined as 
\begin{equation}
\tilde{A}_t=G_t-\tilde{V}_t
\label{eq:online_rl_advantage}
\end{equation}
Where $G_t$ and $\tilde{V}_t$ denote the cumulative return and value estimate, respectively.

Rollout segments are weighted according to their estimated advantage:
successful and higher-progress segments receive greater training weight,
while failed or lower-progress behaviors are down-weighted.
This allows the policy to learn preferentially from better experience
without requiring additional environment interaction during optimization.

The offline stage converts already-collected experience into improved
behavior and provides a stronger initialization for online learning,
substantially reducing the costly exploration required in the next stage.

\paragraph{Online reinforcement with human correction.}

After offline refinement, we redeploy the policy and collect on-policy
trajectories.
When the robot reaches a difficult state, a human operator intervenes and
provides corrective actions.
The corrective demonstration is recorded as a labeled action segment at
the precise failure state and enters the training stream, providing direct
supervision on the hardest states of the current policy distribution.

We use an actor--critic framework for online optimization, retaining
System~1 as the deployment actor and training a separate critic for
value estimation.
Rewards combine chunk-level progress differences with terminal sparse
signals.
Training alternates between rollout collection and model updates:
\begin{equation}
\pi^{(k)}
\;\longrightarrow\;
\mathcal{D}_{\mathrm{rollout}}^{(k)}
\;\longrightarrow\;
\operatorname{Update}
\!\left(
\pi^{(k)},\,
\mathcal{D}_{\mathrm{rollout}}^{(k)}
\right)
\;\longrightarrow\;
\pi^{(k+1)}.
\label{eq:online_rl_loop}
\end{equation}

Each improved policy generates the experience for subsequent iterations,
progressively shifting training toward the state distribution encountered
during deployment.

\paragraph{Continual improvement.}

The four training stages form a progressive lifecycle:
pre-training establishes broad embodied priors;
world-model pre-training adds future prediction and progress estimation;
post-training specializes these signals for target tasks;
and experience-driven reinforcement improves the policy on states produced
by its own execution.
The online stage can be repeated as new rollout data becomes available,
forming a continual \emph{rollout--correction--update} cycle.
A detailed treatment of the human-intervention protocol and the full online
RL formulation will be provided in a dedicated future GigaBrain report.

%% file: sections/experiment.tex
\section{Experiments}
\label{sec:experiments}

We evaluate GigaBrain-0.7 through a series of experiments designed to
answer the following questions:
(1)~Which backbone and VLM--action expert architecture provide a strong
basis for large-scale embodied pretraining?
(2)~How do data scale and data composition affect pretrained robot
capabilities?
(3)~How does temporal context help resolve ambiguous repeated states during manipulation?
(4)~What are the individual and joint contributions of System~3's
future-state and value-derived conditioning signals?
(5)~Does heterogeneous pretraining produce executable out-of-the-box
behavior and generalization beyond task configurations observed during
data collection?
(6)~How much can task-specific post-training and subsequent
experience-driven reinforcement learning further improve performance?

\subsection{Experimental Setup}
\label{sec:exp_setup}

\paragraph{Evaluation platforms.}
We conduct real-world evaluations on two robot platforms: the AgileX PiPER/PiPER-X, a low-cost dual-arm manipulator widely used for tabletop manipulation benchmarking, and our in-house Maker~H01, an embodied-native humanoid platform with a head, waist, and bimanual arms.
The two platforms differ substantially in degrees of freedom, control interface, camera configuration, and workspace geometry, providing a rigorous test of cross-embodiment generalization.
 
\paragraph{Baselines.}
For real-world policy evaluation, we compare GigaBrain-0.7 against the following baselines:
(i)~$\pi_{0.5}$~\citep{pi05}, a state-of-the-art VLA model with heterogeneous co-training and hybrid discrete--continuous supervision;
(ii)~GigaBrain-0.1, the predecessor model in the GigaBrain series with PaliGemma2 backbone;
(iii)~Xiaomi-Robotics-1~\citep{xiaomi_robotics_1};
(iv)~Galaxea~G0.5~\citep{galaxea_g05}.
These real-world policy baselines are evaluated under the same protocol and task definitions. Separately, the embodied vision-language benchmark evaluation includes Xiaomi-Robotics-0 and G0.5-base, as well as Spirit-v1.5~\citep{spiritv15}, 
Wall-OSS-0.5~\citep{walloss05}, 
and Hy-Embodied-0.5-VLA-UMI~\citep{hyvla05}.
 
\paragraph{Evaluation protocol.}
Each task is evaluated over a fixed number of trials (typically 10--20 per configuration).
We report success rate as the primary metric, defined as the fraction of trials in which the robot completes the full task objective.
For language-following evaluations, each task is conditioned on a specific natural-language prompt that specifies attributes such as color, target identity, or spatial direction (e.g., ``pick up the \emph{red} spoon''), and a trial is considered successful only if the correct target is manipulated.
We distinguish between \textbf{in-distribution} (ID) tasks whose objects and configurations are represented in the training set and \textbf{out-of-distribution} (OOD) tasks that require generalization to unseen objects, layouts, or instructions.
 
\paragraph{Post-training.}
For post-training evaluations, all models are fine-tuned on
task-specific demonstration data collected on the corresponding target platform.
The resulting policies are evaluated using the same task definitions
and evaluation protocol described above.

\subsection{Backbone and Architecture Comparison}
\label{sec:architecture_comparison}

Before studying data scaling, we first establish the System~1 configuration
used in the subsequent experiments.
We compare candidate VLM backbones and alternative mechanisms for coupling
the vision-language model with the continuous action expert.
These experiments are intended to identify a robust model configuration for
large-scale pretraining.

\paragraph{VLM backbone comparison.}

We first compare PaliGemma2\citep{paligemma2}, Qwen3.5\citep{qwen3.5}, and Gemma~4\citep{gemma4} while keeping the
dual-stream VLA formulation fixed.
The models are evaluated on representative real-robot tasks spanning
structured manipulation, language-conditioned object interaction, and
long-horizon deformable-object manipulation.
\begin{table}[htpb]
\centering
\caption{
\textbf{Comparison of VLM backbones for System~1.}
The evaluated configurations differ in model size and image resolution;
the comparison is used to determine the backbone for subsequent experiments
rather than as a controlled model-scaling study. Model size denotes the total parameter count of the full model, including both the VLM backbone and the action expert.
\textsuperscript{*}Gemma~4 Clean Desk was evaluated under a limited-range
position-generalization setting.
}
\label{tab:backbone_comparison}

\small
\setlength{\tabcolsep}{4pt}
\renewcommand{\arraystretch}{1.05}

\begin{tabular}{lcccc}
\toprule
\textbf{Backbone}
& \textbf{Model Size}
& \makecell{Clean\\Desk}
& \makecell{Fruit\\Picking}
& \makecell{Shirt\\Folding} \\
\midrule

PaliGemma2
& 3.5B
& 50\%
& 88\%
& 30\% \\

Qwen3.5
& 5B
& 60\%
& 12\%
& 0\% \\

Gemma~4
& 8.5B
& 100\%\textsuperscript{*}
& 92\%
& 0\% \\

\bottomrule
\end{tabular}
\end{table}
As shown in Tab.~\ref{tab:backbone_comparison}, increasing backbone size
does not lead to uniformly stronger robot performance in the evaluated
setting.
Gemma~4 performs strongly on the more structured tasks, whereas PaliGemma2
is the only backbone that achieves nonzero success on shirt folding while
remaining competitive on fruit manipulation.
We therefore use PaliGemma2 as the backbone for the following architecture
comparison and data-scaling experiments.

\paragraph{VLM--action expert coupling.}

With the backbone fixed to PaliGemma2, we next compare three ways of
connecting the vision-language representation to the continuous action
expert:
\emph{Dual Stream},
\emph{Last-Layer Cross Attention}, and
\emph{Multi-Layer Cross Attention}.
A dual-stream expert interleaves with the VLM at every layer, attending to concatenated VLM–expert keys and values within a single attention computation, without any separate cross-attention module. A last-layer cross-attention expert instead decouples the action head entirely from the backbone, with every expert block cross-attending only to the VLM's final-layer hidden states. A multi-layer cross-attention expert maintains a per-layer interface, with expert block i cross-attending to the hidden states of the corresponding VLM layer i. The three variants thus span a spectrum of coupling strength: from per-layer interleaving, through a per-layer interface, to a single terminal representation, progressively decoupling the expert from the backbone at the cost of a narrower information pathway.

\begin{table}[htpb]
\centering
\caption{
\textbf{Comparison of VLM--action expert coupling architectures.}
All variants use PaliGemma2 as the VLM backbone.
Success rates are measured on real robots; training and inference costs are
reported for the corresponding configurations.
}
\label{tab:action_architecture_comparison}

\small
\setlength{\tabcolsep}{4pt}
\renewcommand{\arraystretch}{1.05}

\begin{tabular}{lccccc}
\toprule
\textbf{Architecture}
& \makecell{Clean\\Desk}
& \makecell{Fruit\\Picking}
& \makecell{Shirt\\Folding}
& \makecell{Train\\s/step}
& \makecell{Infer.\\s} \\
\midrule

Dual Stream
& 50\%
& 88\%
& 30\%
& 4.93
& 0.221 \\

Last-Layer Cross Attention
& 20\%
& 40\%
& 0\%
& 4.65
& 0.073 \\

Multi-Layer Cross Attention
& 20\%
& 36\%
& 0\%
& 6.59
& 0.108 \\

\bottomrule
\end{tabular}
\end{table}

Tab.~\ref{tab:action_architecture_comparison} shows a clear trade-off
between action-generation capability and computational efficiency.
Restricting the action expert to the final VLM representation yields the
lowest inference latency, whereas the multi-layer cross-attention variant
incurs additional computational cost without improving task success in the
evaluated settings.
In contrast, the dual-stream architecture achieves the strongest
real-robot performance across all three tasks, including nonzero success
on the more challenging shirt-folding task, although at a higher inference
cost.

Based on these comparisons, we adopt the PaliGemma2-based dual-stream
architecture as the default System~1 configuration in the remainder of the
experiments.
This fixes the model architecture before we study the effect of increasing
the scale and diversity of the pretraining data.

\subsection{Data Scaling}
\label{sec:data_scaling}

With the model architecture fixed, we next study how GigaBrain-0.7 benefits
from scaling heterogeneous embodied data.
We consider three complementary questions:
whether increasing the overall amount of pretraining data improves
optimization,
whether additional robot experience translates into stronger real-robot
capability,
and whether human demonstrations provide complementary supervision beyond
robot trajectories alone.

\begin{figure*}[htb]
    \centering
    \captionsetup{
        type=figure,
        justification=justified,
        singlelinecheck=false
    }
    \captionsetup[subfigure]{
        justification=centering,
        singlelinecheck=true
    }

    \begin{subfigure}[t]{0.72\textwidth}
        \centering
        \includegraphics[width=\linewidth]
        {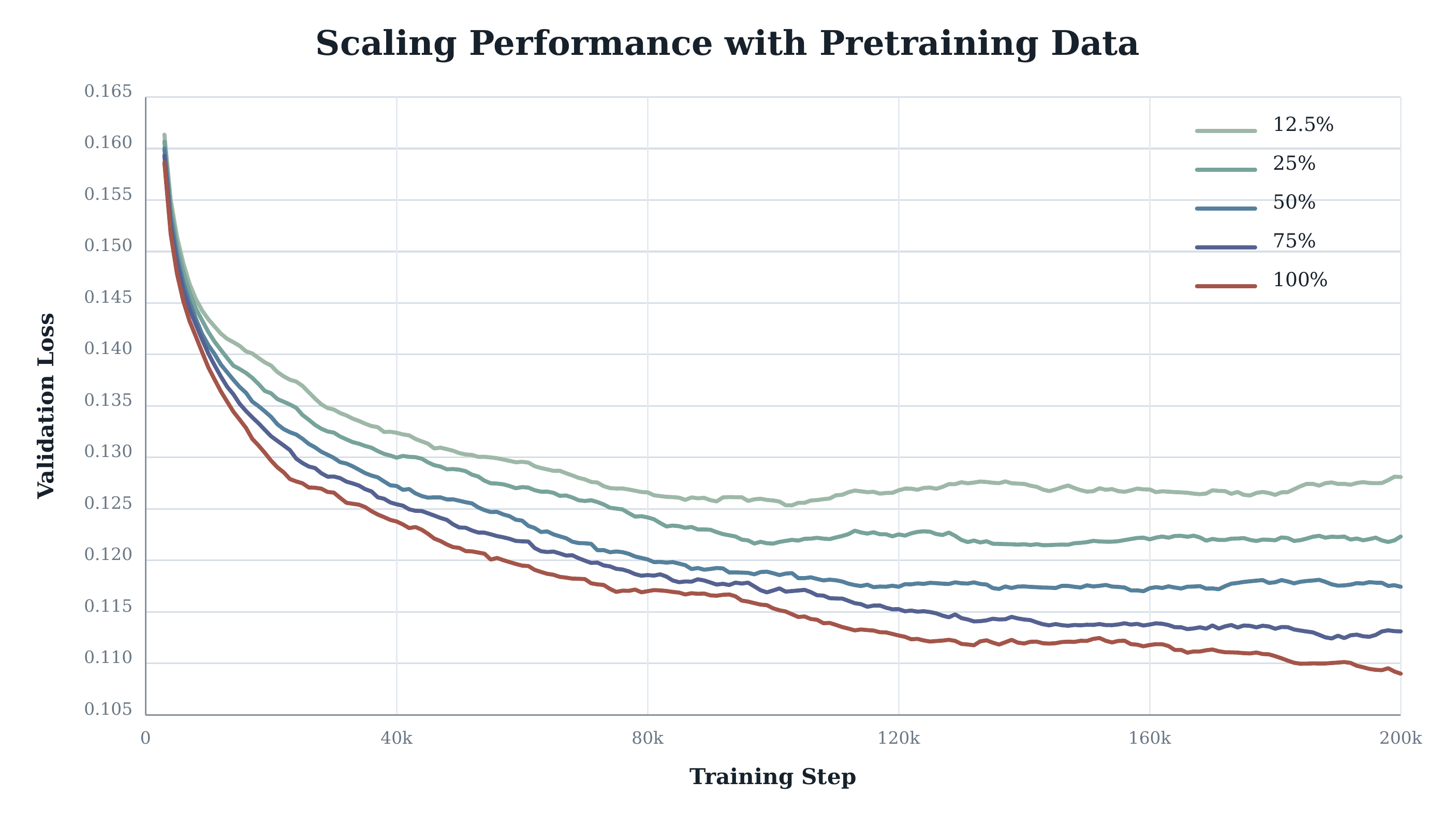}
        \caption{
        Validation loss under different pretraining-data scales.
        }
        \label{fig:data_scaling_loss}
    \end{subfigure}

    \par\vspace{0.5em}

    \makebox[\textwidth][c]{%
        \begin{subfigure}[t]{0.4\textwidth}
            \centering
            \includegraphics[width=\linewidth]
            {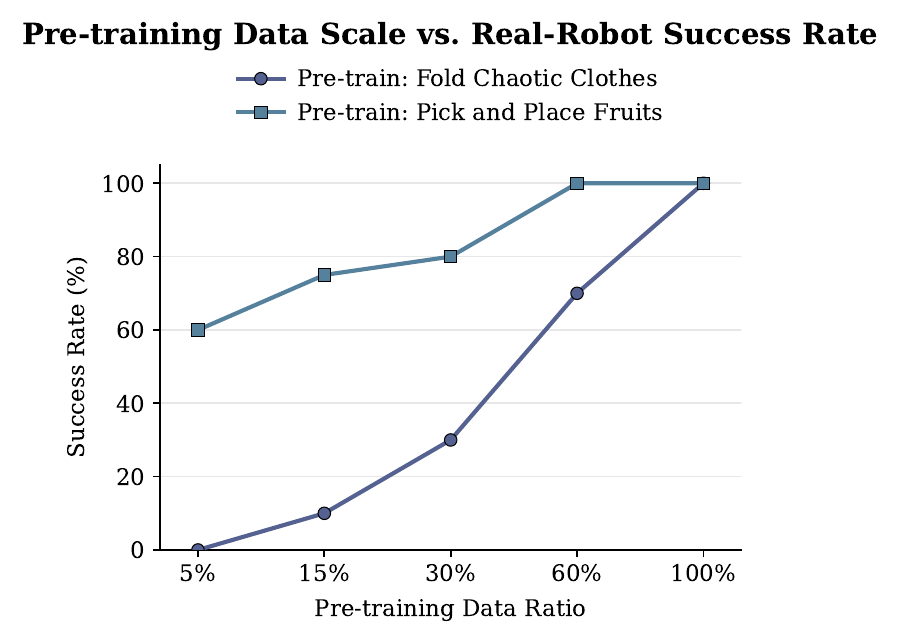}
            \caption{
            Real-robot performance as robot-data scale increases.
            }
            \label{fig:data_scaling_robot}
        \end{subfigure}
        \hspace{0.02\textwidth}
        \begin{subfigure}[t]{0.4\textwidth}
            \centering
            \includegraphics[width=\linewidth]
            {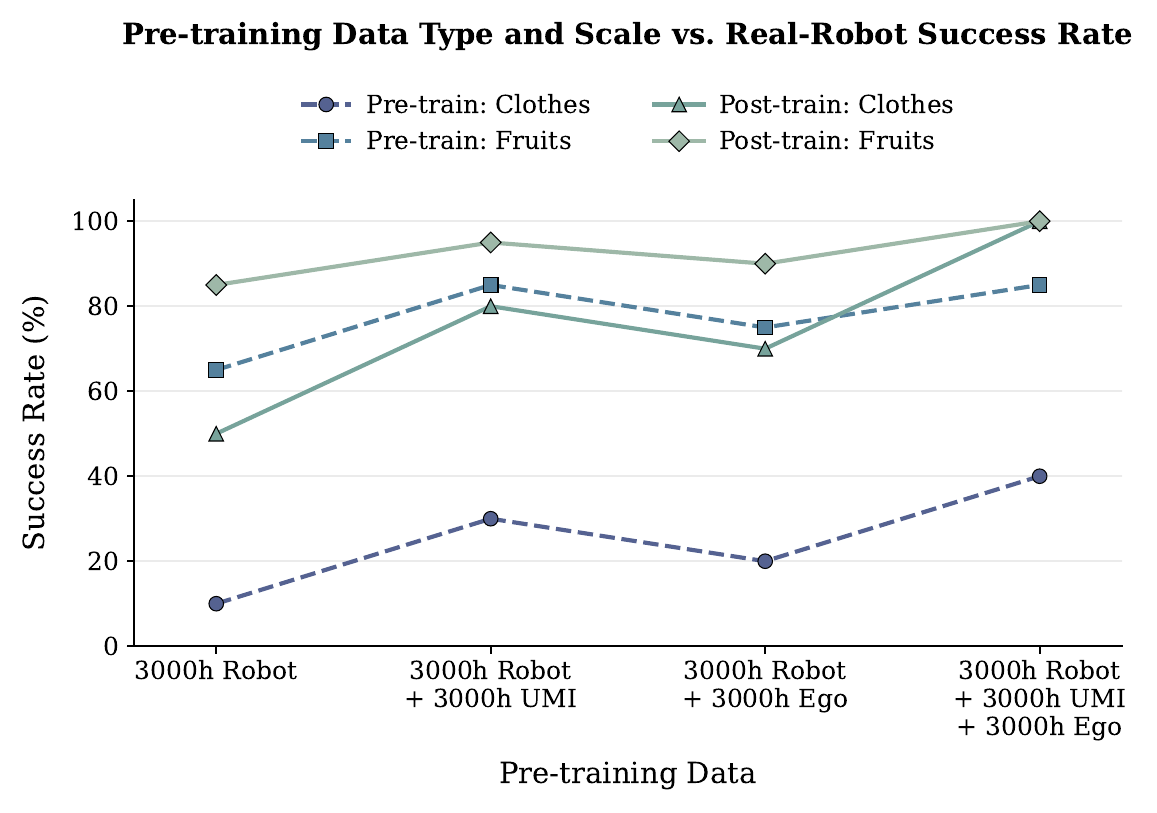}
            \caption{
            Effect of UMI and EGO data in the pretraining mixture.
            }
            \label{fig:data_scaling_sources}
        \end{subfigure}%
    }

    \caption{
    \textbf{Scaling GigaBrain-0.7 with heterogeneous embodied data.}
    (a) Increasing the amount of pretraining data consistently reduces
    validation loss under a fixed model configuration.
    (b) Increasing robot-data scale improves real-robot performance,
    with a stronger effect on the more challenging deformable-object task.
    (c) UMI and EGO provide complementary supervision to robot trajectories,
    and their combination yields the strongest overall performance.
    }
    \label{fig:data_scaling}
\end{figure*}

\paragraph{Scaling the pretraining data.}

We first vary the amount of pretraining data while keeping the model
configuration unchanged.
As shown in Fig.~\ref{fig:data_scaling_loss}, larger training sets
consistently reach lower validation loss, with the separation between data
scales remaining visible throughout optimization.
Within the evaluated range, we therefore observe continued improvement as
the available pretraining data increases.

\paragraph{Scaling robot experience.}

We next examine whether the same trend translates to physical execution.
Fig.~\ref{fig:data_scaling_robot} evaluates models trained with
progressively larger subsets of the robot data on fruit manipulation and
chaotic clothes folding.

Both tasks benefit from additional robot experience, but the scaling behavior
differs substantially with task complexity.
Fruit manipulation improves relatively smoothly, whereas clothes folding
shows a much sharper dependence on data scale.
The latter requires sustained interaction with deformable objects and a
broader range of intermediate states, suggesting that complex long-horizon
skills place substantially greater demands on the coverage of the pretraining
distribution.
Together with the consistently decreasing validation loss, this trend
suggests that scaling does more than improve the training objective:
manipulation behaviors that are weak or unreliable at smaller scales
become increasingly reliable as embodied experience grows.
We view this progression as a direct manifestation of capability
emergence with scale.

\paragraph{Scaling across data sources.}

Finally, we investigate whether scaling data diversity provides benefits
beyond increasing robot trajectories alone.
Starting from 3,000~hours of robot data, we add equal amounts of UMI or EGO
demonstrations, and further evaluate their combination.

As shown in Fig.~\ref{fig:data_scaling_sources}, both human-data sources
improve performance over robot-only pretraining, and the combination of
robot, UMI, and EGO data provides the strongest results across the evaluated
settings.
The improvement persists after task-specific post-training, indicating that
the additional human experience changes the quality of the learned
pretraining prior rather than providing only an immediate improvement to the
base policy.

The different sources also exhibit complementary effects.
Robot trajectories provide direct embodiment-specific control supervision,
UMI contributes manipulation experience from a near-hand observation and
motion interface, and EGO broadens the diversity of first-person human
interaction.
Their combined benefit supports the central motivation of our data pyramid:
effective scaling depends not only on the number of trajectories, but also on
bringing heterogeneous forms of physical experience into a common training
representation.

\subsection{Temporal Context Analysis}
\label{sec:temporal_analysis}

Many manipulation tasks contain visually similar states that recur at
different stages of execution.
When action generation relies only on the current observation, these
states can become ambiguous because similar visual configurations may
require different subsequent behaviors depending on the preceding
interaction history.
We qualitatively examine this effect by comparing otherwise matched
policies with and without temporal visual context.

\begin{figure*}[t]
    \centering
    \captionsetup{
        type=figure,
        justification=justified,
        singlelinecheck=false
    }
    \captionsetup[subfigure]{
        justification=centering,
        singlelinecheck=false
    }

    \begin{subfigure}[t]{0.96\textwidth}
    \centering
    \includegraphics[
    width=\linewidth
    ]{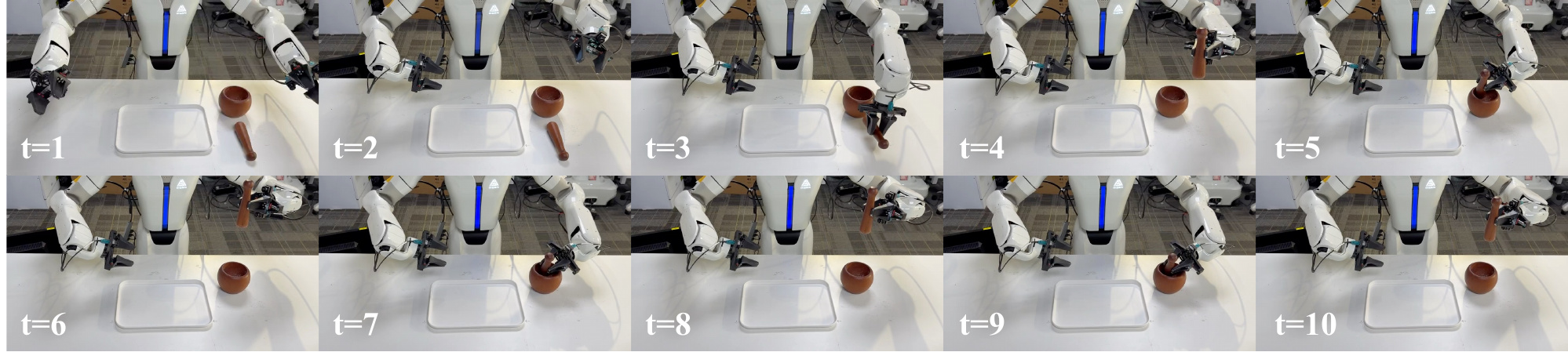}
    \caption{
    \textbf{Without temporal context.}
    The policy repeatedly revisits similar interaction states and
    falls into a repeated-action cycle.
    }
    \label{fig:temporal_without_context}
    \end{subfigure}
    
    \vspace{4pt}
    
    \begin{subfigure}[t]{0.96\textwidth}
    \centering
    \includegraphics[
    width=\linewidth
    ]{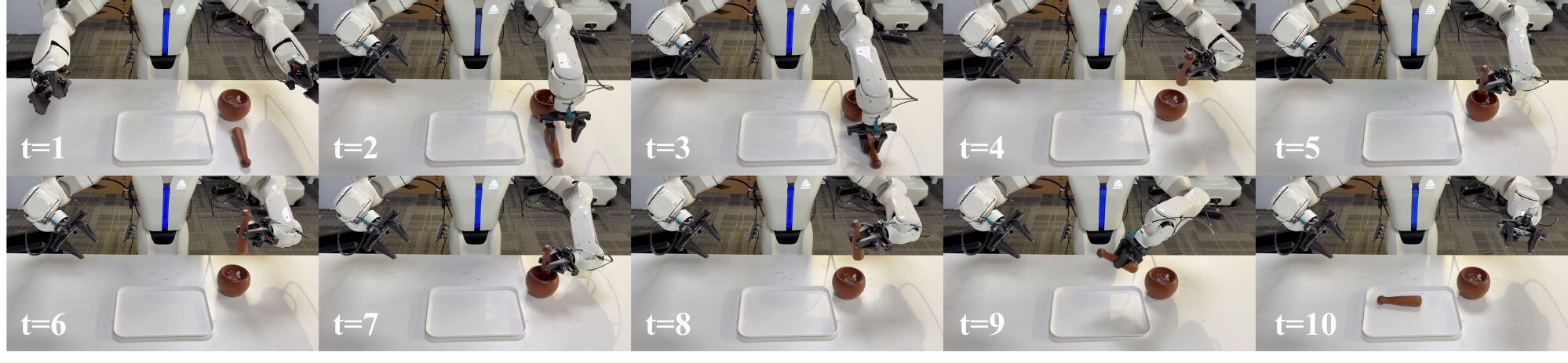}
    \caption{
    \textbf{With temporal context.}
    Historical observations help disambiguate the revisited state,
    allowing the policy to exit the repeated interaction and continue
    task execution.
    }
    \label{fig:temporal_with_context}
    \end{subfigure}

    \caption{
    \textbf{Qualitative analysis of temporal context in repeated-state
    manipulation.}
    Each rollout is represented by ten temporally ordered frames.
    Without temporal context, the policy repeatedly returns to visually
    similar interaction states and produces a recurring action pattern.
    With temporal context, recent interaction history provides additional
    information about the current stage of execution, enabling the policy
    to break the repeated-action cycle and continue toward the task goal.
    }
    \label{fig:temporal_context_analysis}
\end{figure*}

As illustrated in Fig.~\ref{fig:temporal_context_analysis}, the
single-frame policy can become trapped in a repeated-action cycle when
the current observation closely resembles states encountered earlier in
the episode.
Although these instantaneous visual observations are similar, they occur
at different stages of the interaction and do not necessarily require the
same subsequent behavior.

Temporal context provides information about the preceding interaction
trajectory and helps distinguish such visually similar but temporally
distinct states.
In the representative example, the policy with temporal context eventually
departs from the recurring interaction pattern and proceeds to the next
stage of the task, whereas the single-frame policy continues to repeat
similar actions.
This qualitative comparison illustrates that temporal modeling can support
closed-loop manipulation not only by retaining longer-horizon context, but
also by resolving local ambiguity when similar visual states recur during
execution.

\subsection{System~3 Prediction and Evaluation Analysis}
\label{sec:system3_ablation}

\subsubsection{Prediction and Evaluation}
\begin{figure*}[t]
\centering
\captionsetup{
        type=figure,
        justification=justified,
        singlelinecheck=false
    }
\includegraphics[width=\textwidth]
{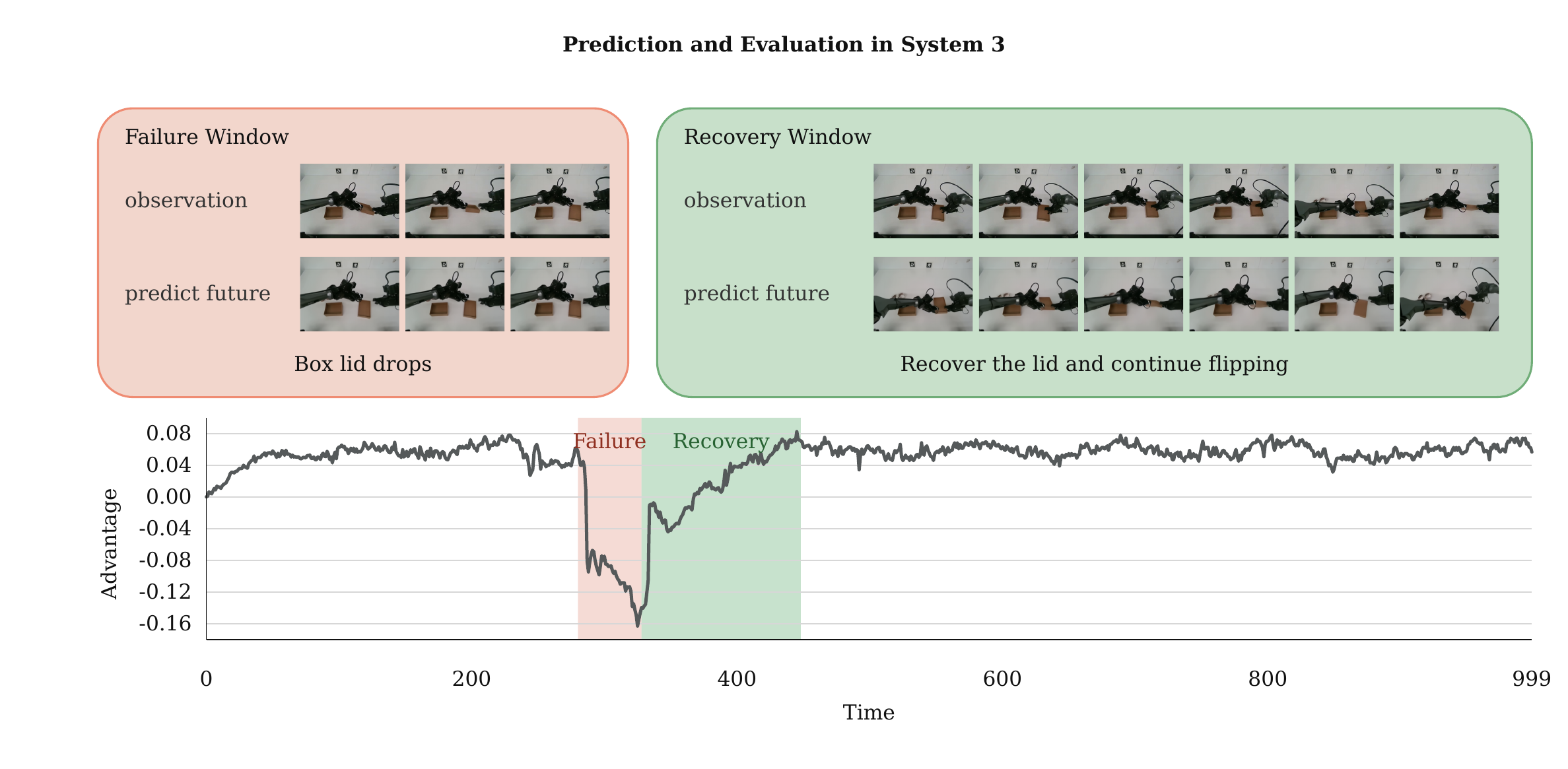}

\caption{
\textbf{Prediction and evaluation with the World Value Model.}
The model predicts both future visual observations and future state values.
The inferred advantage decreases sharply when the imagined trajectory leads
to failure and increases when the trajectory recovers toward successful task
execution.
}
\label{fig:system3_prediction_evaluation}
\end{figure*}

System 3 jointly predicts future visual observations and future
state values. The predicted future state values can then be used to estimate
the advantage of a candidate action trajectory. As shown in
Fig.~\ref{fig:system3_prediction_evaluation}, when the imagined future
trajectory results in task failure, the estimated advantage exhibits a
pronounced decrease. In contrast, once the trajectory recovers and returns to
a successful execution mode, the advantage increases substantially. These
results suggest that the model captures value-relevant changes in predicted
future outcomes and can use them to distinguish unfavorable failure
trajectories from recovered task-progressing behaviors.

Compared to $\pi_{0.7}$\citep{pi07} and $\pi^{*}_{0.6}$~\citep{pi06}, our assessment of action
quality does not rely directly on human annotations or VLM-based reasoning.
Instead, it is derived from a more fine-grained and dense action evaluation
produced by the world model through future imagination. This approach yields
action-value estimates that are more closely aligned with the resulting
rollout outcomes and exhibits greater robustness in generalization across
scenarios.
Compared with GigaBrain-0.5M*, we extend the future prediction horizon beyond the length of the action chunk. We find that this longer horizon improves the success rate on complex tasks and enhances the robustness of the model.

\subsubsection{Ablation in System 3}
To isolate future image and value effects, we compare four configurations:
\textsc{Base},
\textsc{+SubImage},
\textsc{+Value}, and
\textsc{+SubImage+Value}.

The four variants are evaluated on clothes folding with AgileX PiPER,
gift wrapping with AgileX PiPER-X, and cube sorting with Maker~H01.
We jointly report task success, average task score, and completion time,
since success alone can become saturated on easier tasks.

\begin{figure*}[t]
\centering

\captionsetup{
    justification=justified,
    singlelinecheck=false
}

\captionsetup[subfigure]{
    justification=centering,
    singlelinecheck=false
}

\begin{subfigure}[t]{0.32\textwidth}
    \centering
    \includegraphics[width=\linewidth]
    {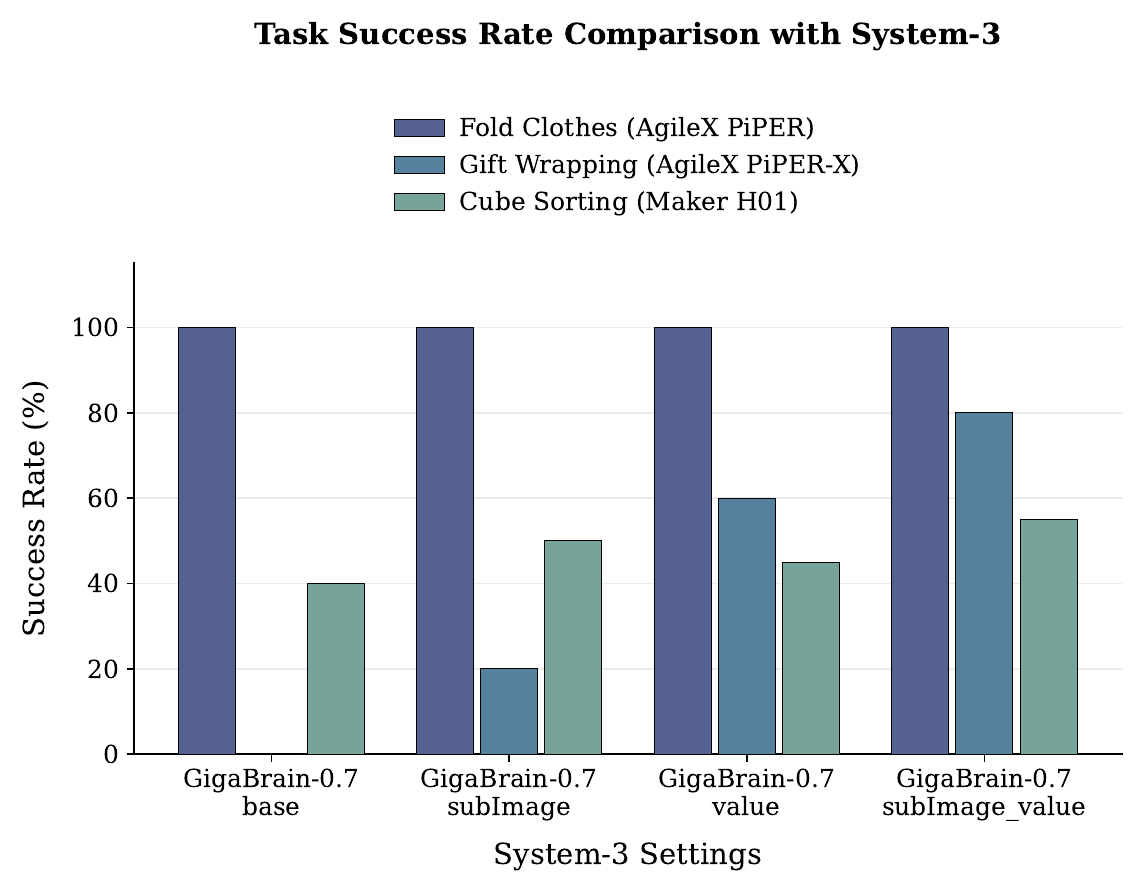}
    \caption{Task success rate.}
    \label{fig:system3_success}
\end{subfigure}
\hfill
\begin{subfigure}[t]{0.32\textwidth}
    \centering
    \includegraphics[width=\linewidth]
    {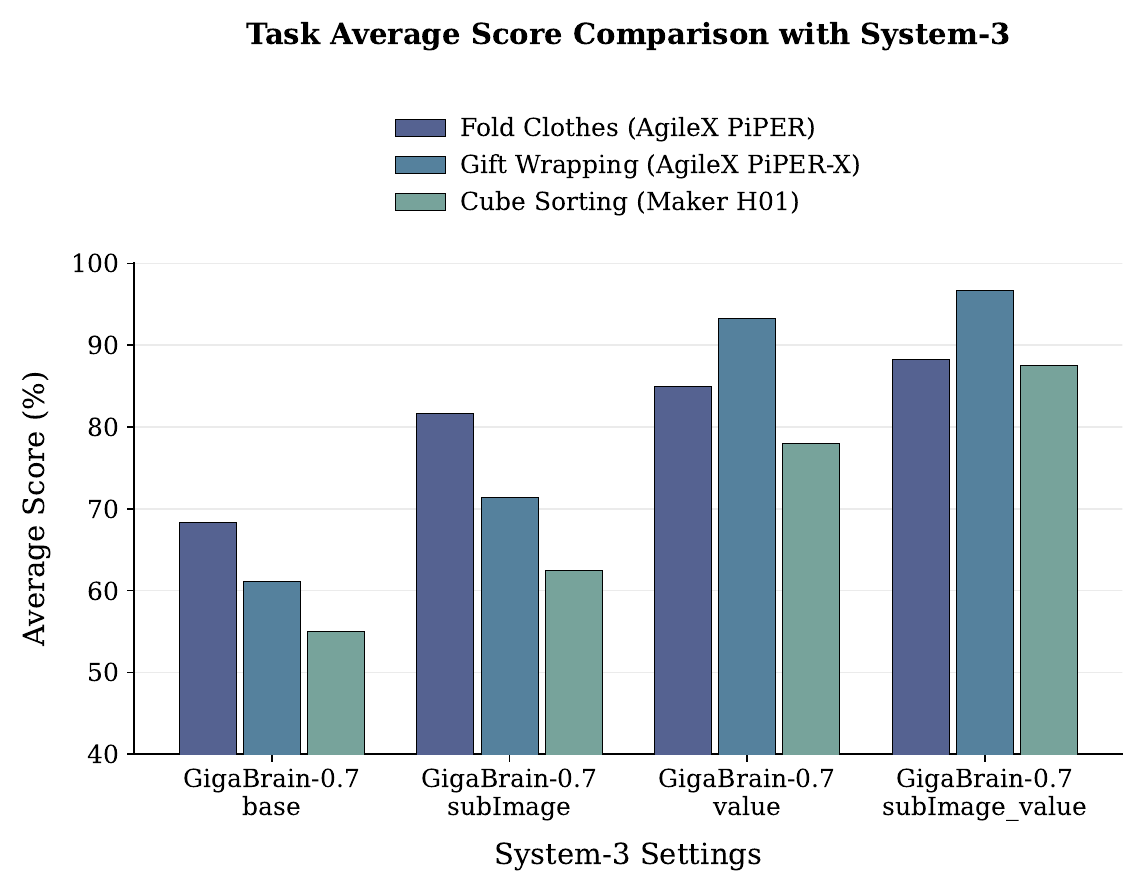}
    \caption{Average task score.}
    \label{fig:system3_score}
\end{subfigure}
\hfill
\begin{subfigure}[t]{0.32\textwidth}
    \centering
    \includegraphics[width=\linewidth]
    {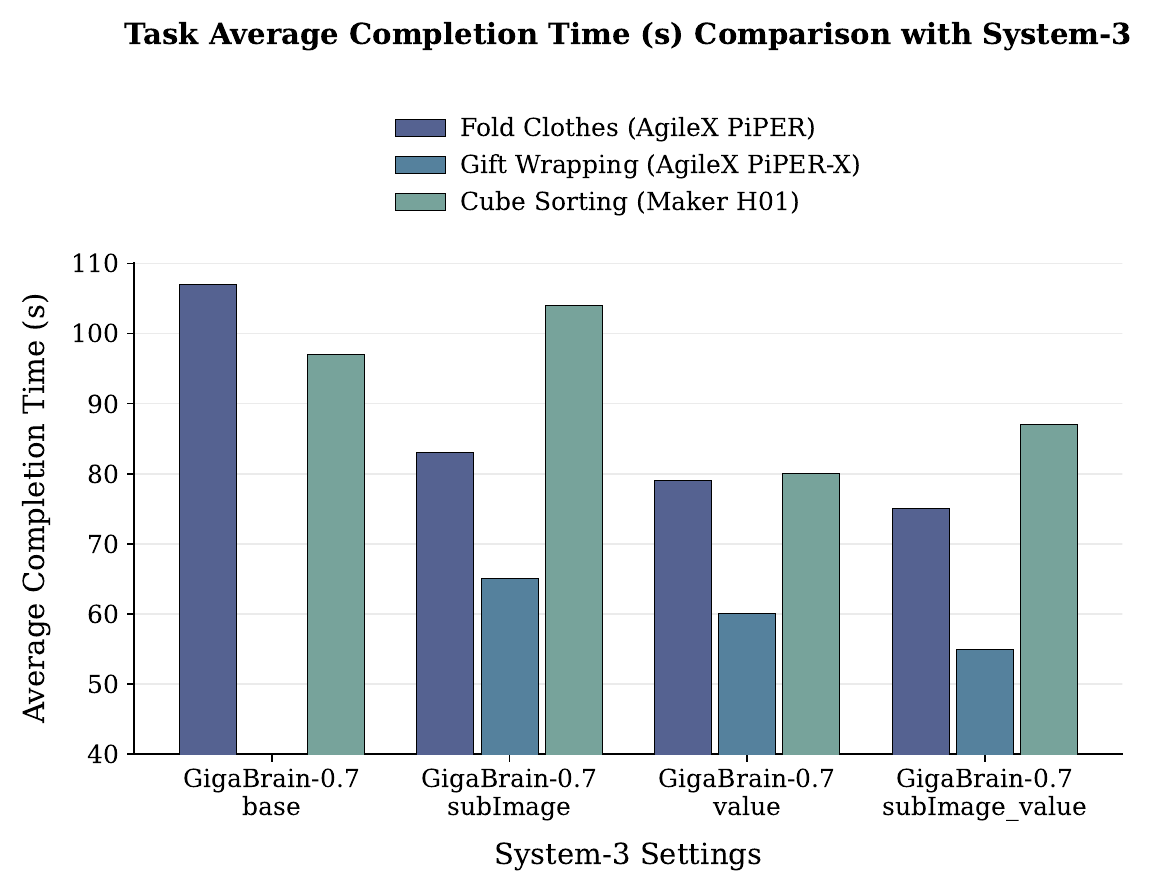}
    \caption{Average completion time.}
    \label{fig:system3_time}
\end{subfigure}

\caption{
\textbf{Ablation of the prediction and evaluation interfaces in System~3.}
The base VLA is compared with image conditioning
(\textsc{+SubImage}), value conditioning (\textsc{+Value}),
and their combination.
Completion time is reported only when successful task completion is
observed.
}
\label{fig:system3_ablation}

\end{figure*}
On clothes folding, all four configurations achieve 100\% task success.
The additional System~3 signals nevertheless improve both task score and
execution efficiency.
The average score increases from 68.3\% for \textsc{Base} to
81.7\%, 85.0\%, and 88.3\% for
\textsc{+SubImage}, \textsc{+Value}, and
\textsc{+SubImage+Value}, respectively.
Average completion time decreases from 107\,s to 83\,s, 79\,s, and 75\,s.

The effect is more pronounced on gift wrapping.
The base model fails to complete the task, while
\textsc{+SubImage}, \textsc{+Value}, and
\textsc{+SubImage+Value} achieve 20\%, 60\%, and 80\% success.
The corresponding task scores increase from 61.1\% to
71.4\%, 93.3\%, and 96.7\%.
Among variants with successful episodes, completion time decreases from
65\,s to 60\,s and finally 55\,s.

Cube sorting exhibits a more mixed pattern.
Success increases from 40\% for the base model to
50\% with image conditioning, 45\% with value conditioning,
and 55\% when both are enabled.
The combined model obtains the highest average task score, increasing from
55.0\% to 87.5\%.
Value conditioning alone yields the shortest observed completion time
(80\,s), whereas the full configuration provides the strongest overall
success and progress metrics.

Taken together, these experiments show that System~3 provides effective
\textbf{predictive and progress-aware guidance} for task execution.
Future-state and value conditioning improve not only whether a task is
completed, but also how effectively the policy progresses through
imperfect intermediate states, as reflected by higher task scores and
shorter completion times.
This effect remains visible even when binary success is already saturated, and becomes more pronounced on the more challenging gift-wrapping task.
These results indicate that predictive and evaluative context can improve the robustness of behaviors acquired by the underlying policy, rather than merely increasing task success.

\subsection{Foundation Model Evaluation}
\label{sec:foundation_eval}

We next evaluate GigaBrain-0.7 directly after pretraining, without
task-specific adaptation.
Rather than measuring specialization on a small set of downstream tasks,
this evaluation asks whether large-scale heterogeneous pretraining produces
a policy that can already execute diverse instructions and manipulation
behaviors out of the box.

We evaluate the model on both AgileX PiPER and Maker~H01, covering
language-conditioned multi-task execution and longer-horizon manipulation.
For each platform, we additionally introduce out-of-distribution settings
that alter task configurations, objects, or interaction requirements beyond
those represented by the corresponding in-distribution evaluation.

\subsubsection{Out-of-the-Box Language Following}
\label{sec:foundation_language}


\begin{figure*}[t]
    \centering
    \captionsetup{
        type=figure,
        justification=justified,
        singlelinecheck=false
    }
    \includegraphics[width=0.80\textwidth]
    {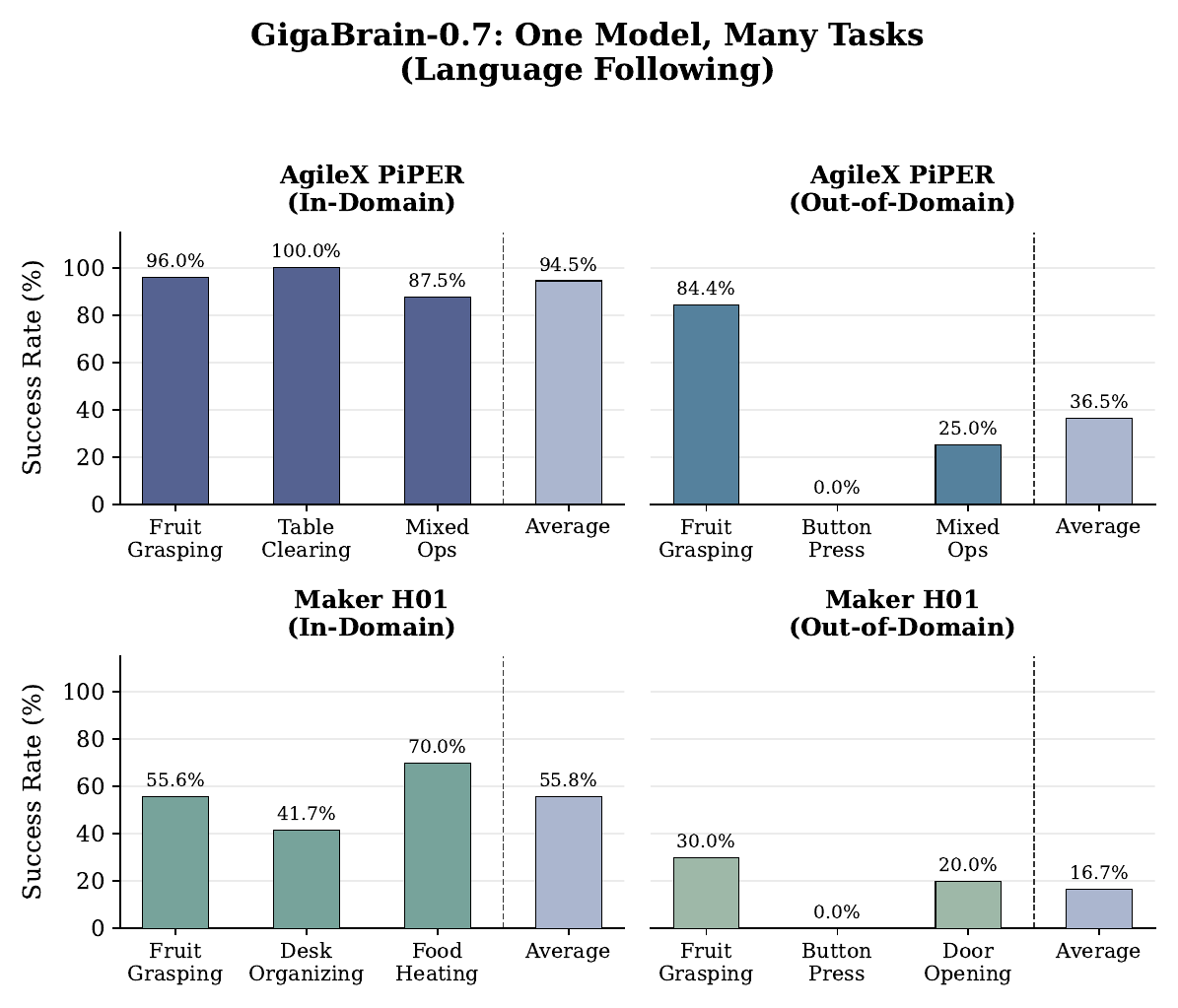}
    \caption{
    \textbf{Out-of-the-box multi-task language following of GigaBrain-0.7.}
    We evaluate the same pretrained policy on AgileX PiPER and Maker~H01
    under both in-distribution and out-of-distribution settings.
    }
    \label{fig:foundation_language}
\end{figure*}


\begin{figure*}[htbp]
    \centering
    \captionsetup{
        type=figure,
        justification=justified,
        singlelinecheck=false
    }

    \begin{subfigure}[t]{\textwidth}
        \centering
        \setlength{\tabcolsep}{1.5pt}
        \renewcommand{\arraystretch}{1.02}

        \begin{tabularx}{\linewidth}{
            @{}
            *{5}{>{\centering\arraybackslash}X}
            @{}
        }

        \includegraphics[width=\linewidth]
        {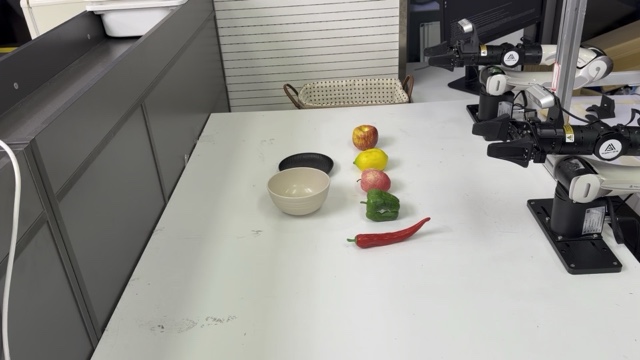}
        &
        \includegraphics[width=\linewidth]
        {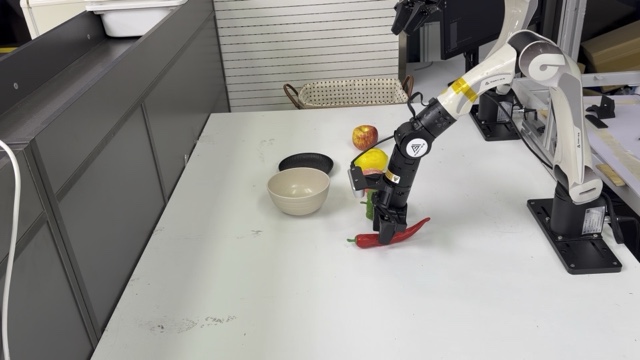}
        &
        \includegraphics[width=\linewidth]
        {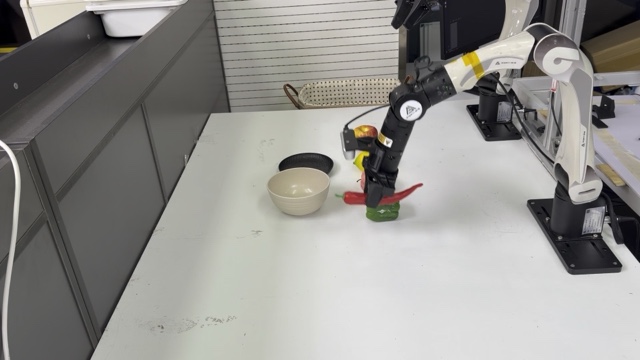}
        &
        \includegraphics[width=\linewidth]
        {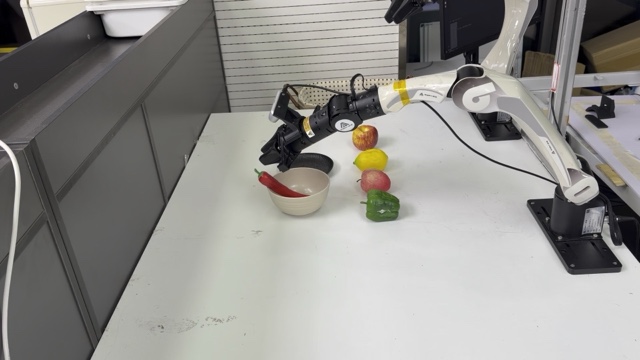}
        &
        \includegraphics[width=\linewidth]
        {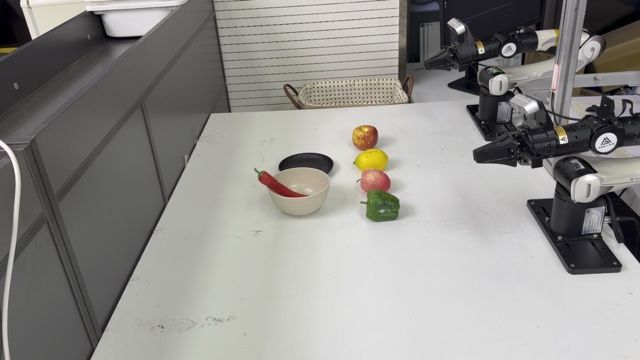}
        \\[1pt]

        \includegraphics[width=\linewidth]
        {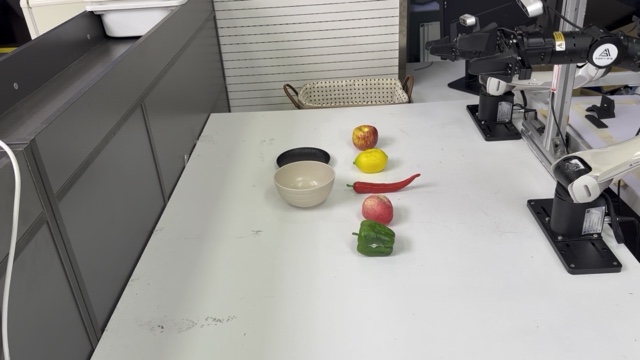}
        &
        \includegraphics[width=\linewidth]
        {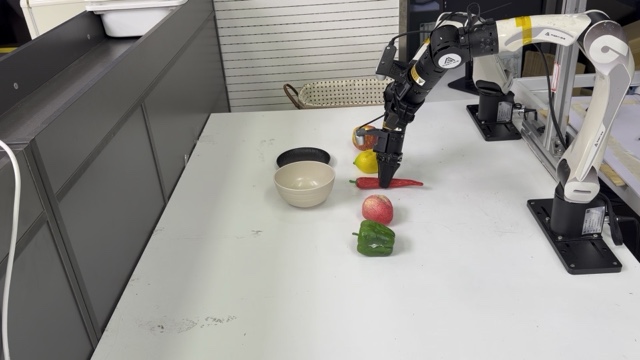}
        &
        \includegraphics[width=\linewidth]
        {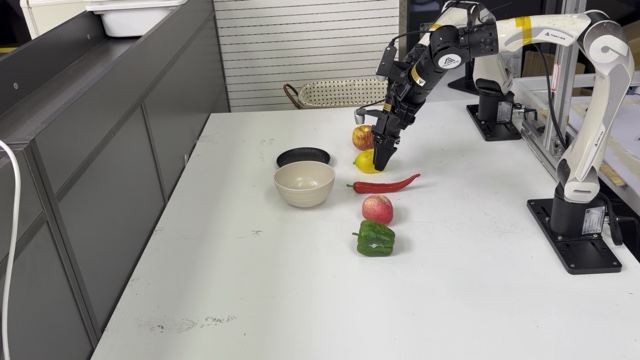}
        &
        \includegraphics[width=\linewidth]
        {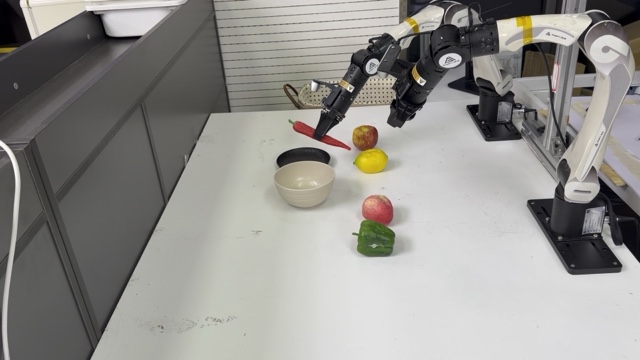}
        &
        \includegraphics[width=\linewidth]
        {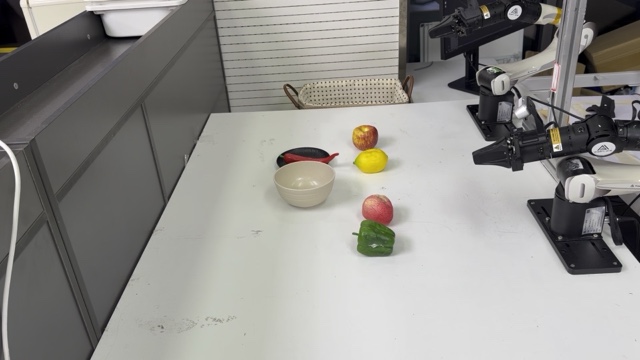}

        \end{tabularx}

        \caption{
AgileX PiPER: the target pepper instance is absent from the
 training data, while the two containers were not
collected together in the same task configuration. 
        }
        \label{fig:foundation_language_ood_agilex}
    \end{subfigure}

    \vspace{2pt}

    \begin{subfigure}[t]{\textwidth}
        \centering
        \setlength{\tabcolsep}{1.5pt}

        \begin{tabularx}{\linewidth}{
            @{}
            *{5}{>{\centering\arraybackslash}X}
            @{}
        }

        \includegraphics[width=\linewidth]
        {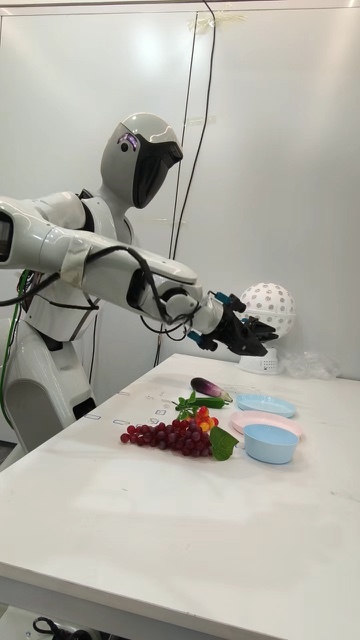}
        &
        \includegraphics[width=\linewidth]
        {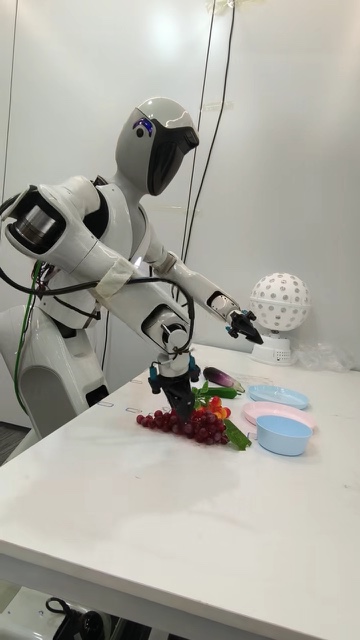}
        &
        \includegraphics[width=\linewidth]
        {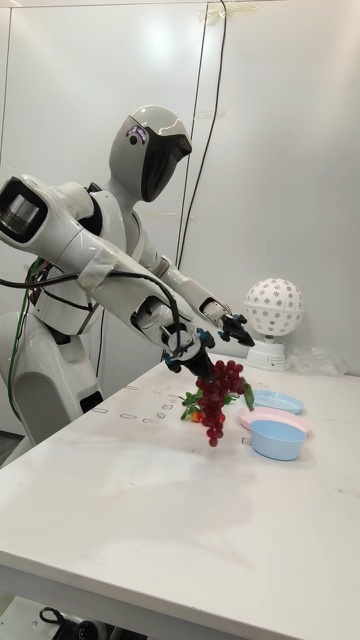}
        &
        \includegraphics[width=\linewidth]
        {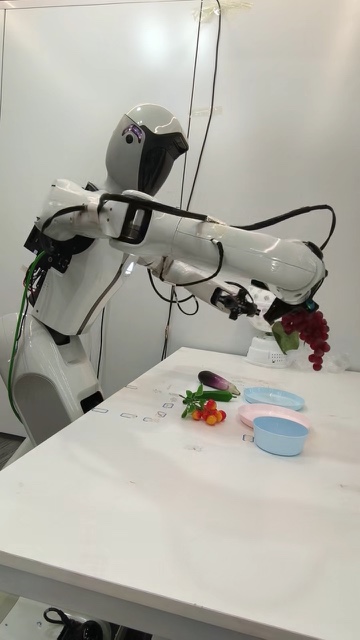}
        &
        \includegraphics[width=\linewidth]
        {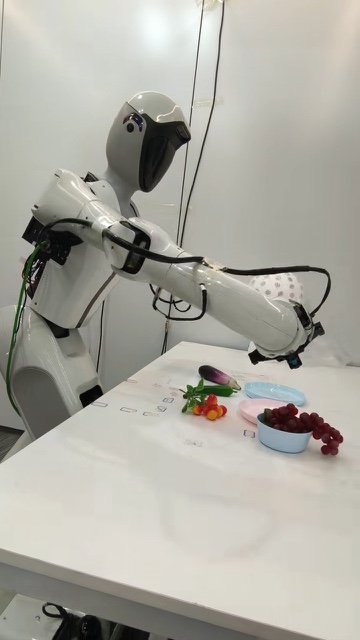}

        \end{tabularx}

        \caption{
        Maker H01: grapes are absent from the H01 training data, and neither the evaluation scene nor the task composition was observed during data collection.
        }
        \label{fig:foundation_language_ood_h01}
    \end{subfigure}

    \caption{
    \textbf{Object, spatial, and compositional generalization of the
pretrained GigaBrain-0.7 policy.}
    The same pretrained GigaBrain-0.7 policy is deployed on AgileX PiPER
    and Maker~H01 without task-specific adaptation.
    Panel~(a) evaluates a pepper instance that is absent from the
    corresponding training data; moreover, the two target containers
    were not collected together in the same task configuration and their
    relative placement is randomized at evaluation time.
    Panel~(b) introduces a broader compound distributional shift on
    Maker~H01: grapes are absent from the H01 training data, and neither
    the evaluation scene nor the corresponding task composition was
    observed during H01 data collection.
    }
    \label{fig:foundation_language_ood_rollouts}
\end{figure*}

As shown in Fig.~\ref{fig:foundation_language}, the same pretrained
GigaBrain-0.7 policy can follow multiple language instructions on both
AgileX PiPER and Maker~H01 without task-specific adaptation.
Across the in-distribution settings, the policy executes diverse
instruction-conditioned behaviors involving object selection, scene
rearrangement, and multi-step manipulation.
These results show that heterogeneous pretraining produces an executable
generalist policy before downstream adaptation, rather than merely
providing a stronger initialization for task-specific post-training.

We next evaluate the same pretrained policy under controlled
out-of-distribution settings that probe several distinct axes of
generalization.
Rather than changing only visual appearance, these evaluations vary
task-relevant object instances, spatial arrangements, scenes, and
compositions of previously acquired manipulation skills.
This follows the broader principle that foundation-model quality is better
revealed when the evaluation distribution departs from the configurations
observed during data collection.

Representative OOD executions are shown in
Fig.~\ref{fig:foundation_language_ood_rollouts}.
The two embodiments probe complementary forms of generalization.
On AgileX PiPER, the evaluated pepper instance is absent from the
corresponding training data.
More importantly, the two target containers were not collected together
in the same task configuration and their relative placement is
randomized during evaluation.
Successful execution therefore requires more than recognizing a
familiar scene template: the policy must identify the
language-specified target, localize the relevant container in the
current arrangement, and compose the appropriate pick-and-place
behavior for a previously unobserved object--container configuration.
Maker~H01 introduces a broader compound distributional shift.
The H01 training data contain neither the evaluated grapes nor the
test scene or the corresponding task composition.
Nevertheless, the pretrained policy follows the human instruction,
selects the relevant object, and completes the requested manipulation
without task-specific adaptation.
Together, these examples provide qualitative evidence that
heterogeneous pretraining gives rise to target understanding, spatial
grounding, and compositional generalization beyond the task
configurations explicitly observed during data collection.

The ID--OOD gap also highlights the current limits of foundation transfer.
Generalization remains stronger when the underlying interaction structure
is preserved and becomes more challenging when multiple factors---such as
object identity, scene configuration, and task composition---shift
simultaneously.
Both AgileX PiPER and Maker~H01 exhibit degradation under the
OOD settings, indicating that robust generalization across task
distributions and heterogeneous embodiments remains an important direction
for continued scaling.

\subsubsection{Out-of-the-Box Complex Manipulation}
\label{sec:foundation_complex}


\begin{figure*}[t]
    \centering
    \captionsetup{
        type=figure,
        justification=justified,
        singlelinecheck=false
    }
    \includegraphics[width=0.80\textwidth]
    {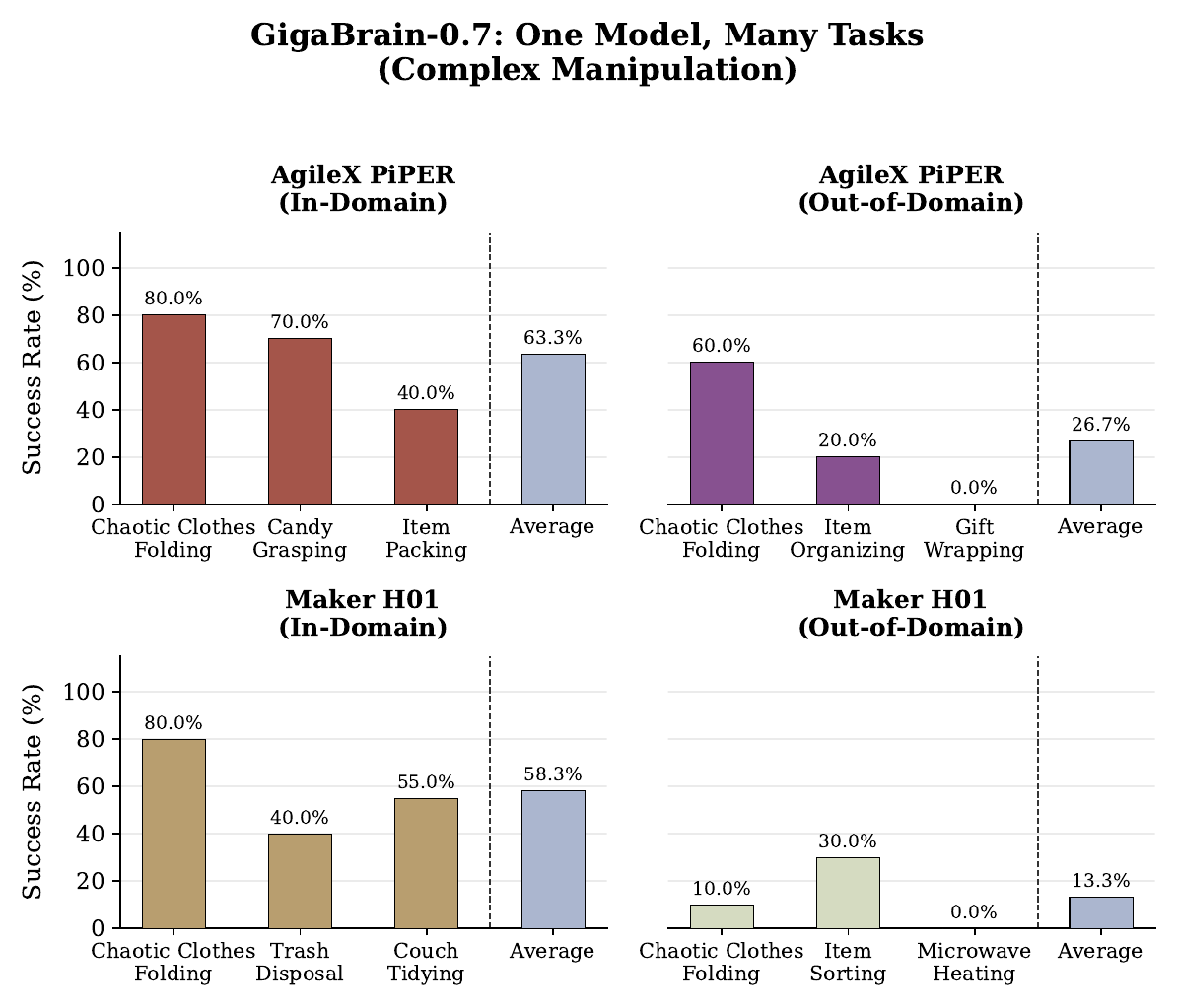}
    \caption{
    \textbf{Out-of-the-box complex manipulation with GigaBrain-0.7.}
    The evaluation covers longer-horizon manipulation under ID and OOD
    settings on AgileX PiPER and Maker~H01.
    }
    \label{fig:foundation_complex}
\end{figure*}


\begin{figure*}[htbp]
    \centering
    \captionsetup{
        type=figure,
        justification=justified,
        singlelinecheck=false
    }

    \begin{subfigure}[t]{\textwidth}
        \centering
        \setlength{\tabcolsep}{1.5pt}
        \renewcommand{\arraystretch}{1.02}

        \begin{tabularx}{\linewidth}{
            @{}
            *{5}{>{\centering\arraybackslash}X}
            @{}
        }

        \includegraphics[width=\linewidth]
        {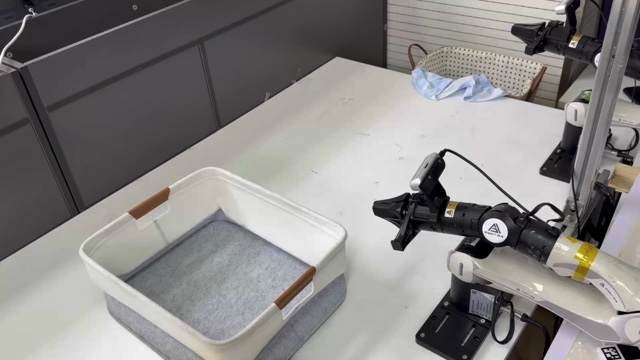}
        &
        \includegraphics[width=\linewidth]
        {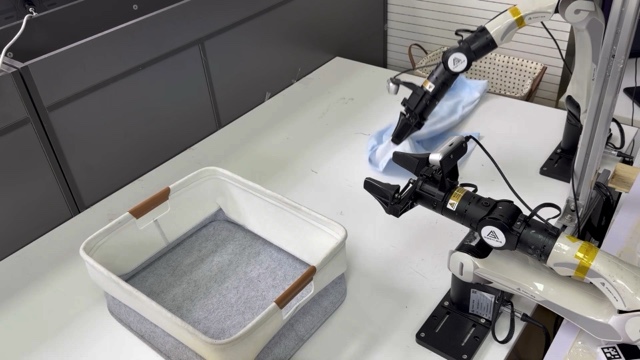}
        &
        \includegraphics[width=\linewidth]
        {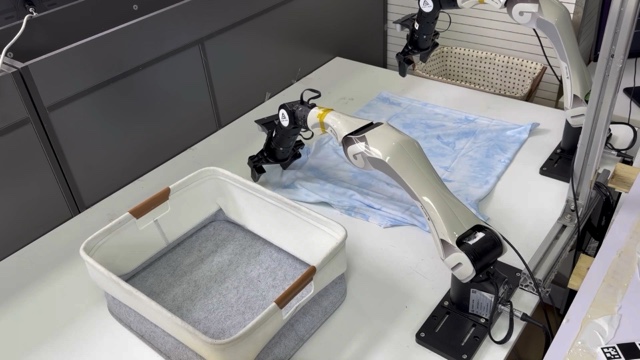}
        &
        \includegraphics[width=\linewidth]
        {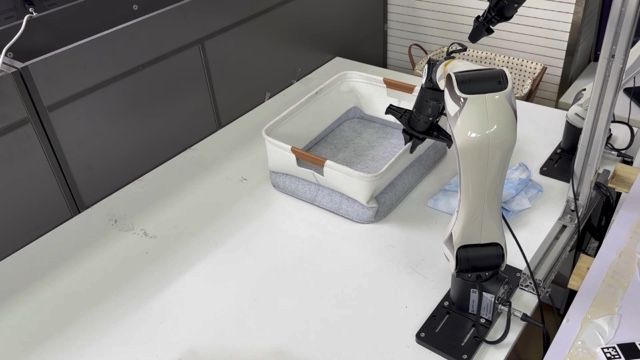}
        &
        \includegraphics[width=\linewidth]
        {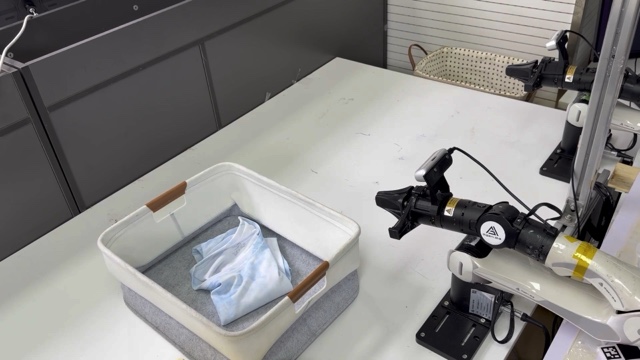}
        \\[1pt]

        \includegraphics[width=\linewidth]
        {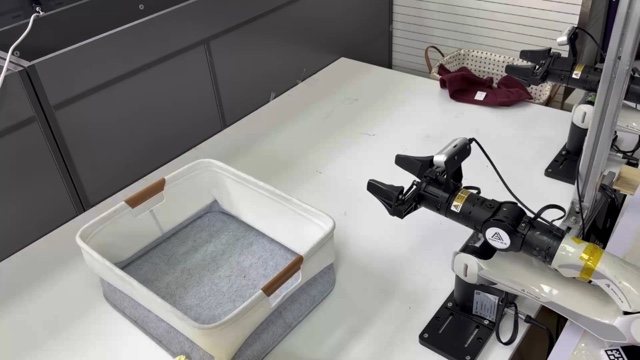}
        &
        \includegraphics[width=\linewidth]
        {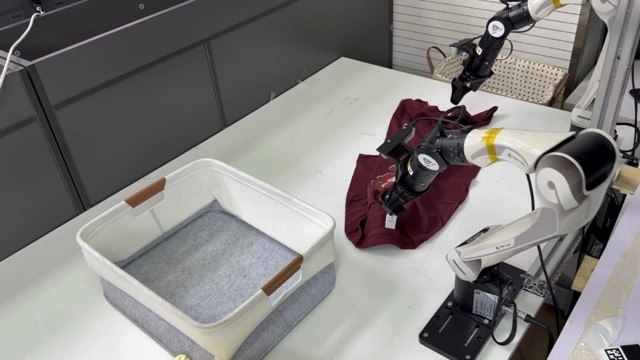}
        &
        \includegraphics[width=\linewidth]
        {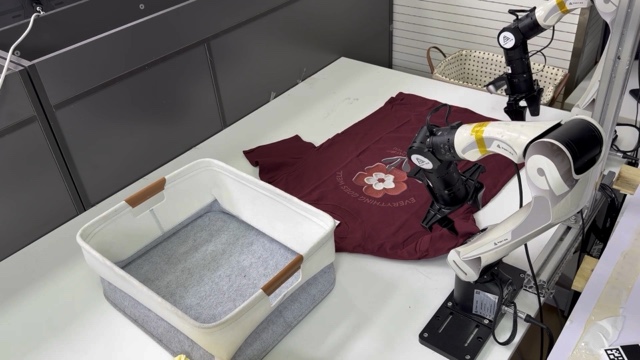}
        &
        \includegraphics[width=\linewidth]
        {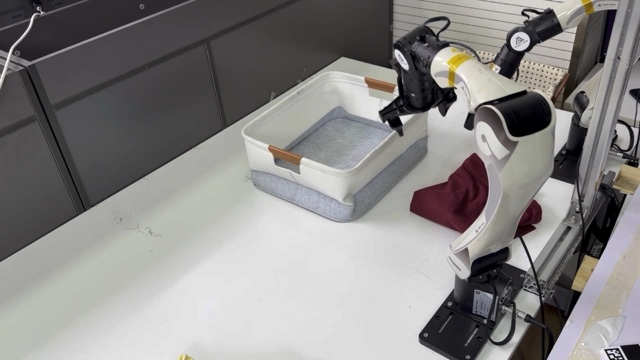}
        &
        \includegraphics[width=\linewidth]
        {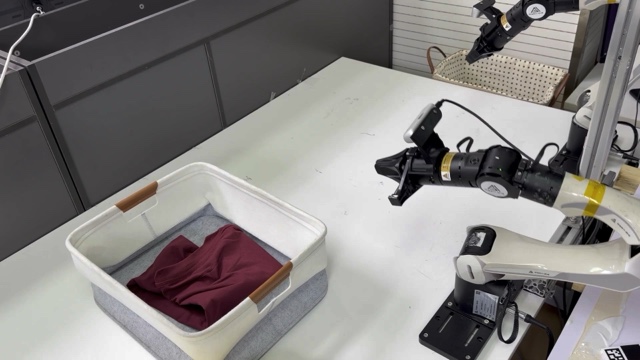}

        \end{tabularx}

        \caption{
        AgileX PiPER: folding unseen garment instances from diverse
unstructured initial states.
        }
        \label{fig:foundation_complex_ood_agilex}
    \end{subfigure}

    \vspace{4pt}

    \begin{subfigure}[t]{\textwidth}
        \centering
        \setlength{\tabcolsep}{1.5pt}

        \begin{tabularx}{\linewidth}{
            @{}
            *{5}{>{\centering\arraybackslash}X}
            @{}
        }

        \includegraphics[width=\linewidth]
        {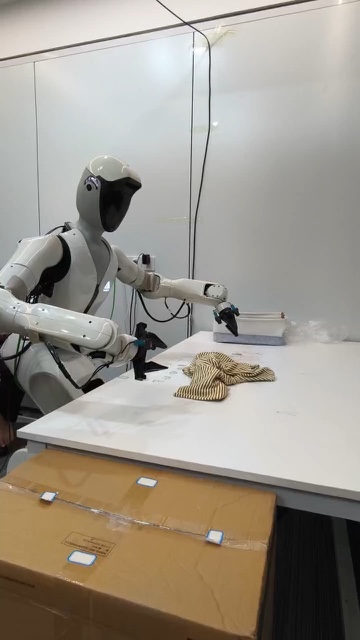}
        &
        \includegraphics[width=\linewidth]
        {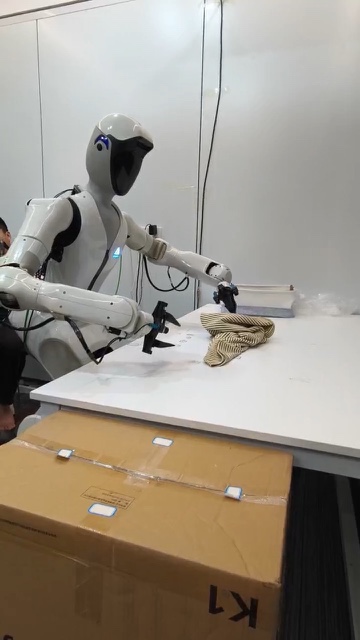}
        &
        \includegraphics[width=\linewidth]
        {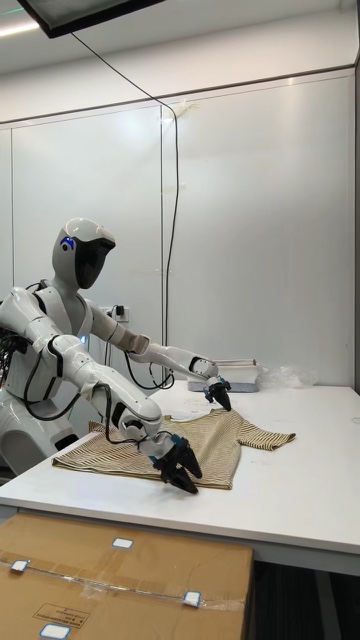}
        &
        \includegraphics[width=\linewidth]
        {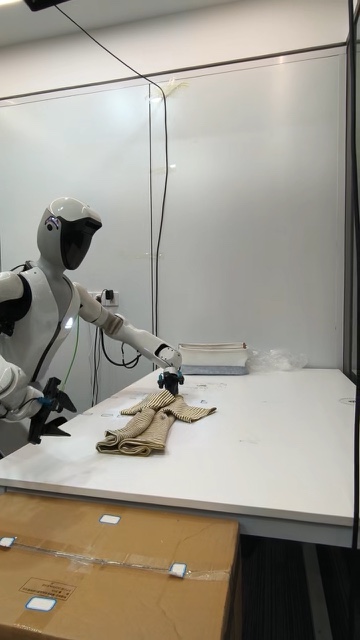}
        &
        \includegraphics[width=\linewidth]
        {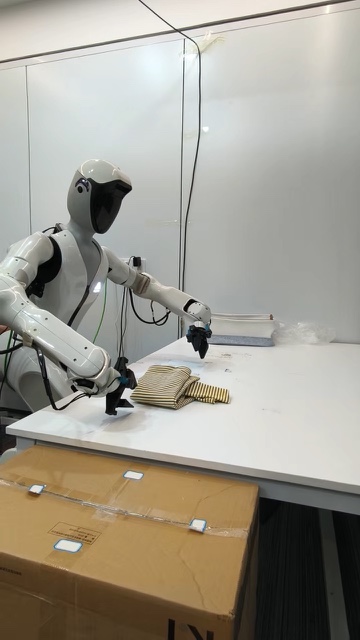}

        \end{tabularx}

        \caption{
        Maker H01: folding an unseen garment instance from an unstructured
initial state.
        }
        \label{fig:foundation_complex_ood_h01}
    \end{subfigure}

    \caption{
    \textbf{Out-of-the-box garment-instance and deformable-state
generalization.}
    The pretrained GigaBrain-0.7 policy is deployed on garment-folding
    tasks on AgileX PiPER and Maker~H01 without task-specific adaptation.
    The evaluated garment instances are absent from the corresponding
    training data and are initialized in diverse, highly unstructured
    configurations.
    Panel~(a) shows two representative AgileX PiPER rollouts with
    different unseen garments and initial states, while Panel~(b) shows
    a Maker~H01 rollout under the same evaluation principle.
    Each row contains five temporally ordered stages from a single
    rollout, illustrating how the policy maintains sustained interaction
    as the deformable garment geometry evolves throughout execution.
    }
    \label{fig:foundation_complex_ood_rollouts}
\end{figure*}

We further evaluate whether the pretrained policy can perform manipulation
tasks that require sustained interaction over multiple control stages.
As shown in Fig.~\ref{fig:foundation_complex}, GigaBrain-0.7 exhibits
non-trivial out-of-the-box performance on both AgileX PiPER and Maker~H01,
including deformable-object manipulation and longer-horizon task execution,
without task-specific adaptation.

Representative out-of-the-box executions are visualized in
Fig.~\ref{fig:foundation_complex_ood_rollouts}.
These evaluations probe generalization along two coupled dimensions:
garment identity and deformable-object state.
The evaluated garments are absent from the corresponding training data
and are presented in diverse, highly unstructured initial
configurations.
This setting differs from conventional appearance-only generalization.
Because cloth geometry changes continuously during manipulation, the
policy cannot rely on a fixed object shape or replay a memorized
trajectory.
Successful execution requires repeatedly establishing appropriate
contact, responding to the evolving garment state, and maintaining
progress across successive action chunks.
Despite these variations, the pretrained policy can reorganize the
garment and sustain the folding behavior across both AgileX PiPER
and Maker~H01 without task-specific adaptation.
These results suggest that heterogeneous pretraining transfers
reusable interaction structure across unseen garment instances and
substantial variation in deformable-object configuration, rather
than only memorizing specific object appearances or initial states.

Out-of-distribution complex manipulation remains substantially more
challenging, particularly when changes in the task configuration require
interaction patterns that differ from those more strongly represented in
pretraining.
Nevertheless, the policy retains executable behavior on multiple OOD
settings, suggesting that interaction patterns acquired from heterogeneous
experience can be reused and recombined as task conditions change.

\textbf{Taken together with the language-following results, these experiments show
that large-scale pretraining provides more than task-specific motor
patterns: it yields a policy that can execute diverse instructions,
sustain longer-horizon interaction, and retain useful behavior beyond
familiar task configurations.
We view this progression as one manifestation of the emergent embodied
capabilities enabled by scaling heterogeneous pretraining.}

\subsection{Post-Training Evaluation}
\label{sec:posttraining_eval}

The pretrained policy provides a generalist starting point, while practical
deployment often requires specialization to a particular embodiment or task
distribution.
We therefore evaluate GigaBrain-0.7 after task-specific post-training on
two complementary settings:
multi-task language following, which emphasizes instruction grounding and
target discrimination, and complex manipulation, which requires sustained
physical interaction and multi-stage execution.

\subsubsection{Multi-Task Language Following}
\label{sec:posttrain_language}

\begin{figure*}[htpb]
    \centering
    \captionsetup{
        type=figure,
        justification=justified,
        singlelinecheck=false
    }
    \begin{subfigure}[t]{0.8\textwidth}
        \centering
        \includegraphics[width=\linewidth]
        {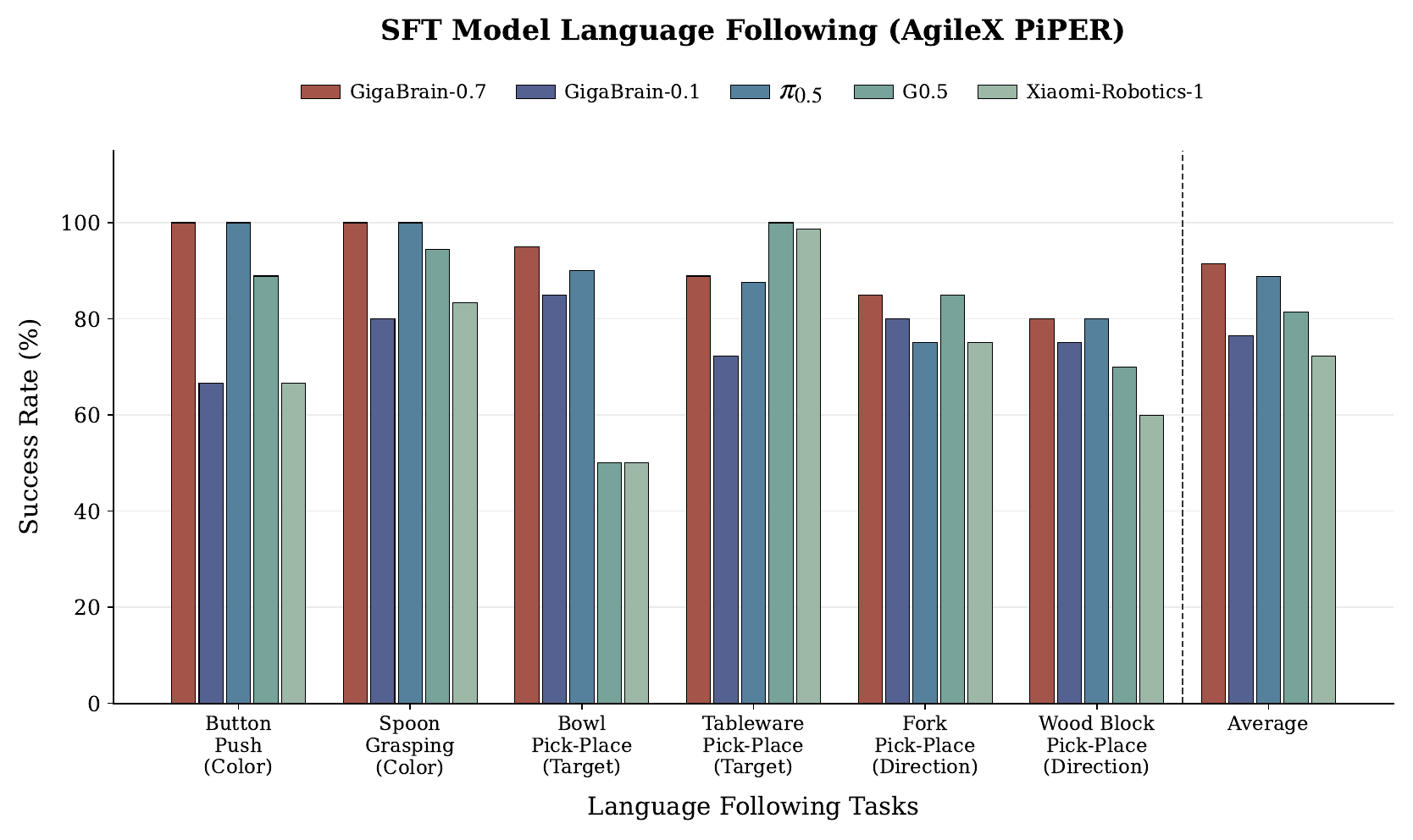}
        \label{fig:posttrain_language_PiPER}
    \end{subfigure}
    \hfill
    \begin{subfigure}[t]{0.8\textwidth}
        \centering
        \includegraphics[width=\linewidth]
        {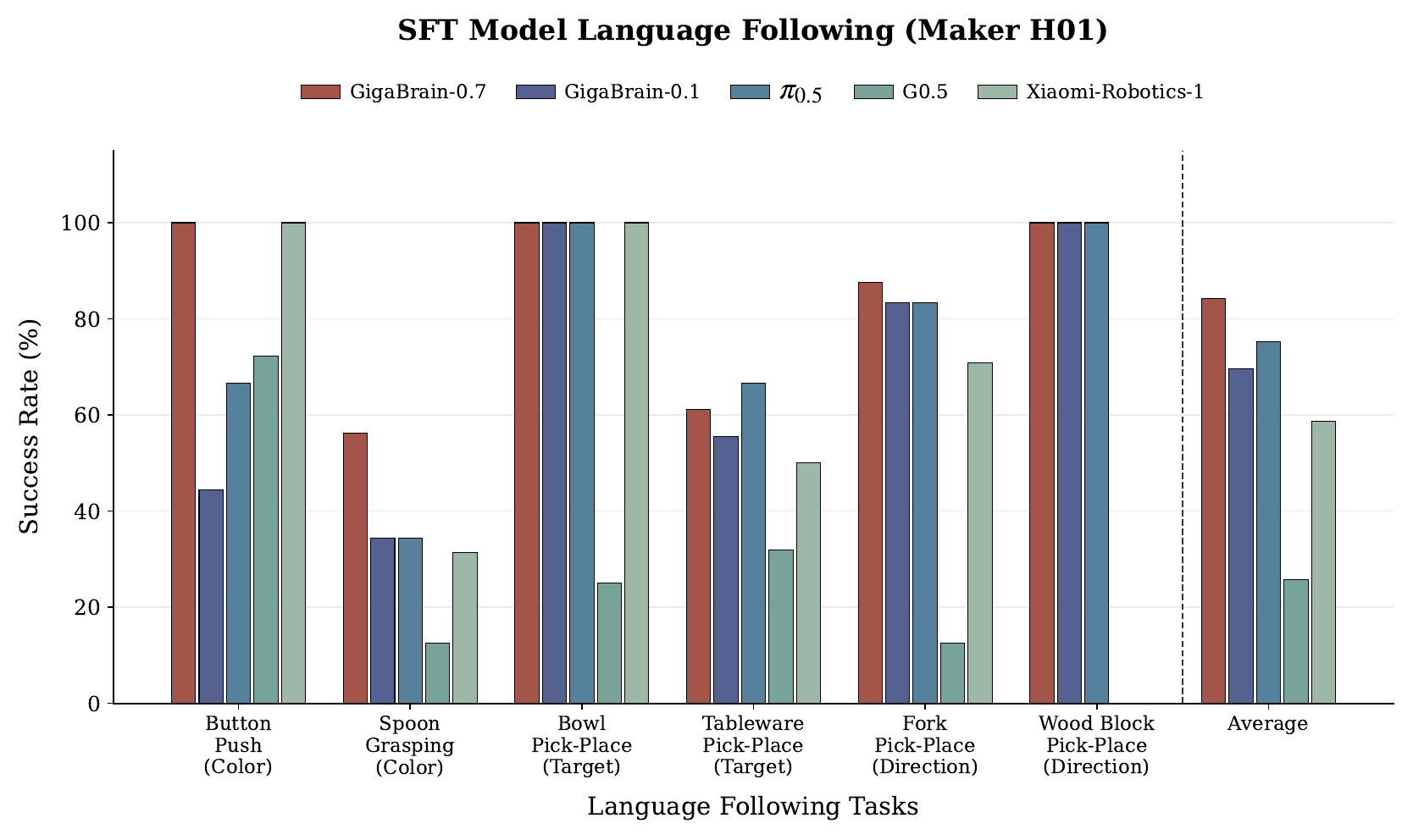}
        \label{fig:posttrain_language_h01}
    \end{subfigure}

    \caption{
    \textbf{Post-training multi-task language following.}
    We compare GigaBrain-0.7 with representative VLA baselines on six
    manipulation tasks that probe color, target, and directional grounding
    across two robot embodiments.
    }
    \label{fig:posttrain_language}
\end{figure*}

We first evaluate six language-conditioned manipulation tasks covering
color discrimination, target selection, and directional reasoning.
Fig.~\ref{fig:posttrain_language} provides an overview of performance on
AgileX PiPER and Maker~H01, while Tab.~\ref{tab:posttrain_language}
reports the complete quantitative results.


\begin{table*}[t]
\centering
\caption{
\textbf{Post-training multi-task language-following evaluation.}
We evaluate six instruction-conditioned manipulation tasks on
AgileX PiPER and Maker H01, covering three complementary
language-following capabilities:
\emph{color discrimination} (button push and spoon grasping),
\emph{target discrimination} (bowl and tableware pick-and-place),
and \emph{direction discrimination} (fork and wood-block pick-and-place).
Each cell reports success rate (\%).
Avg.\ denotes the mean across all six tasks and is reported only for
models with complete evaluations.
Best available results in each column are shown in \textbf{bold}.
}
\label{tab:posttrain_language}

\scriptsize
\setlength{\tabcolsep}{3.5pt}
\renewcommand{\arraystretch}{1.08}


\begin{tabular}{lccccccc}
\toprule

& \multicolumn{7}{c}{\textbf{AgileX PiPER}} \\
\cmidrule(lr){2-8}

\textbf{Model}
& \multicolumn{2}{c}{\textbf{Color}}
& \multicolumn{2}{c}{\textbf{Target}}
& \multicolumn{2}{c}{\textbf{Direction}}
& \textbf{Avg.} \\

\cmidrule(lr){2-3}
\cmidrule(lr){4-5}
\cmidrule(lr){6-7}

&
\makecell{Button\\Push}
&
\makecell{Spoon\\Grasping}
&
\makecell{Bowl\\Pick-Place}
&
\makecell{Tableware\\Pick-Place}
&
\makecell{Fork\\Pick-Place}
&
\makecell{Wood Block\\Pick-Place}
&
\\
\midrule

$\pi_{0.5}$
& \textbf{100.0}
& \textbf{100.0}
& 90.0
& 87.5
& 75.0
& \textbf{80.0}
& 88.8 \\

GigaBrain-0.1
& 66.7
& 77.8
& 85.0
& 72.2
& 80.0
& 75.0
& 76.1 \\

G0.5
& 88.9
& 94.4
& 50.0
& \textbf{100.0}
& \textbf{85.0}
& 70.0
& 81.4 \\

Xiaomi-Robotics-1
& 66.7
& 83.3
& 50.0
& 98.6
& 75.0
& 60.0
& 72.3 \\

GigaBrain-0.7
& \textbf{100.0}
& \textbf{100.0}
& \textbf{95.0}
& 88.9
& \textbf{85.0}
& \textbf{80.0}
& \textbf{91.5} \\

\bottomrule
\end{tabular}

\vspace{7pt}


\begin{tabular}{lccccccc}
\toprule

& \multicolumn{7}{c}{\textbf{Maker H01}} \\
\cmidrule(lr){2-8}

\textbf{Model}
& \multicolumn{2}{c}{\textbf{Color}}
& \multicolumn{2}{c}{\textbf{Target}}
& \multicolumn{2}{c}{\textbf{Direction}}
& \textbf{Avg.} \\

\cmidrule(lr){2-3}
\cmidrule(lr){4-5}
\cmidrule(lr){6-7}

&
\makecell{Button\\Push}
&
\makecell{Spoon\\Grasping}
&
\makecell{Bowl\\Pick-Place}
&
\makecell{Tableware\\Pick-Place}
&
\makecell{Fork\\Pick-Place}
&
\makecell{Wood Block\\Pick-Place}
&
\\
\midrule

$\pi_{0.5}$
& 66.7
& 34.4
& \textbf{100.0}
& \textbf{66.7}
& 83.3
& \textbf{100.0}
& 75.2 \\

GigaBrain-0.1
& 44.4
& 34.4
& \textbf{100.0}
& 55.6
& 83.3
& \textbf{100.0}
& 69.6 \\

G0.5
& 72.2
& 12.5
& 25.0
& 31.9
& 12.5
& 0.0
& 25.7 \\

Xiaomi-Robotics-1
& \textbf{100.0}
& 31.3
& \textbf{100.0}
& 50.0
& 70.8
& 0.0
& 58.7 \\

GigaBrain-0.7
& \textbf{100.0}
& \textbf{56.3}
& \textbf{100.0}
& 61.1
& \textbf{87.5}
& \textbf{100.0}
& \textbf{84.2} \\

\bottomrule
\end{tabular}

\end{table*}

Across the PiPER evaluation, GigaBrain-0.7 consistently improves over the
preceding GigaBrain model and matches or exceeds $\pi_{0.5}$ across the
reported tasks.
The gains are particularly visible when language must be grounded into
target-specific or directional physical behavior, indicating that the
pretrained vision-language-action representation remains effective after
specialization.

\begin{figure*}[htbp]
    \centering
    \captionsetup{
        type=figure,
        justification=justified,
        singlelinecheck=false
    }
    \includegraphics[
        width=0.99\textwidth
    ]{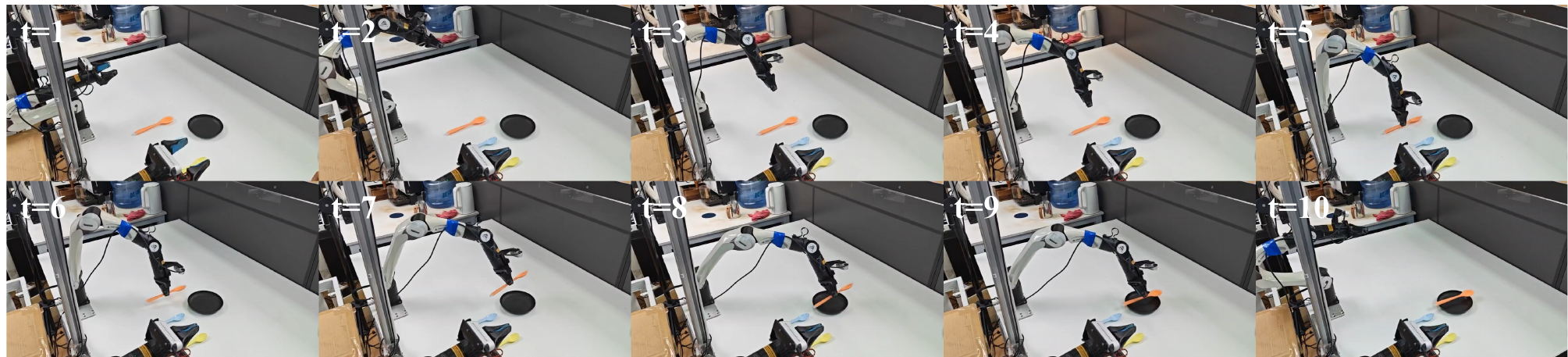}

    \caption{
    \textbf{Post-training language following on AgileX PiPER: color grounding.}
    Given the instruction \emph{``Grab the orange spoon and put it on the
    plate.''}, GigaBrain-0.7 identifies the language-specified object by
    color and executes the corresponding pick-and-place behavior.
    The ten frames show temporally ordered stages of a representative rollout.
    }
    \label{fig:posttrain_language_spoon}
\end{figure*}

\begin{figure*}[tbp]
    \centering
    \captionsetup{
        type=figure,
        justification=justified,
        singlelinecheck=false
    }
    \includegraphics[
        width=0.99\textwidth
    ]{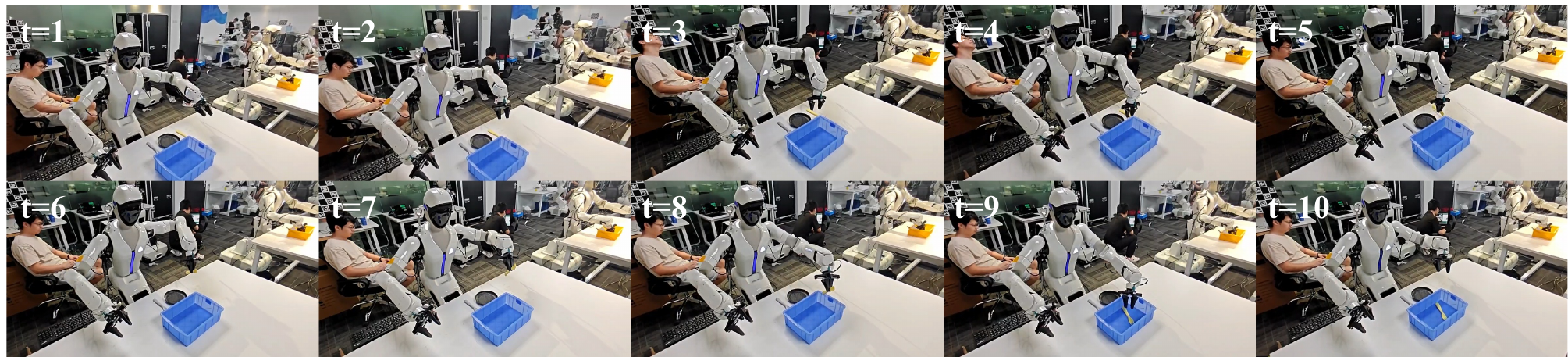}

    \caption{
    \textbf{Post-training language following on Maker H01: spatial grounding.}
    Given the instruction \emph{``Grab the fork on the left side of the
    plate and put it in the basket.''}, the policy grounds the relative
    spatial relation before executing the instructed pick-and-place sequence.
    }
    \label{fig:posttrain_language_fork_left}
\end{figure*}

The comparison on Maker~H01 is more challenging but exhibits the same
overall trend.
GigaBrain-0.7 improves the average success rate from 69.6\% for
GigaBrain-0.1 to 84.2\%, with particularly large gains on button push
and spoon grasping while maintaining strong performance on target- and
direction-conditioned tasks.
Together, the two embodiments indicate that the post-training gains are not
restricted to a single robot configuration, although performance remains
sensitive to the underlying embodiment and task.
Beyond aggregate success rates, we further visualize representative
post-training rollouts to examine whether the learned policies remain
responsive to fine-grained language conditions.
Figures~\ref{fig:posttrain_language_spoon}--%
\ref{fig:posttrain_language_fork_left} cover three complementary forms
of instruction grounding---color attributes, object identity, and
relative spatial relations---across AgileX PiPER and Maker~H01.

\begin{figure*}[htbp]
    \centering
    \captionsetup{
        type=figure,
        justification=justified,
        singlelinecheck=false
    }
    \includegraphics[
        width=0.99\textwidth
    ]{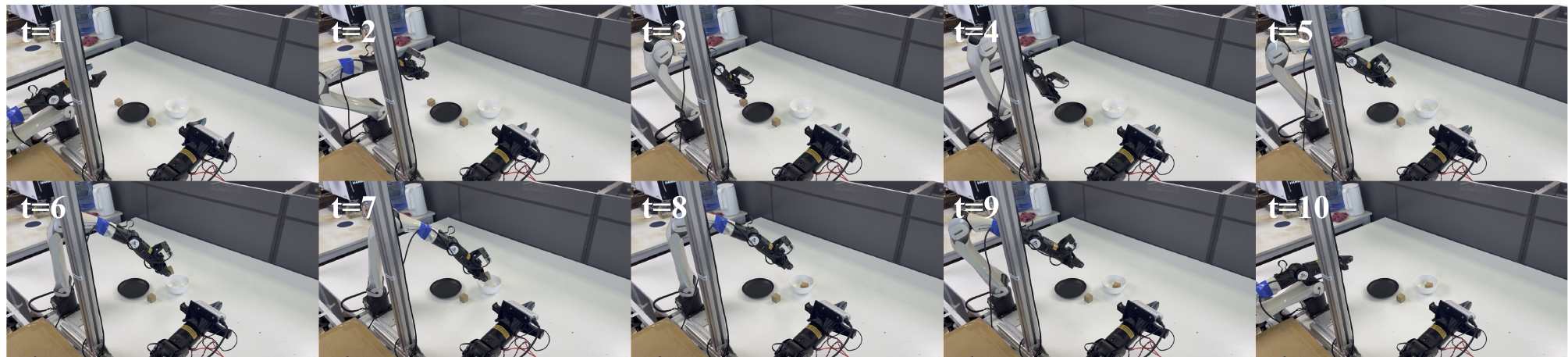}

    \caption{
    \textbf{Post-training language following on AgileX PiPER: spatial grounding.}
    Given the instruction \emph{``Grab the block on the left side of the
    plate and put it in the bowl.''}, the policy resolves the relative
    spatial reference, selects the corresponding block, and executes the
    instructed transfer.
    The ten frames show temporally ordered stages of a representative rollout.
    }
    \label{fig:posttrain_language_block}
\end{figure*}

\begin{figure*}[htbp]
    \centering
    \captionsetup{
        type=figure,
        justification=justified,
        singlelinecheck=false
    }
    \includegraphics[
        width=0.99\textwidth
    ]{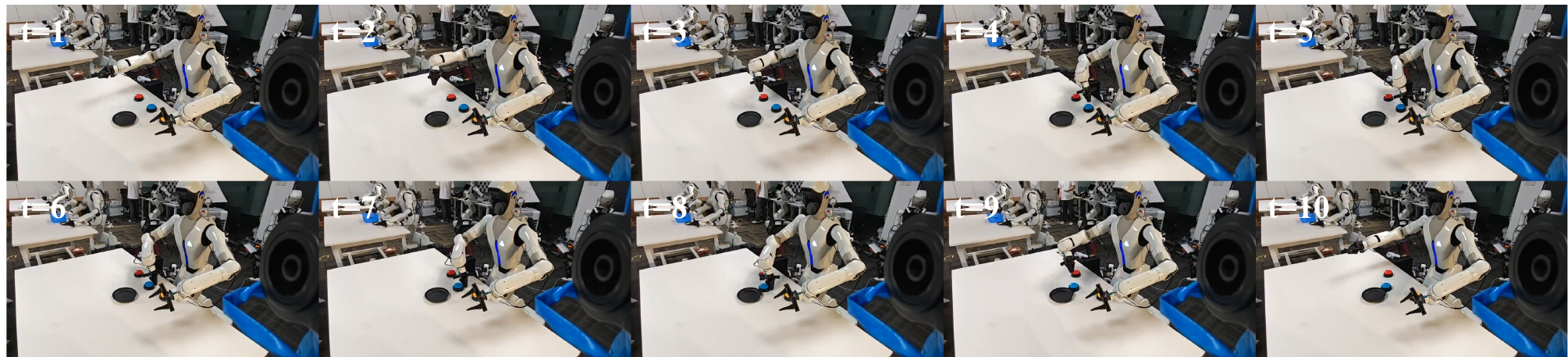}

    \caption{
    \textbf{Post-training language following on Maker H01: color grounding.}
    In a scene containing multiple colored alternatives, GigaBrain-0.7
    identifies the language-specified blue target and executes the
    corresponding manipulation behavior.
    The rollout illustrates sensitivity to fine-grained visual attributes
    on the humanoid embodiment.
    }
    \label{fig:posttrain_language_button}
\end{figure*}

\begin{figure*}[htbp]
    \centering
    \captionsetup{
        type=figure,
        justification=justified,
        singlelinecheck=false
    }
    \includegraphics[
        width=0.99\textwidth
    ]{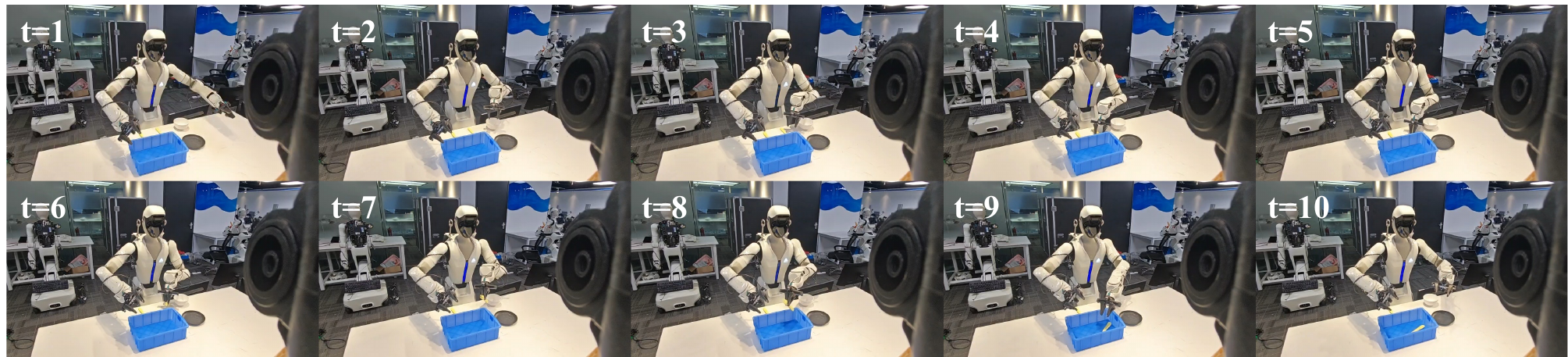}

    \caption{
    \textbf{Post-training language following on Maker H01: target grounding.}
    Given the instruction \emph{``Pick up the fork and move it to the
    basket.''}, the policy identifies the requested utensil and maintains
    the language-conditioned target throughout the subsequent manipulation
    sequence.
    }
    \label{fig:posttrain_language_fork}
\end{figure*}

Across these examples, the post-trained policy conditions its behavior
on distinctions expressed in the instruction rather than reproducing a
fixed manipulation trajectory.
Color attributes, object identity, and relative spatial relations all
affect target selection and subsequent execution, and the same qualitative
behavior is observed across both AgileX PiPER and Maker~H01.
These rollouts demonstrate that post-training preserves fine-grained instruction grounding while adapting the policy to heterogeneous robot embodiments.

\subsubsection{Complex Manipulation}
\label{sec:posttrain_complex}

\begin{figure*}[htpb]
    \centering
    \captionsetup{
        type=figure,
        justification=justified,
        singlelinecheck=false
    }
    \begin{subfigure}[t]{0.8\textwidth}
        \centering
        \includegraphics[width=\linewidth]
        {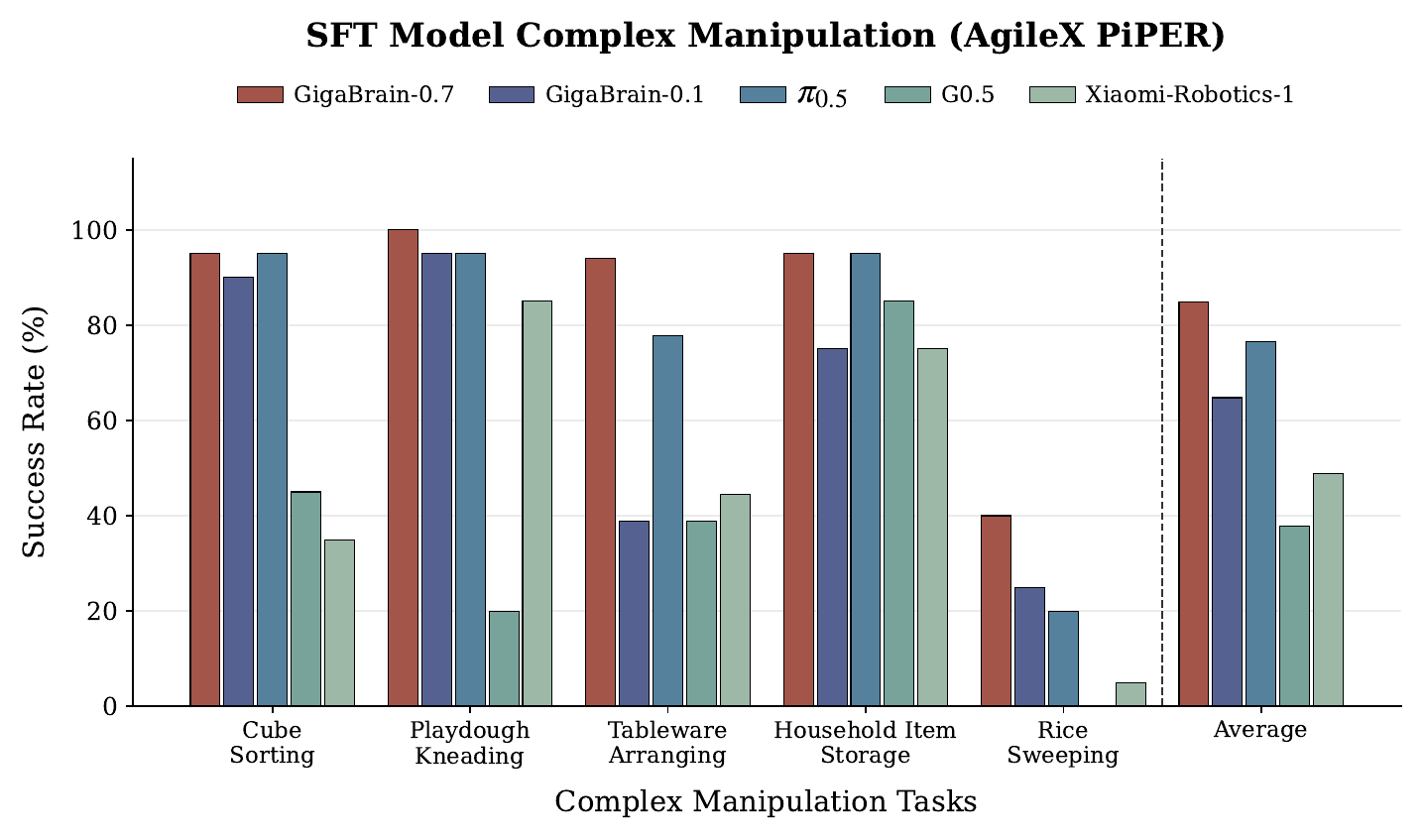}
        \label{fig:posttrain_complex_PiPER}
    \end{subfigure}
    \hfill
    \begin{subfigure}[t]{0.8\textwidth}
        \centering
        \includegraphics[width=\linewidth]
        {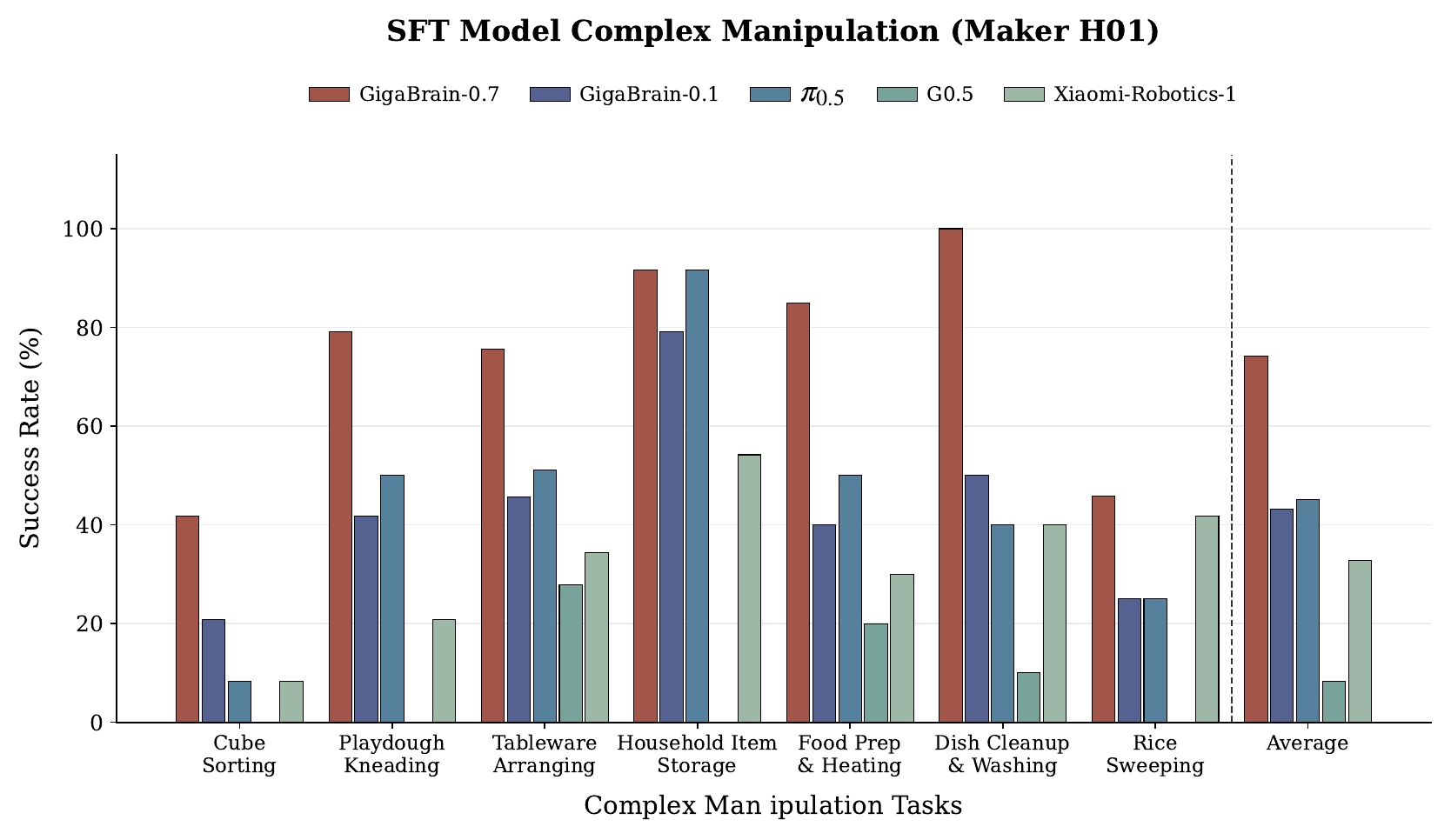}
        \label{fig:posttrain_complex_h01}
    \end{subfigure}

    \caption{
    \textbf{Post-training complex manipulation.}
    The evaluation covers manipulation tasks requiring sustained interaction,
    multi-stage sequencing, deformable-object handling, and tool use on
    AgileX PiPER and Maker~H01.
    }
    \label{fig:posttrain_complex}
\end{figure*}

We next evaluate tasks requiring sustained contact, multi-stage execution,
deformable-object manipulation, or tool-mediated interaction.
Fig.~\ref{fig:posttrain_complex} visualizes the comparison across
platforms, and Tab.~\ref{tab:posttrain_complex} provides the exact results.

\begin{table*}[t]
\centering
\caption{
\textbf{Post-training single-task complex manipulation evaluation.}
We evaluate post-trained policies on a diverse set of complex manipulation
tasks across AgileX PiPER and Maker H01.
The benchmark covers object sorting, deformable-object manipulation,
tableware arrangement, household-item storage, sweeping, food preparation
and heating, and dish cleanup.
Each cell reports task success rate (\%).
Avg.\ denotes the mean over all evaluated tasks on the corresponding
embodiment and is reported only for models with complete evaluations.
Best available results in each column are shown in \textbf{bold}.
}
\label{tab:posttrain_complex}

\scriptsize
\setlength{\tabcolsep}{3.5pt}
\renewcommand{\arraystretch}{1.08}


\begin{tabular}{lcccccc}
\toprule

& \multicolumn{6}{c}{\textbf{AgileX PiPER}} \\
\cmidrule(lr){2-7}

\textbf{Model}
& \makecell{Cube\\Sorting}
& \makecell{Play-Dough\\Kneading}
& \makecell{Tableware\\Arranging}
& \makecell{Household Item\\Storage}
& \makecell{Rice\\Sweeping}
& \textbf{Avg.} \\
\midrule

$\pi_{0.5}$
& \textbf{95.0}
& 95.0
& 77.8
& \textbf{95.0}
& 20.0
& 76.6 \\

GigaBrain-0.1
& 90.0
& 95.0
& 38.9
& 75.0
& 25.0
& 64.8 \\

G0.5
& 45.0
& 20.0
& 38.9
& 85.0
& 0.0
& 37.8 \\

Xiaomi-Robotics-1
& 35.0
& 85.0
& 44.4
& 75.0
& 5.0
& 48.9 \\

GigaBrain-0.7
& \textbf{95.0}
& \textbf{100.0}
& \textbf{94.4}
& \textbf{95.0}
& \textbf{40.0}
& \textbf{84.9} \\

\bottomrule
\end{tabular}

\vspace{7pt}


\begin{tabular}{lcccccccc}
\toprule

& \multicolumn{8}{c}{\textbf{Maker H01}} \\
\cmidrule(lr){2-9}

\textbf{Model}
& \makecell{Cube\\Sorting}
& \makecell{Play-Dough\\Kneading}
& \makecell{Tableware\\Arranging}
& \makecell{Household Item\\Storage}
& \makecell{Food Prep.\\\& Heating}
& \makecell{Dish Cleanup\\\& Washing}
& \makecell{Rice\\Sweeping}
& \textbf{Avg.} \\
\midrule

$\pi_{0.5}$
& 8.3
& 50.0
& 51.1
& \textbf{91.7}
& 50.0
& 40.0
& 25.0
& 45.2 \\

GigaBrain-0.1
& 20.8
& 41.7
& 45.6
& 79.2
& 40.0
& 50.0
& 25.0
& 43.2 \\

G0.5
& 0.0
& 0.0
& 27.8
& 0.0
& 20.0
& 10.0
& 0.0
& 8.3 \\

Xiaomi-Robotics-1
& 8.3
& 20.8
& 34.4
& 54.2
& 30.0
& 40.0
& 41.7
& 32.8 \\

GigaBrain-0.7
& \textbf{41.7}
& \textbf{79.2}
& \textbf{75.6}
& \textbf{91.7}
& \textbf{85.0}
& \textbf{100.0}
& \textbf{45.8}
& \textbf{74.1} \\

\bottomrule
\end{tabular}

\end{table*}

Across both embodiments, GigaBrain-0.7 provides consistent improvements over
the preceding GigaBrain model and remains competitive with or stronger than
$\pi_{0.5}$ across the completed complex-manipulation evaluations.

\begin{figure*}[tbp]
    \centering
    \captionsetup{
        type=figure,
        justification=justified,
        singlelinecheck=false
    }
    \includegraphics[
        width=0.99\textwidth
    ]{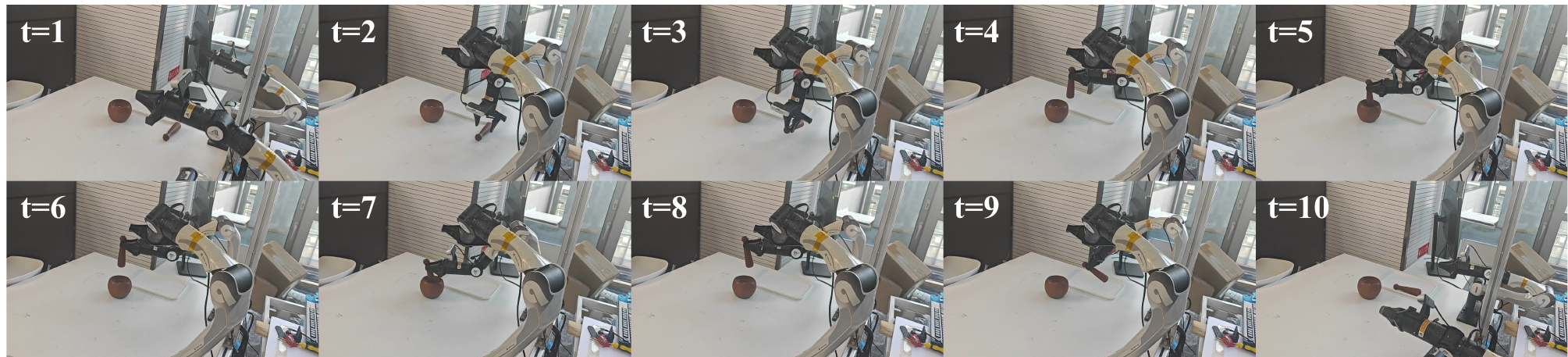}

    \caption{
    \textbf{Post-training complex manipulation on AgileX PiPER:
    play-dough manipulation.}
    The policy performs repeated contact-rich interactions with deformable
    material, requiring subsequent actions to adapt as the object geometry
    changes throughout execution.
    The ten frames show temporally ordered stages of a representative rollout.
    }
    \label{fig:posttrain_complex_playdough}
\end{figure*}

\begin{figure*}[tbp]
    \centering
    \captionsetup{
        type=figure,
        justification=justified,
        singlelinecheck=false
    }
    \includegraphics[
        width=0.99\textwidth
    ]{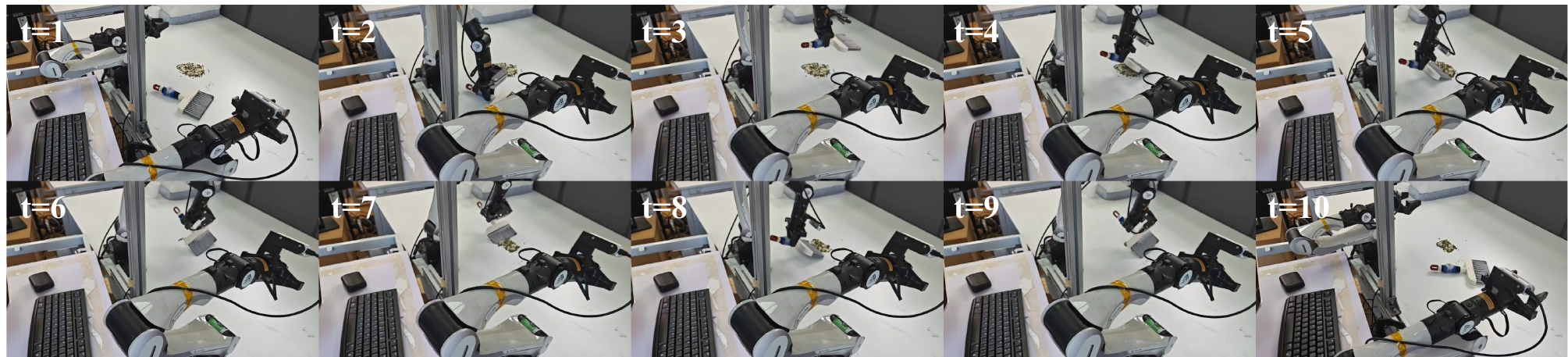}

    \caption{
    \textbf{Post-training complex manipulation on AgileX PiPER:
    tabletop sweeping.}
    GigaBrain-0.7 uses a brush to manipulate distributed material across
    the tabletop, requiring sustained tool--object contact and coordinated
    motion over multiple stages of execution.
    }
    \label{fig:posttrain_complex_brush}
\end{figure*}

\begin{figure*}[htbp]
    \centering
    \captionsetup{
        type=figure,
        justification=justified,
        singlelinecheck=false
    }
    \includegraphics[
        width=0.99\textwidth
    ]{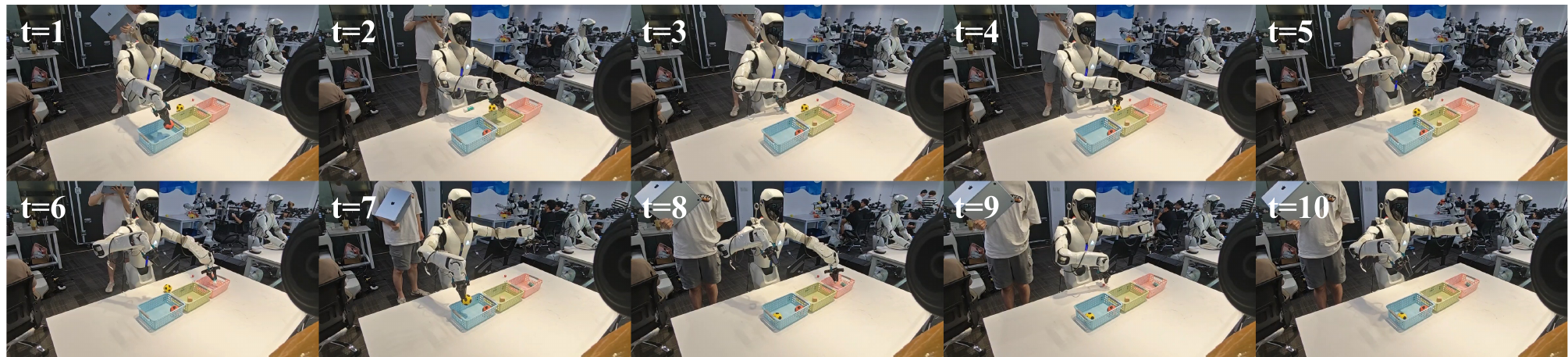}

    \caption{
    \textbf{Post-training complex manipulation on Maker H01:
    household-item organization.}
    The humanoid policy performs a sequence of object selection, grasping,
    transport, and placement behaviors to organize multiple items across
    the workspace.
    The rollout illustrates sustained multi-stage execution on a
    substantially different robot embodiment.
    }
    \label{fig:posttrain_complex_organize}
\end{figure*}

\begin{figure*}[htbp]
    \centering
    \captionsetup{
        type=figure,
        justification=justified,
        singlelinecheck=false
    }
    \includegraphics[
        width=0.99\textwidth
    ]{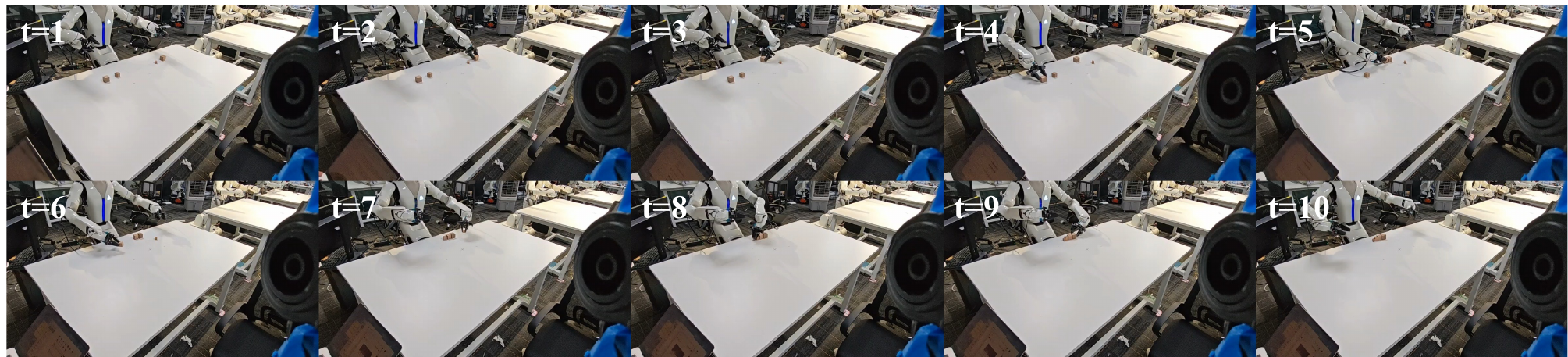}

    \caption{
    \textbf{Post-training complex manipulation on Maker H01:
    block sorting.}
    GigaBrain-0.7 performs repeated object selection and placement across
    successive manipulation stages, requiring the policy to maintain task
    progress as the scene changes after each interaction.
    }
    \label{fig:posttrain_complex_sort_block}
\end{figure*}

On PiPER, differences are relatively modest on several structured tasks
where the strongest baselines already perform well, while clearer gains
appear on interactions requiring more sustained contact or multi-stage
execution.
On Maker~H01, the separation is substantially larger and extends across a
broader range of skills, including deformable-object manipulation,
food-related interaction, and multi-stage cleanup.
We further visualize representative post-training executions on complex
manipulation tasks in
Figures~\ref{fig:posttrain_complex_playdough}--%
\ref{fig:posttrain_complex_sort_block}.
The selected tasks span deformable-object interaction, tool-mediated
manipulation, household-item organization, and sequential object sorting,
placing qualitatively different demands on contact, temporal coordination,
and task progress.

The consistency of these gains across task categories suggests that the
benefit of GigaBrain-0.7 is not confined to a single manipulation primitive.
At the same time, several contact-rich tasks remain far from saturation,
indicating that complex physical interaction remains an important direction
for additional data scaling and experience-based post-training.


\subsection{Benchmark Evaluation}

Beyond real-robot evaluation, we further assess GigaBrain-0.7 on
standardized benchmarks that probe complementary capabilities of an
embodied foundation model.
We evaluate embodied vision-language understanding through
MiMo-Embodied~\citep{mimo}, and policy execution through three complementary
simulation benchmarks: RoboTwin~2.0~\citep{robotwin2},
EBench~\citep{ebench}, and RoboColiseum~\citep{robocoliseum}.

\subsubsection{Embodied Vision-Language Evaluation}
\label{sec:vlm_eval}

Beyond action generation, an embodied foundation model should retain the
visual-semantic and spatial reasoning capabilities required to interpret
physical scenes and support downstream interaction.
We therefore evaluate the vision-language component of GigaBrain-0.7 on
MiMo-Embodied~\citep{mimo}, a suite of embodied vision-language
benchmarks covering spatial understanding and affordance reasoning.
We exclude VSI-Bench because it requires video input, whereas the VLM
interface evaluated here operates on image observations.

Following the benchmark taxonomy, we organize the evaluated tasks into two
capability groups.
\emph{Spatial Understanding} contains eight benchmarks covering spatial
reasoning, visual grounding, referring expressions, and scene-level
understanding.
\emph{Affordance} contains five benchmarks covering interaction-oriented
referring, placement reasoning, point-level affordance prediction,
part-level affordance understanding, and robotic affordance reasoning.
For a fair comparison with existing models, Tab.~\ref{tab:mimo_summary}
reports GigaBrain-0.7 before MiMo data are introduced into the subsequent
pretraining mixture.

\begin{table}[t]
\centering
\caption{
\textbf{Summary of embodied vision-language evaluation on
MiMo-Embodied~\citep{mimo}.}
Spatial Avg.\ is averaged over eight Spatial Understanding benchmarks,
and Affordance Avg.\ over five Affordance benchmarks.
Overall denotes the mean over all 13 evaluated benchmarks.
For GigaBrain-0.7, the reported comparison is obtained before MiMo data
are incorporated into subsequent continued pretraining.
\textbf{Bold} indicates the best result among the compared models.
}
\label{tab:mimo_summary}

\small
\setlength{\tabcolsep}{5pt}
\renewcommand{\arraystretch}{1.05}

\begin{tabular}{lccc}
\toprule
\textbf{Model}
& \textbf{Spatial Avg.}
& \textbf{Affordance Avg.}
& \textbf{Overall} \\
\midrule

Xiaomi-Robotics-0~\citep{xiaomi_robotics_0}
& .1372
& .0687
& .1108 \\

Spirit-v1.5~\citep{spiritv15}
& .3524
& .2995
& .3320 \\

Wall-OSS-0.5~\citep{walloss05}
& .2273
& .0367
& .1540 \\

G0.5-base~\citep{galaxea_g05}
& .3890
& \textbf{.3956}
& .3916 \\

Hy-Embodied-0.5-VLA-UMI~\citep{hyvla05}
& .2532
& .0000
& .1558 \\


GigaBrain-0.7
& \textbf{.5215}
& .3669
& \textbf{.4621} \\

\bottomrule
\end{tabular}
\end{table}



As shown in Tab.~\ref{tab:mimo_summary}, GigaBrain-0.7 already exhibits
strong embodied vision-language capability before MiMo data are introduced
into the training mixture, achieving the strongest overall and Spatial
Understanding averages among the compared models.

We subsequently incorporate MiMo data into the same one-stage VLA
pretraining mixture and continue joint optimization of vision-language and
action objectives, without introducing a separate VLM adaptation stage.
The resulting final checkpoint further reaches an overall MiMo score of
0.5704, indicating that additional embodied vision-language supervision
can be absorbed directly within the unified VLA pretraining framework.

Since the later training mixture is no longer disjoint from the MiMo
evaluation set, we treat this final score as a continued-pretraining
diagnostic rather than a held-out benchmark result.
The released GigaBrain-0.7 checkpoint corresponds to this final
MiMo-enhanced pretraining stage.

\subsubsection{Simulation Benchmark Evaluation}
\label{sec:simulation_benchmark}

We further evaluate GigaBrain-0.7 on three simulation benchmarks with
complementary task distributions and evaluation settings:
RoboTwin~2.0~\citep{robotwin2},
EBench~\citep{ebench}, and
RoboColiseum~\citep{robocoliseum}.
Together, these benchmarks cover tabletop bimanual manipulation,
mobile and long-horizon interaction, and reconstruction-based
Real2Sim2Real evaluation.

For readability and reproducibility, the main text focuses on
representative publicly available methods, while complete leaderboard
snapshots, including additional submissions are provided in Appendix~\ref{app:sim_leaderboards} together with their source URLs and snapshot dates.

\paragraph{Physics-based simulation benchmarks.}

We first evaluate on RoboTwin~2.0~\citep{robotwin2}, a large-scale
benchmark for bimanual manipulation under both clean and randomized
simulation conditions.
Following the official \emph{Co-Train} protocol, a single GigaBrain-0.7
policy is post-trained jointly on all 50 benchmark tasks using 50 clean
demonstrations per task, and is subsequently evaluated under both the
clean \emph{Easy} setting and the domain-randomized \emph{Hard} setting.
This differs from the \emph{Single} protocol, where a separate checkpoint
is fine-tuned for each individual task.

As shown in Tab.~\ref{tab:robotwin2}, GigaBrain-0.7 achieves the
strongest overall performance among the evaluated methods and ranks first
under the challenging Hard setting.
While several models obtain higher success in the clean environment,
GigaBrain-0.7 retains substantially stronger performance after domain
randomization.
This result highlights the robustness of the learned policy to visual and
environmental variations while operating as a single multi-task policy.

\begin{table}[t]
\centering
\caption{
\textbf{Evaluation on RoboTwin~2.0~\citep{robotwin2}.}
Under the official Co-Train protocol, one policy is jointly post-trained
on all 50 tasks using 50 clean demonstrations per task.
Single denotes task-specific fine-tuning with a separate checkpoint for
each task.
Success rates (\%) are reported under the Easy (clean) and
Hard (domain-randomized) settings.
Overall denotes the mean of Easy and Hard.
Best results in each column are shown in \textbf{bold}.
}
\label{tab:robotwin2}

\small
\setlength{\tabcolsep}{5.5pt}
\renewcommand{\arraystretch}{1.05}

\begin{tabular}{lcccc}
\toprule
\textbf{Model}
& \textbf{Method}
& \textbf{Easy}
& \textbf{Hard}
& \textbf{Overall} \\
\midrule

$\pi_0$~\citep{pi0}
& Single
& 46.42
& 16.34
& 31.38 \\

Xiaomi-Robotics-0~\citep{xiaomi_robotics_0}
& Co-Train
& 62.9
& 18.2
& 40.55 \\

X-VLA~\citep{x-vla}
& Co-Train
& 68.0
& 20.9
& 44.45 \\

X-WAM~\citep{x-wam}
& Co-Train
& 70.0
& 25.8
& 47.90 \\

$\pi_{0.5}$~\citep{pi05}
& Co-Train
& \textbf{70.7}
& 46.0
& 58.35 \\

GigaBrain-0.7
& Co-Train
& 66.8
& \textbf{67.9}
& \textbf{67.35} \\

\bottomrule
\end{tabular}
\end{table}

We next evaluate on EBench~\citep{ebench}, which extends simulation
evaluation to mobile manipulation, long-horizon execution, and
dexterous and precise interaction.
Among the publicly available VLA models compared in
Tab.~\ref{tab:ebench}, GigaBrain-0.7 achieves the strongest performance
in both overall success rate and aggregate task score, outperforming
representative generalist VLA baselines including $\pi_0$ and
$\pi_{0.5}$.
This result complements RoboTwin~2.0 by showing that the model's advantage
extends beyond fixed-base bimanual manipulation to tasks involving
mobility, extended execution horizons, and precise interaction.
The complete official EBench leaderboard, including additional submissions
not included in the main-text comparison, is provided in
Appendix~\ref{app:ebench_leaderboard}.

\begin{table}[t]
\centering
\caption{
\textbf{Evaluation on EBench~\citep{ebench} among publicly available VLA models.}
SR denotes the overall task success rate, and Score measures task progress.
Results for $\pi_0$, $\pi_{0.5}$, X-VLA, and GigaBrain-0.7 follow the
corresponding leaderboard evaluations.
\textsuperscript{*}The InternVLA-A1 result is reproduced from the EBench
evaluation reported in Qwen-RobotManip~\citep{qwen_robotmanip}, rather
than from the current EBench leaderboard.
Best results among the compared models are shown in \textbf{bold}.
The complete official leaderboard, including additional submissions not
shown here, is provided in Appendix~\ref{app:ebench_leaderboard}.
}
\label{tab:ebench}

\small
\setlength{\tabcolsep}{7pt}
\renewcommand{\arraystretch}{1.05}

\begin{tabular}{lcc}
\toprule
\textbf{Model}
& \textbf{SR (\%)}
& \textbf{Score} \\
\midrule

$\pi_0$~\citep{pi0}
& 23.59
& 37 \\

X-VLA~\citep{x-vla}
& 23.72
& 35 \\

InternVLA-A1\textsuperscript{*}~\citep{internvla}
& 23.90
& 36 \\

$\pi_{0.5}$~\citep{pi05}
& 28.08
& 42 \\

GigaBrain-0.7
& \textbf{33.30}
& \textbf{46.1} \\



\bottomrule
\end{tabular}
\end{table}

\paragraph{Reconstruction-based Real2Sim2Real evaluation.}

Physics-based simulation provides scalable and reproducible evaluation,
but its synthetic visual distribution inevitably introduces a
simulation-to-reality gap.
We therefore complement RoboTwin~2.0 and EBench with
RoboColiseum~\citep{robocoliseum}, which adopts a reconstruction-based
Real2Sim2Real evaluation pipeline to more closely align simulated
observations with real-world environments.

RoboColiseum evaluates policies along four complementary dimensions:
instruction following, spatial reasoning, robustness, and general
manipulation.
As shown in Tab.~\ref{tab:robocoliseum}, GigaBrain-0.7 achieves the
strongest performance across all four dimensions among the evaluated
open-source models.
The advantage is particularly evident in spatial reasoning, while strong
performance is retained across instruction-conditioned execution,
robustness, and general manipulation.
Together with the real-robot evaluations in the preceding sections, these
results provide complementary evidence that the capabilities acquired by
GigaBrain-0.7 remain effective across both simulated and real-world
manipulation settings.

\begin{table*}[t]
\centering
\caption{
\textbf{Evaluation on RoboColiseum~\citep{robocoliseum}.}
The benchmark evaluates four complementary dimensions of robot capability.
We report representative open-source models with complete evaluations.
Best results in each column are shown in \textbf{bold}.
}
\label{tab:robocoliseum}

\small
\setlength{\tabcolsep}{8pt}
\renewcommand{\arraystretch}{1.05}

\begin{tabular}{lcccc}
\toprule
\textbf{Model}
& \makecell{\textbf{Instruction}\\\textbf{Following}}
& \makecell{\textbf{Spatial}\\\textbf{Reasoning}}
& \textbf{Robustness}
& \makecell{\textbf{General}\\\textbf{Manipulation}} \\
\midrule

$\pi_0$~\citep{pi0}
& .3680
& .1300
& .3130
& .3470 \\

Xiaomi-Robotics-0~\citep{xiaomi_robotics_0}
& .6460
& .2230
& .5460
& .3070 \\

GR00T~N1.7~\citep{groot_n17}
& .6460
& .2490
& .5380
& .4380 \\

$\pi_{0.5}$~\citep{pi05}
& .7460
& .3560
& .6130
& .5820 \\

ACoT-VLA~\citep{acot-vla}
& .7570
& .3970
& .6220
& .4770 \\

GigaBrain-0.7
& \textbf{.8166}
& \textbf{.4729}
& \textbf{.6800}
& \textbf{.6092} \\

\bottomrule
\end{tabular}
\end{table*}

Overall, GigaBrain-0.7 exhibits consistently strong performance across
three complementary simulation regimes.
It achieves the best overall and Hard-setting performance on
RoboTwin~2.0 under the multi-task Co-Train protocol, obtains the strongest
EBench result among the publicly available models compared in the main
text, and leads all four evaluated capability dimensions on RoboColiseum.
These results demonstrate that the benefits of large-scale heterogeneous
pretraining extend across different robot embodiments, task structures,
and evaluation domains.
To facilitate reproducible evaluation, our implementation will also be
integrated into XPolicyLab~\citep{xpolicy}, providing a unified workflow for training,
inference, and deployment across robotic learning benchmarks such as
RoboTwin~2.0 and RoboDojo\citep{robodojo}.

%
\subsection{Experience-Driven Reinforcement Learning}
\label{sec:rl_eval}

We finally examine whether GigaBrain-0.7 can continue improving through experience generated during its own real-robot execution, complementing the preceding evaluations of large-scale pretraining and task-specific post-training. Starting from the task-specific SFT policy, we apply the staged experience-reinforcement pipeline described in Section~5.4: rollout experience first supports offline policy refinement, after which the refined policy is redeployed for online reinforcement with human correction. This evaluation measures the incremental gains from demonstration-based specialization to offline experience learning and then to learning from the state distribution induced by the updated policy.

\begin{figure*}[t]
\centering
\captionsetup{
        type=figure,
        justification=justified,
        singlelinecheck=false
    }
\includegraphics[
width=0.99\textwidth
]{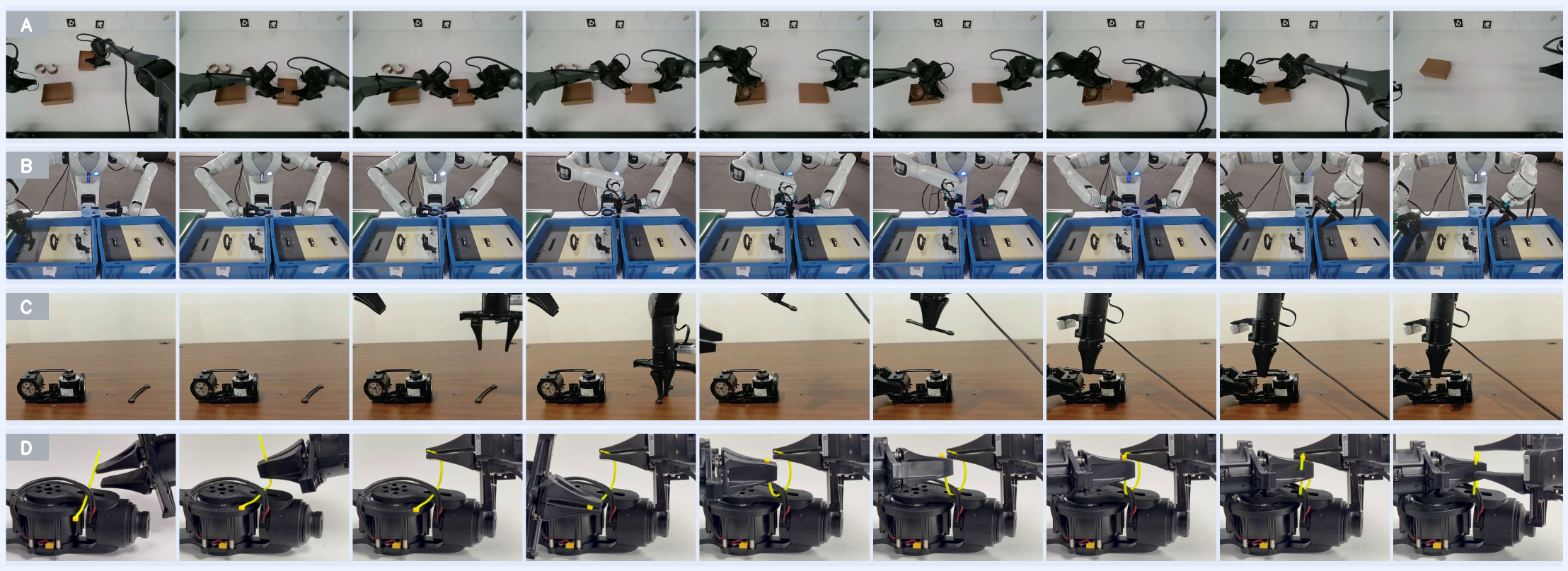}
\caption{
\textbf{Representative real-robot tasks for experience-driven reinforcement learning.}
(A)~Gift Box Packing (PiPER-X).
(B)~Bearing Installation (H01).
(C)~Link Installation (PiPER).
(D)~Cable Tie Insertion and Tying (PiPER-X).
}
\label{fig:experience_driven_rl_tasks}
\end{figure*}

Four representative real-robot tasks expose complementary challenges. Two emphasize long-horizon, multi-stage execution. In \emph{Gift Box Packing} on AgileX PiPER-X, the policy must open the box, place the target object inside, and then manipulate and align the lid; an early error changes the configuration encountered in later stages. \emph{Bearing Installation} on Maker~H01 involves an even longer sequence: the humanoid retrieves a workpiece from the parts bin, transfers it to the assembly station, picks up a bearing, aligns and installs it at the designated location, and returns the assembled workpiece. Two additional tasks emphasize precision and sustained contact at critical stages. In \emph{Link Installation} on AgileX PiPER, the policy must adjust the link pose, align one end with a constrained mounting interface, and perform the insertion. In \emph{Cable Tie Insertion and Tying} on AgileX PiPER-X, the robot must control a slender cable tie while routing it around the target component, threading its free end through the locking head, and pulling it through to tighten the connection. Fig.~\ref{fig:experience_driven_rl_tasks} shows representative executions. Together, these tasks evaluate experience-driven improvement in sequential progress and fine-grained physical interaction across PiPER-family manipulators and the Maker~H01 humanoid.

Three successive training stages are compared under identical task definitions and success criteria. The \emph{SFT Baseline} is the task-specific supervised policy before experience reinforcement. For \emph{Offline RL}, the deployed policy collects real-robot rollouts containing successful, failed, and corrective experience. As described in Section~5.4, rollout segments are evaluated using progress-derived returns and value estimates, and the policy is refined with an advantage-weighted objective that assigns greater weight to higher-progress experience. Failures and partial progress can therefore provide training signal without new interaction during optimization. The refined policy then initializes \emph{Online RL}: it is redeployed to collect on-policy trajectories, with human corrections introduced at difficult states. These corrections enter the training stream at the states that elicited them, while the actor--critic update combines progress-based feedback with terminal task outcomes. All stages refine the same System~1 deployment policy rather than replacing it with a separate task-specific controller.

\begin{table}[t]
\centering
\caption{\textbf{Task success rates at different training stages.}}
\label{tab:performance}
\small
\setlength{\tabcolsep}{4pt}
\begin{tabular}{lccc}
\toprule
Task & SFT Baseline & Offline RL & Online RL \\
\midrule
Link Installation (PiPER)
& 20\% & 40\% & 100\% \\
Gift Box Packing (PiPER-X)
& 80\% & 90\% & 100\% \\
Cable Tie Insertion and Tying (PiPER-X)
& 0\% & 40\% & 100\% \\
Bearing Installation (H01)
& 20\% & 60\% & 100\% \\
\bottomrule
\end{tabular}
\end{table}

Results in Tab.~\ref{tab:performance} show consistent stage-wise improvement on all four tasks. Average success rises from 30.0\% for the SFT baseline to 57.5\% after offline RL and 100\% after online RL. Offline gains depend strongly on the capability of the starting policy. Gift Box Packing, already at 80\% after SFT, improves to 90\%. Bearing Installation and Cable Tie Insertion and Tying each gain 40 percentage points, rising from 20\% to 60\% and from 0\% to 40\%, respectively, while Link Installation improves from 20\% to 40\%. Offline reinforcement thus yields its largest gains where the SFT policy encounters substantial real-robot failures, while still benefiting a task with a strong supervised initialization.

The transition to online reinforcement produces a second substantial improvement. Link Installation and Cable Tie Insertion and Tying rise from 40\% to 100\%, Bearing Installation from 60\% to 100\%, and Gift Box Packing from 90\% to 100\%. The largest gains occur on the two tasks that require resolving precision-critical states during alignment, insertion, or threading. Redeployment is especially valuable in such cases: after offline refinement changes the policy, online rollouts expose the remaining difficult states under the updated policy distribution, enabling human corrections to provide targeted action supervision precisely where needed.

Taken together, the three stages demonstrate complementary roles for supervised demonstrations and deployment experience. SFT provides the task-specific initialization required for physical execution; offline reinforcement learns from differences in the quality of resulting rollouts; and online reinforcement continues this process using experience and corrections collected from the improved policy itself. This pattern holds for both the long-horizon Gift Box Packing and Bearing Installation tasks and the precision-critical Link Installation and Cable Tie Insertion and Tying tasks. Success increases at every stage across all four real-robot evaluations, with the final online-RL policies achieving 100\% success on every task in the reported evaluations.

%% file: sections/conclusion.tex
\section{Conclusion and Future Work}

In this work, we present GigaBrain-0.7, an embodied foundation model that
scales heterogeneous embodied experience and coordinates understanding,
prediction, and action through a three-system architecture.
GigaBrain-0.7 is pretrained on over 37,000 hours of embodied trajectory
data spanning 16 robot morphologies, together with large-scale
vision-language supervision.
Its one-stage VLA pretraining jointly optimizes multimodal understanding,
hierarchical task prediction, discrete action supervision, and continuous
action generation across heterogeneous embodiments.
System~3 is separately pretrained for future-state prediction and
task-progress estimation, and its outputs provide additional conditioning
during task-specific post-training and positive-progress guidance at
inference.
Across real-robot, embodied vision-language, and simulation evaluations,
GigaBrain-0.7 demonstrates strong out-of-the-box generalization,
post-training performance, and robustness across diverse tasks and
embodiments.
Experience-driven offline and online reinforcement learning further
improves the policy using rollout experience and corrective feedback.

Looking ahead, we will continue to scale heterogeneous embodied data and
improve cross-source alignment, strengthen long-horizon predictive modeling
and task-progress estimation, and develop more scalable closed-loop learning
from autonomous experience and human correction.
We hope these efforts will move embodied foundation models toward more
general, robust, and continually improving real-world robot intelligence.

%% file: sections/Appendix.tex
\clearpage
\appendix
\section{Simulation Benchmark Leaderboard Snapshots}
\label{app:sim_leaderboards}

For completeness and reproducibility, we provide snapshots of the
benchmark leaderboards associated with the simulation comparisons in
Section~\ref{sec:simulation_benchmark}.
The main text focuses on representative publicly available methods for
concise and reproducible comparison, whereas this appendix preserves the
complete leaderboard status at the time of manuscript preparation,
including additional submissions not shown in the main-text tables.

Online leaderboards may change as new submissions are added.
We therefore provide the source URL and snapshot access date for each
benchmark below.

\subsection{RoboTwin~2.0}
\label{app:robotwin_leaderboard}

\noindent
\textbf{Leaderboard URL:}
\url{https://robotwin-platform.github.io/leaderboard}

\noindent
\textbf{Snapshot accessed:} August~15, 2026.

\begin{figure*}[htbp]
    \centering
    \captionsetup{
        type=figure,
        justification=justified,
        singlelinecheck=false
    }
    \includegraphics[
        width=0.98\textwidth
    ]{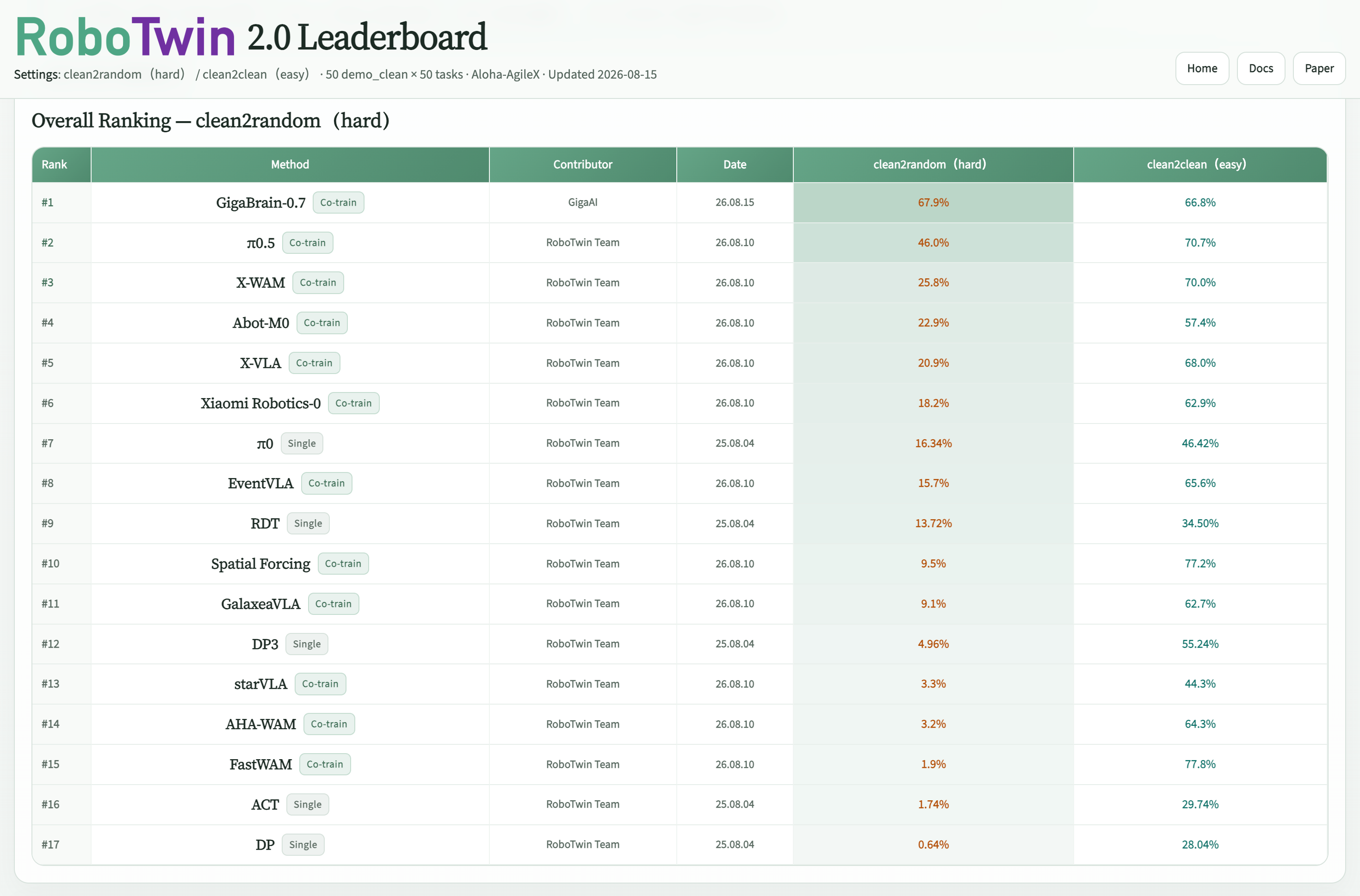}

    \caption{
    \textbf{Snapshot of the official RoboTwin~2.0 leaderboard.}
    The leaderboard reports the evaluation protocol together with
    performance under the clean Easy and domain-randomized Hard settings.
    GigaBrain-0.7 is evaluated under the official Co-Train protocol used
    for the comparison in Tab.~\ref{tab:robotwin2}.
    }
    \label{fig:app_robotwin_leaderboard}
\end{figure*}

\subsection{EBench}
\label{app:ebench_leaderboard}

\noindent
\textbf{Leaderboard URL:}
\url{https://internrobotics.shlab.org.cn/eval/}

\noindent
\textbf{Snapshot accessed:} August~15, 2026.

\begin{figure*}[htbp]
    \centering
    \captionsetup{
        type=figure,
        justification=justified,
        singlelinecheck=false
    }
    \includegraphics[
        width=0.96\textwidth
    ]{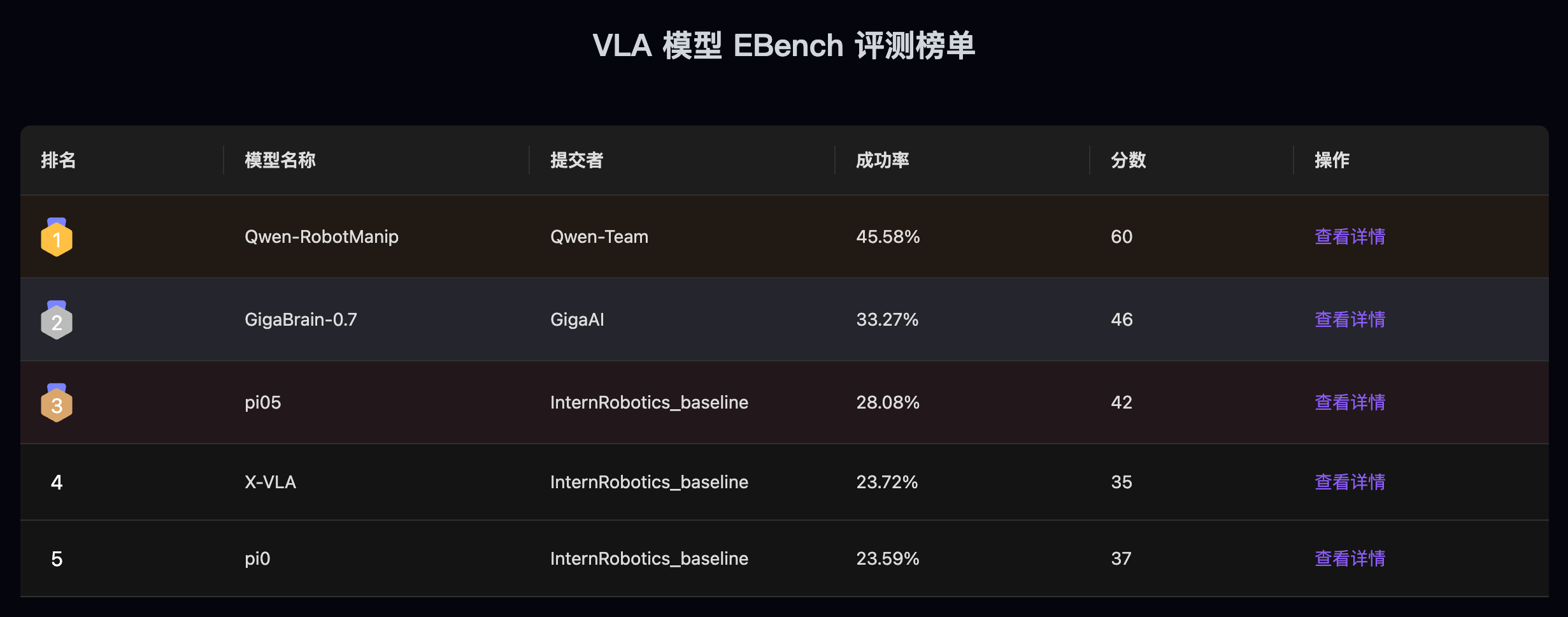}

    \caption{
\textbf{Snapshot of the complete official EBench leaderboard.}
The leaderboard reports overall task success rate and aggregate task
score for all listed submissions at the time of manuscript preparation.
}
    \label{fig:app_ebench_leaderboard}
\end{figure*}

The complete leaderboard snapshot above includes submissions that are not
part of the representative publicly available model comparison in the
main text.
In particular, Qwen-RobotManip is retained here as part of the complete
official leaderboard record.

The InternVLA-A1 result reported in Tab.~\ref{tab:ebench} is not taken
from the online leaderboard snapshot above.
It is reproduced from the EBench evaluation reported by
Qwen-RobotManip~\citep{qwen_robotmanip} and is marked with a superscript
asterisk in the main-text table.

\subsection{RoboColiseum}
\label{app:robocoliseum_leaderboard}

\noindent
\textbf{Leaderboard URL:}
\url{https://robocoliseum.ai/}

\noindent
\textbf{Snapshot accessed:} August~15, 2026.

\begin{figure*}[htbp]
    \centering
    \captionsetup{
        type=figure,
        justification=justified,
        singlelinecheck=false
    }

    \includegraphics[
        width=0.96\textwidth
    ]{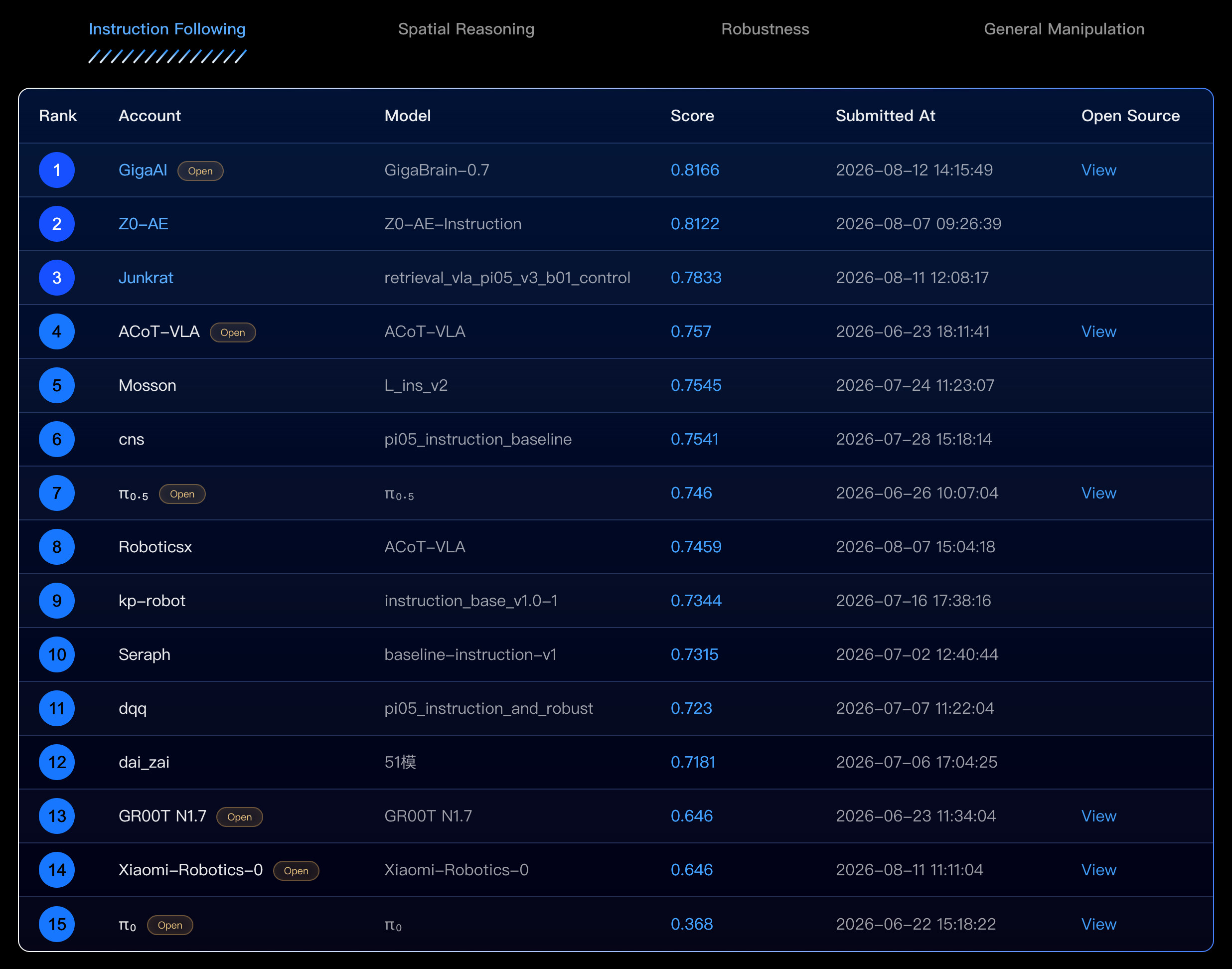}

    \caption{
    \textbf{Snapshot of the RoboColiseum Instruction Following leaderboard.}
    The leaderboard evaluates the ability of robot policies to follow
    language-conditioned instructions at the time of manuscript preparation.
    }
    \label{fig:app_robocoliseum_instruction}
\end{figure*}

\begin{figure*}[htbp]
    \centering
    \captionsetup{
        type=figure,
        justification=justified,
        singlelinecheck=false
    }

    \includegraphics[
        width=0.96\textwidth
    ]{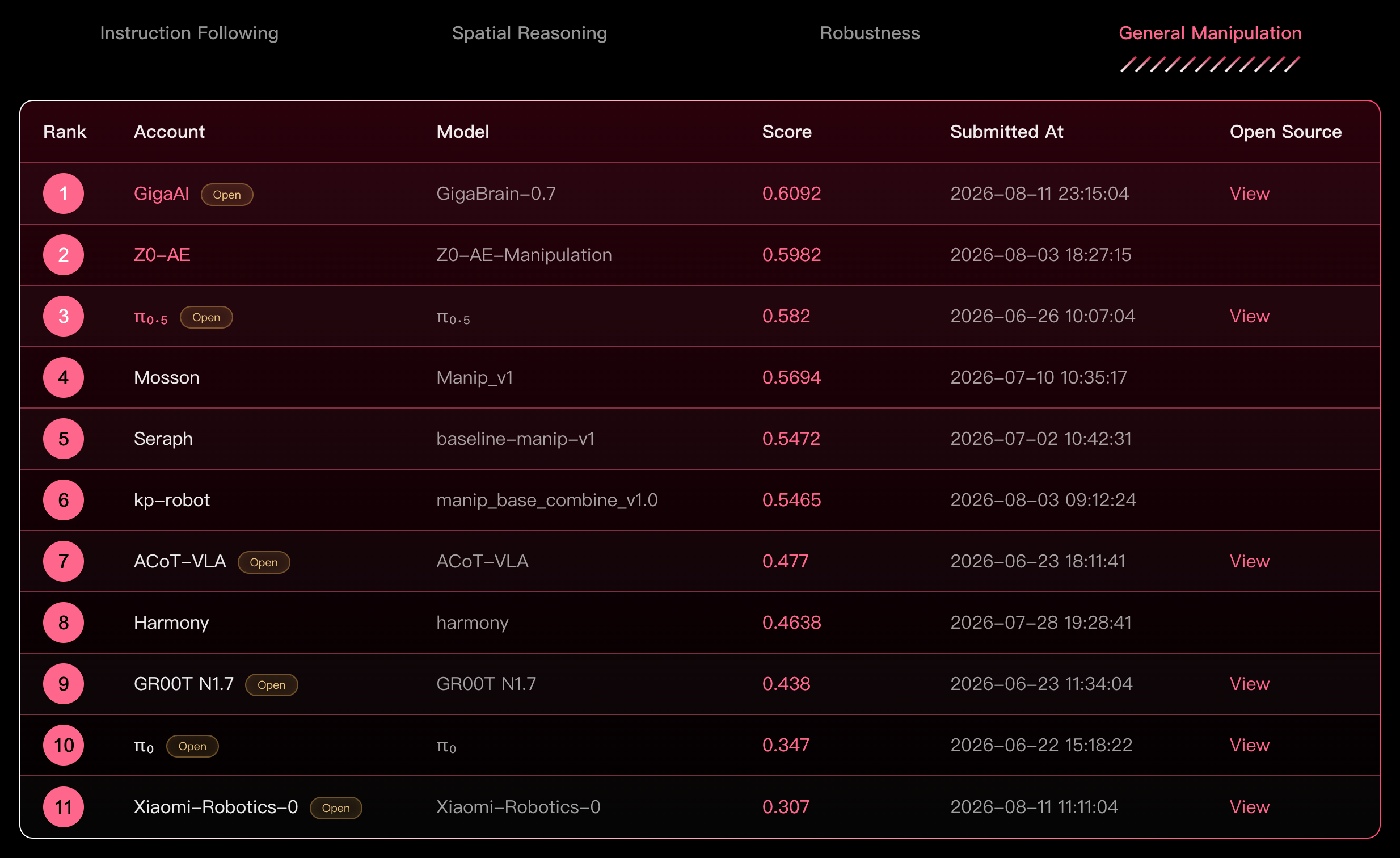}

    \caption{
    \textbf{Snapshot of the RoboColiseum General Manipulation leaderboard.}
    The leaderboard evaluates general manipulation capability at the time
    of manuscript preparation.
    }
    \label{fig:app_robocoliseum_manipulation}
\end{figure*}

RoboColiseum reports separate leaderboards for four complementary
capability dimensions: instruction following, spatial reasoning,
robustness, and general manipulation.
Figures~\ref{fig:app_robocoliseum_instruction}--%
\ref{fig:app_robocoliseum_manipulation} provide separate snapshots of
the four leaderboard dimensions associated with the comparison in
Tab.~\ref{tab:robocoliseum}.

\begin{figure*}[htbp]
    \centering
    \captionsetup{
        type=figure,
        justification=justified,
        singlelinecheck=false
    }

    \includegraphics[
        width=0.96\textwidth
    ]{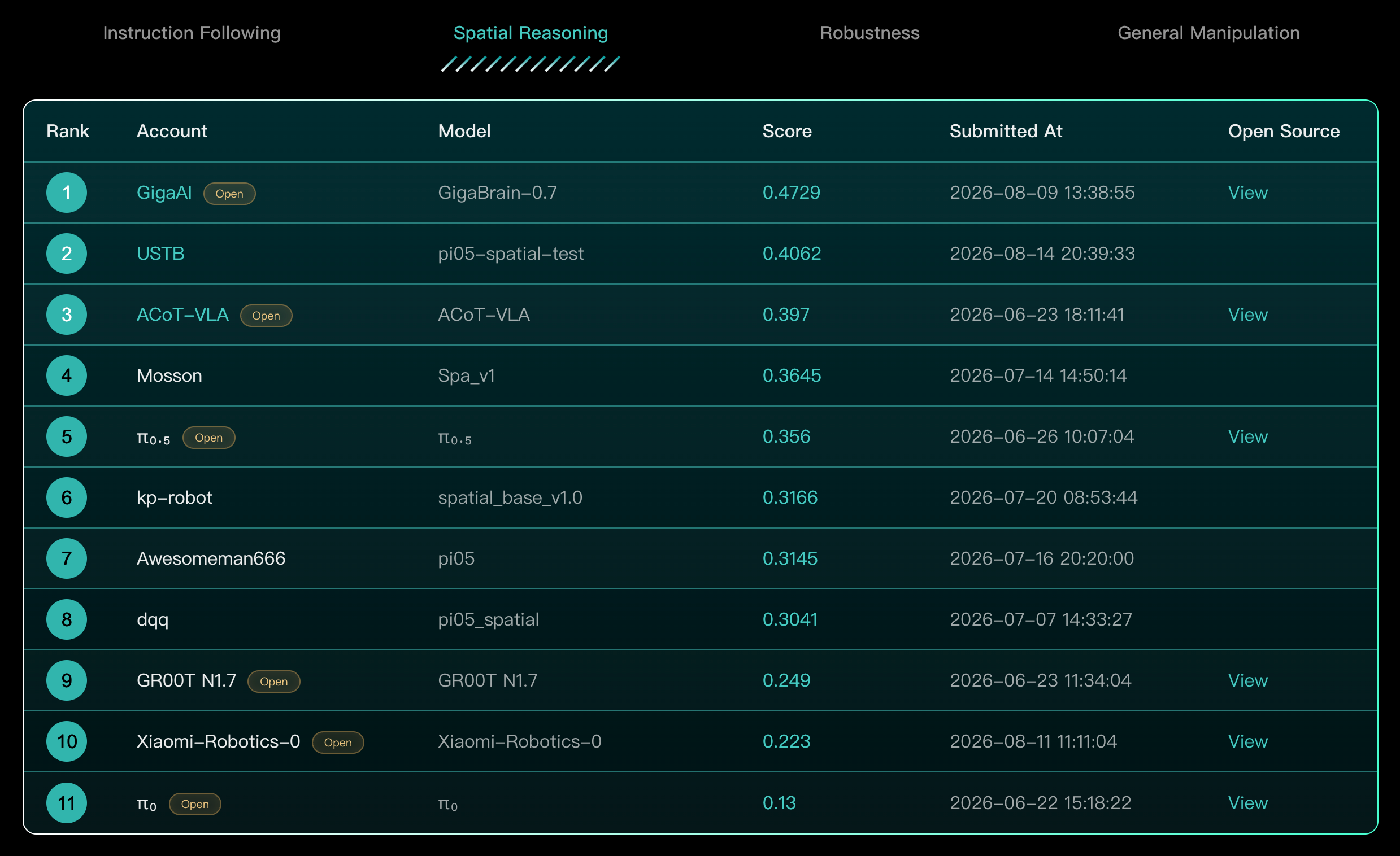}

    \caption{
    \textbf{Snapshot of the RoboColiseum Spatial Reasoning leaderboard.}
    The leaderboard evaluates spatial reasoning capability at the time of
    manuscript preparation.
    }
    \label{fig:app_robocoliseum_spatial}
\end{figure*}

\begin{figure*}[htbp]
    \centering
    \captionsetup{
        type=figure,
        justification=justified,
        singlelinecheck=false
    }

    \includegraphics[
        width=0.96\textwidth
    ]{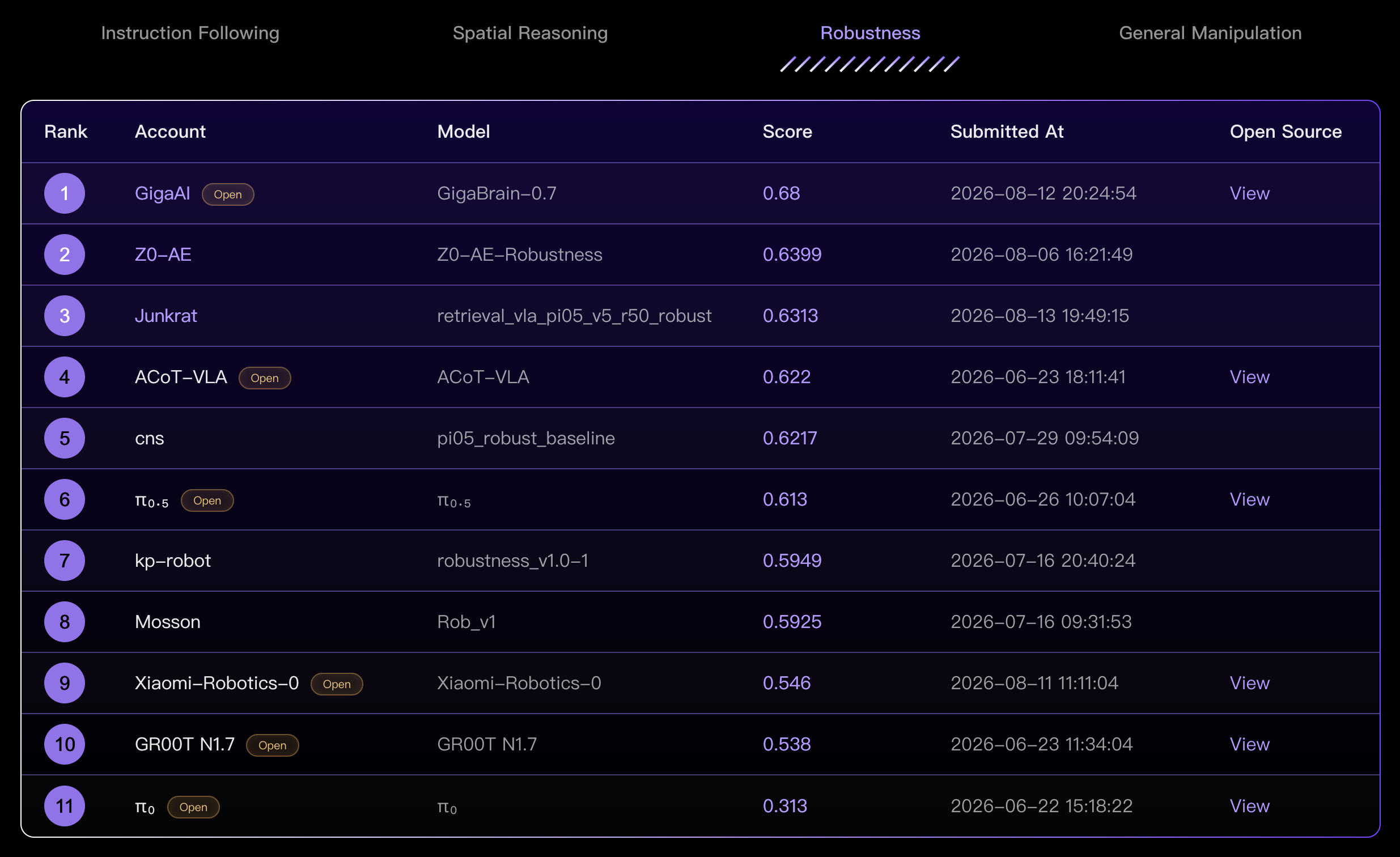}

    \caption{
    \textbf{Snapshot of the RoboColiseum Robustness leaderboard.}
    The leaderboard evaluates policy robustness under the corresponding
    RoboColiseum protocol at the time of manuscript preparation.
    }
    \label{fig:app_robocoliseum_robustness}
\end{figure*}

%% file: main.bbl
\begin{thebibliography}{95}
\providecommand{\natexlab}[1]{#1}
\providecommand{\url}[1]{\texttt{#1}}
\expandafter\ifx\csname urlstyle\endcsname\relax
  \providecommand{\doi}[1]{doi: #1}\else
  \providecommand{\doi}{doi: \begingroup \urlstyle{rm}\Url}\fi

\bibitem[rob(2026)]{robocoliseum}
Robocoliseum: A comprehensive, multi-dimensional benchmark for evaluating core capabilities of robotic models.
\newblock \url{https://robocoliseum.ai/}, 2026.
\newblock Accessed: 2026-08-14.

\bibitem[{AgiBot-World-Contributors} et~al.(2025){AgiBot-World-Contributors}, Bu, Cai, Chen, Cui, Ding, Feng, Gao, He, Huang, et~al.]{agibotworld2025colosseo}
{AgiBot-World-Contributors}, Qingwen Bu, Jisong Cai, Li~Chen, Xiuqi Cui, Yan Ding, Siyuan Feng, Shenyuan Gao, Xindong He, Xu~Huang, et~al.
\newblock Agibot world colosseo: A large-scale manipulation platform for scalable and intelligent embodied systems.
\newblock \emph{arXiv preprint arXiv:2503.06669}, 2025.
\newblock URL \url{https://arxiv.org/abs/2503.06669}.

\bibitem[{AgiBot World Team}(2026)]{agibotworld2026}
{AgiBot World Team}.
\newblock Agibot world 2026.
\newblock \url{https://huggingface.co/datasets/agibot-world/AgiBotWorld2026}, 2026.

\bibitem[Bain et~al.(2021)Bain, Nagrani, Varol, and Zisserman]{webvid10m}
Max Bain, Arsha Nagrani, G{\"u}l Varol, and Andrew Zisserman.
\newblock Frozen in time: A joint video and image encoder for end-to-end retrieval.
\newblock In \emph{Proceedings of the IEEE/CVF International Conference on Computer Vision}, pages 1728--1738, 2021.
\newblock URL \url{https://openaccess.thecvf.com/content/ICCV2021/html/Bain_Frozen_in_Time_A_Joint_Video_and_Image_Encoder_for_ICCV_2021_paper.html}.

\bibitem[Black et~al.(2024)Black, Brown, Driess, Esmail, Equi, Finn, Fusai, Groom, Hausman, Ichter, Jakubczak, Jones, Ke, Levine, Li-Bell, Mothukuri, Nair, Pertsch, Shi, Tanner, Vuong, Walling, Wang, and Zhilinsky]{pi0}
Kevin Black, Noah Brown, Danny Driess, Adnan Esmail, Michael Equi, Chelsea Finn, Niccolo Fusai, Lachy Groom, Karol Hausman, Brian Ichter, Szymon Jakubczak, Tim Jones, Liyiming Ke, Sergey Levine, Adrian Li-Bell, Mohith Mothukuri, Suraj Nair, Karl Pertsch, Lucy~Xiaoyang Shi, James Tanner, Quan Vuong, Anna Walling, Haohuan Wang, and Ury Zhilinsky.
\newblock {$\pi_0$: A Vision-Language-Action Flow Model for General Robot Control}, 2024.
\newblock URL \url{https://arxiv.org/abs/2410.24164}.
\newblock arXiv preprint arXiv:2410.24164.

\bibitem[Brohan et~al.(2023)Brohan, Brown, Carbajal, Chebotar, Chen, Choromanski, Ding, Driess, Dubey, Finn, Florence, Fu, Arenas, Gopalakrishnan, Han, Hausman, Herzog, Hsu, Ichter, Irpan, Joshi, Julian, Kalashnikov, Kuang, Leal, Lee, Lee, Levine, Lu, Michalewski, Mordatch, Pertsch, Rao, Reymann, Ryoo, Salazar, Sanketi, Sermanet, Singh, Singh, Soricut, Tran, Vanhoucke, Vuong, Wahid, Welker, Wohlhart, Wu, Xia, Xiao, Xu, Xu, Yu, and Zitkovich]{rt2}
Anthony Brohan, Noah Brown, Justice Carbajal, Yevgen Chebotar, Xi~Chen, Krzysztof Choromanski, Tianli Ding, Danny Driess, Avinava Dubey, Chelsea Finn, Pete Florence, Chuyuan Fu, Montse~Gonzalez Arenas, Keerthana Gopalakrishnan, Kehang Han, Karol Hausman, Alexander Herzog, Jasmine Hsu, Brian Ichter, Alex Irpan, Nikhil Joshi, Ryan Julian, Dmitry Kalashnikov, Yuheng Kuang, Isabel Leal, Lisa Lee, Tsang-Wei~Edward Lee, Sergey Levine, Yao Lu, Henryk Michalewski, Igor Mordatch, Karl Pertsch, Kanishka Rao, Krista Reymann, Michael Ryoo, Grecia Salazar, Pannag Sanketi, Pierre Sermanet, Jaspiar Singh, Anikait Singh, Radu Soricut, Huong Tran, Vincent Vanhoucke, Quan Vuong, Ayzaan Wahid, Stefan Welker, Paul Wohlhart, Jialin Wu, Fei Xia, Ted Xiao, Peng Xu, Sichun Xu, Tianhe Yu, and Brianna Zitkovich.
\newblock {RT-2: Vision-Language-Action Models Transfer Web Knowledge to Robotic Control}, 2023.
\newblock URL \url{https://arxiv.org/abs/2307.15818}.
\newblock arXiv preprint arXiv:2307.15818.

\bibitem[Cadene et~al.(2026)Cadene, Aliberts, Capuano, Aractingi, Zouitine, Kooijmans, Choghari, Russi, Pascal, Palma, Shukor, Moss, Soare, Aubakirova, Lhoest, Gallouedec, and Wolf]{cadene2026lerobot}
Remi Cadene, Simon Aliberts, Francesco Capuano, Michel Aractingi, Adil Zouitine, Pepijn Kooijmans, Jade Choghari, Martino Russi, Caroline Pascal, Steven Palma, Mustafa Shukor, Jess Moss, Alexander Soare, Dana Aubakirova, Quentin Lhoest, Quentin Gallouedec, and Thomas Wolf.
\newblock {LeRobot}: An open-source library for end-to-end robot learning.
\newblock \emph{arXiv preprint arXiv:2602.22818}, 2026.
\newblock URL \url{https://arxiv.org/abs/2602.22818}.

\bibitem[Cai et~al.(2026{\natexlab{a}})Cai, Cai, Cao, Chen, He, Jiang, Li, Li, Li, Liu, Lu, Lv, Ma, Pang, Qiao, Qiu, Shen, Shi, Tian, Wang, Wang, Wang, Wang, Wei, Wu, Xie, Xing, Yang, Yang, Yu, Yuan, Zeng, Zhang, Zhang, Zhang, Zhaxi, Zhou, Zhou, Zhou, Zhu, Zhu, and Zhu]{internvla}
Junhao Cai, Zetao Cai, Jiafei Cao, Yilun Chen, Zeyu He, Lei Jiang, Hang Li, Hengjie Li, Yang Li, Yufei Liu, Yanan Lu, Qi~Lv, Haoxiang Ma, Jiangmiao Pang, Yu~Qiao, Zherui Qiu, Yanqing Shen, Xu~Shi, Yang Tian, Bolun Wang, Hanqing Wang, Jiaheng Wang, Tai Wang, Xueyuan Wei, Chao Wu, Yiman Xie, Boyang Xing, Yuqiang Yang, Yuyin Yang, Qiaojun Yu, Feng Yuan, Jia Zeng, Jingjing Zhang, Shenghan Zhang, Shi Zhang, Zhuoma Zhaxi, Bowen Zhou, Yuanzhen Zhou, Yunsong Zhou, Hongrui Zhu, Yangkun Zhu, and Yuchen Zhu.
\newblock Internvla-a1: Unifying understanding, generation and action for robotic manipulation, 2026{\natexlab{a}}.
\newblock URL \url{https://arxiv.org/abs/2601.02456}.

\bibitem[Cai et~al.(2026{\natexlab{b}})Cai, Guo, He, Jin, Li, Lin, Liu, Liu, Ma, Ma, Qiu, Qu, Su, Sun, Wang, Wang, Wang, Wu, Xiang, Yang, Ye, Zhang, and Zhou]{xiaomi_robotics_0}
Rui Cai, Jun Guo, Xinze He, Piaopiao Jin, Jie Li, Bingxuan Lin, Futeng Liu, Wei Liu, Fei Ma, Kun Ma, Feng Qiu, Heng Qu, Yifei Su, Qiao Sun, Dong Wang, Donghao Wang, Yunhong Wang, Rujie Wu, Diyun Xiang, Yu~Yang, Hangjun Ye, Yuan Zhang, and Quanyun Zhou.
\newblock {Xiaomi-Robotics-0: An Open-Sourced Vision-Language-Action Model with Real-Time Execution}, 2026{\natexlab{b}}.
\newblock URL \url{https://arxiv.org/abs/2602.12684}.
\newblock arXiv preprint arXiv:2602.12684.

\bibitem[Cao et~al.(2026)Cao, Chen, Li, Wang, Peng, and Li]{z1}
Lang Cao, Renhong Chen, Luyi Li, Peng Wang, Mofan Peng, and Yitong Li.
\newblock Z-1: Efficient reinforcement learning for vision-language-action models, 2026.
\newblock URL \url{https://arxiv.org/abs/2606.31846}.

\bibitem[Changpinyo et~al.(2021)Changpinyo, Sharma, Ding, and Soricut]{cc12m}
Soravit Changpinyo, Piyush Sharma, Nan Ding, and Radu Soricut.
\newblock Conceptual {12M}: Pushing web-scale image--text pre-training to recognize long-tail visual concepts.
\newblock In \emph{Proceedings of the IEEE/CVF Conference on Computer Vision and Pattern Recognition}, pages 3558--3568, 2021.
\newblock URL \url{https://openaccess.thecvf.com/content/CVPR2021/html/Changpinyo_Conceptual_12M_Pushing_Web-Scale_Image-Text_Pre-Training_To_Recognize_Long-Tail_Visual_CVPR_2021_paper.html}.

\bibitem[Cheang et~al.(2024)Cheang, Chen, Jing, Kong, Li, Li, Liu, Wu, Xu, Yang, Zhang, and Zhu]{gr2}
Chi-Lam Cheang, Guangzeng Chen, Ya~Jing, Tao Kong, Hang Li, Yifeng Li, Yuxiao Liu, Hongtao Wu, Jiafeng Xu, Yichu Yang, Hanbo Zhang, and Minzhao Zhu.
\newblock {GR-2: A Generative Video-Language-Action Model with Web-Scale Knowledge for Robot Manipulation}, 2024.
\newblock URL \url{https://arxiv.org/abs/2410.06158}.
\newblock arXiv preprint arXiv:2410.06158.

\bibitem[Chen et~al.(2025)Chen, Xie, Ma, Sanketi, and Goldberg]{robo2vlm}
Kaiyuan~Eric Chen, Shuangyu Xie, Zehan Ma, Pannag Sanketi, and Ken Goldberg.
\newblock {Robo2VLM}: Improving visual question answering using large-scale robot manipulation data.
\newblock In \emph{Advances in Neural Information Processing Systems: Datasets and Benchmarks Track}, volume~38, 2025.
\newblock URL \url{https://proceedings.neurips.cc/paper_files/paper/2025/hash/1f467c3e37abf9f86c78f44c6a27ee7c-Abstract-Datasets_and_Benchmarks_Track.html}.

\bibitem[Chen et~al.(2026{\natexlab{a}})Chen, Chen, Li, Tang, Su, Wan, Chen, Lu, Yan, Su, et~al.]{robodojo}
Tianxing Chen, Yue Chen, Zixuan Li, Junyuan Tang, Kailun Su, Weijie Wan, Baijun Chen, Haoran Lu, Haowen Yan, Honghao Su, et~al.
\newblock {RoboDojo}: A unified sim-and-real benchmark for comprehensive evaluation of generalist robot manipulation policies.
\newblock \emph{arXiv preprint arXiv:2607.04434}, 2026{\natexlab{a}}.

\bibitem[Chen et~al.(2026{\natexlab{b}})Chen, Chen, Chen, Cai, Liu, Li, Liang, Lin, Ge, Gu, Deng, Guo, Nian, Xie, Chen, Su, Xu, Liu, Hu, ang Gao, Wang, Liang, Qin, Yang, Luo, and Mu]{robotwin2}
Tianxing Chen, Zanxin Chen, Baijun Chen, Zijian Cai, Yibin Liu, Zixuan Li, Qiwei Liang, Xianliang Lin, Yiheng Ge, Zhenyu Gu, Weiliang Deng, Yubin Guo, Tian Nian, Xuanbing Xie, Qiangyu Chen, Kailun Su, Tianling Xu, Guodong Liu, Mengkang Hu, Huan ang Gao, Kaixuan Wang, Zhixuan Liang, Yusen Qin, Xiaokang Yang, Ping Luo, and Yao Mu.
\newblock Robotwin 2.0: A scalable data generator and benchmark with strong domain randomization for robust bimanual robotic manipulation.
\newblock In \emph{Forty-third International Conference on Machine Learning}, 2026{\natexlab{b}}.
\newblock URL \url{https://openreview.net/forum?id=itonej9GIV}.

\bibitem[Chi et~al.(2024)Chi, Xu, Pan, Cousineau, Burchfiel, Feng, Tedrake, and Song]{umi}
Cheng Chi, Zhenjia Xu, Chuer Pan, Eric Cousineau, Benjamin Burchfiel, Siyuan Feng, Russ Tedrake, and Shuran Song.
\newblock {Universal Manipulation Interface: In-The-Wild Robot Teaching Without In-The-Wild Robots}.
\newblock In \emph{Proceedings of Robotics: Science and Systems}, Delft, Netherlands, July 2024.
\newblock \doi{10.15607/RSS.2024.XX.045}.
\newblock URL \url{https://www.roboticsproceedings.org/rss20/p045.html}.

\bibitem[Community et~al.(2026)Community, Chen, Chen, Nian, Cai, Chen, Lin, Liang, Xiang, Su, et~al.]{xpolicy}
XPolicyLab Community, Tianxing Chen, Yue Chen, Tian Nian, Zijian Cai, Guangyu Chen, Wenwei Lin, Qiwei Liang, Peicheng Xiang, Kailun Su, et~al.
\newblock {XPolicyLab}: A unified standard and open ecosystem for robot policy evaluation and deployment.
\newblock \emph{arXiv preprint arXiv:2608.09892}, 2026.

\bibitem[Deitke et~al.(2025)Deitke, Clark, Lee, Tripathi, Yang, Park, Salehi, Muennighoff, Lo, Soldaini, Lu, Anderson, Bransom, Ehsani, Ngo, Chen, Patel, Yatskar, Callison-Burch, Head, Hendrix, Bastani, VanderBilt, Lambert, Chou, Chheda, Sparks, Skjonsberg, Schmitz, Sarnat, Bischoff, Walsh, Newell, Wolters, Gupta, Zeng, Borchardt, Groeneveld, Nam, Lebrecht, Wittlif, Schoenick, Michel, Krishna, Weihs, Smith, Hajishirzi, Girshick, Farhadi, and Kembhavi]{pixmo}
Matt Deitke, Christopher Clark, Sangho Lee, Rohun Tripathi, Yue Yang, Jae~Sung Park, Mohammadreza Salehi, Niklas Muennighoff, Kyle Lo, Luca Soldaini, Jiasen Lu, Taira Anderson, Erin Bransom, Kiana Ehsani, Huong Ngo, YenSung Chen, Ajay Patel, Mark Yatskar, Chris Callison-Burch, Andrew Head, Rose Hendrix, Favyen Bastani, Eli VanderBilt, Nathan Lambert, Yvonne Chou, Arnavi Chheda, Jenna Sparks, Sam Skjonsberg, Michael Schmitz, Aaron Sarnat, Byron Bischoff, Pete Walsh, Chris Newell, Piper Wolters, Tanmay Gupta, Kuo-Hao Zeng, Jon Borchardt, Dirk Groeneveld, Crystal Nam, Sophie Lebrecht, Caitlin Wittlif, Carissa Schoenick, Oscar Michel, Ranjay Krishna, Luca Weihs, Noah~A. Smith, Hannaneh Hajishirzi, Ross Girshick, Ali Farhadi, and Aniruddha Kembhavi.
\newblock Molmo and {PixMo}: Open weights and open data for state-of-the-art vision--language models.
\newblock In \emph{Proceedings of the IEEE/CVF Conference on Computer Vision and Pattern Recognition}, pages 91--104, 2025.
\newblock \doi{10.1109/CVPR52734.2025.00018}.
\newblock URL \url{https://openaccess.thecvf.com/content/CVPR2025/html/Deitke_Molmo_and_PixMo_Open_Weights_and_Open_Data_for_State-of-the-Art_CVPR_2025_paper.html}.

\bibitem[Dong et~al.(2026)Dong, Liu, Zhang, Ye, Yuan, Ni, Gong, Qiu, Zhao, Li, and Hao]{actioncodec}
Zibin Dong, Yicheng Liu, Shiduo Zhang, Baijun Ye, Yifu Yuan, Fei Ni, Jingjing Gong, Xipeng Qiu, Hang Zhao, Yinchuan Li, and Jianye Hao.
\newblock {ActionCodec: What Makes for Good Action Tokenizers}, 2026.
\newblock URL \url{https://arxiv.org/abs/2602.15397}.
\newblock arXiv preprint arXiv:2602.15397.

\bibitem[Driess et~al.(2025)Driess, Springenberg, Ichter, Yu, Li-Bell, Pertsch, Ren, Walke, Vuong, Shi, and Levine]{knowledge_insulation}
Danny Driess, Jost Springenberg, Brian Ichter, Lili Yu, Adrian Li-Bell, Karl Pertsch, Allen Ren, Homer Walke, Quan Vuong, Lucy~Xiaoyang Shi, and Sergey Levine.
\newblock {Knowledge Insulating Vision-Language-Action Models: Train Fast, Run Fast, Generalize Better}.
\newblock In \emph{Advances in Neural Information Processing Systems}, volume~38, pages 102867--102888. Curran Associates, Inc., 2025.
\newblock \doi{10.52202/085713-3439}.
\newblock URL \url{https://proceedings.neurips.cc/paper_files/paper/2025/file/94e936034d12bcd04834ec2773f02aff-Paper-Conference.pdf}.

\bibitem[{Fhrozen}(2025)]{localized_narratives}
{Fhrozen}.
\newblock {Open Images Narratives v2}.
\newblock Hugging Face dataset, 2025.
\newblock URL \url{https://huggingface.co/datasets/Fhrozen/openimages-narratives-v2}.
\newblock Accessed 2026-08-13.

\bibitem[{Galaxea Team}(2025)]{galaxea2025openworld}
{Galaxea Team}.
\newblock Galaxea open-world dataset and g0 dual-system vla model.
\newblock \emph{arXiv preprint arXiv:2509.00576}, 2025.
\newblock URL \url{https://arxiv.org/abs/2509.00576}.

\bibitem[{Galaxea Team}(2026)]{galaxea_g05}
{Galaxea Team}.
\newblock {Galaxea G0.5 Technical Report}.
\newblock Technical report, Galaxea AI, 2026.
\newblock URL \url{https://opengalaxea.github.io/G05/Galaxea_G0_5.pdf}.

\bibitem[Gao et~al.(2026)Gao, Zheng, Gao, Ma, Wang, Wang, Chen, Chen, Zhang, Jia, Jiang, Zhu, Li, Wang, Li, Cai, Yang, Xu, Lyu, Mu, Wang, Pang, Zeng, Zhang, and Shen]{ebench}
Ning Gao, Jinliang Zheng, Xing Gao, Haoxiang Ma, Hanqing Wang, Yukai Wang, Jiantong Chen, Zanxin Chen, Shujie Zhang, Mingda Jia, Xuekun Jiang, Zihou Zhu, Xinyu Li, Shuai Wang, Hao Li, Wenzhe Cai, Yuqiang Yang, Xudong Xu, Zhaoyang Lyu, Yao Mu, Tai Wang, Jiangmiao Pang, Jia Zeng, Weinan Zhang, and Chunhua Shen.
\newblock Ebench: Elemental diagnosis of generalist mobile manipulation policies, 2026.
\newblock URL \url{https://arxiv.org/abs/2606.18239}.

\bibitem[{Gemini Robotics Team} et~al.(2025){Gemini Robotics Team}, Abeyruwan, Ainslie, Alayrac, Arenas, Armstrong, Balakrishna, Baruch, Bauza, Blokzijl, Bohez, Bousmalis, Brohan, Buschmann, Byravan, Cabi, Caluwaerts, Casarini, Chang, Chen, Chen, Chiang, Choromanski, D'Ambrosio, Dasari, Davchev, Devin, Palo, Ding, Dostmohamed, Driess, Du, Dwibedi, Elabd, Fantacci, Fong, Frey, Fu, Giustina, Gopalakrishnan, Graesser, Hasenclever, Heess, Hernaez, Herzog, Hofer, Humplik, Iscen, Jacob, Jain, Julian, Kalashnikov, Karagozler, Karp, Kew, Kirkland, Kirmani, Kuang, Lampe, Laurens, Leal, Lee, Lee, Liang, Lin, Maddineni, Majumdar, Michaely, Moreno, Neunert, Nori, Parada, Parisotto, Pastor, Pooley, Rao, Reymann, Sadigh, Saliceti, Sanketi, Sermanet, Shah, Sharma, Shea, Shu, Sindhwani, Singh, Soricut, Springenberg, Sterneck, Surdulescu, Tan, Tompson, Vanhoucke, Varley, Vesom, Vezzani, Vinyals, Wahid, Welker, Wohlhart, Xia, Xiao, Xie, Xie, Xu, Xu, Xu, Xu, Yang, Yao, Yaroshenko, Yu, Yuan, Zhang, Zhang, Zhou, and
  Zhou]{gemini_robotics}
{Gemini Robotics Team}, Saminda Abeyruwan, Joshua Ainslie, Jean-Baptiste Alayrac, Montserrat~Gonzalez Arenas, Travis Armstrong, Ashwin Balakrishna, Robert Baruch, Maria Bauza, Michiel Blokzijl, Steven Bohez, Konstantinos Bousmalis, Anthony Brohan, Thomas Buschmann, Arunkumar Byravan, Serkan Cabi, Ken Caluwaerts, Federico Casarini, Oscar Chang, Jose~Enrique Chen, Xi~Chen, Hao-Tien~Lewis Chiang, Krzysztof Choromanski, David D'Ambrosio, Sudeep Dasari, Todor Davchev, Coline Devin, Norman~Di Palo, Tianli Ding, Adil Dostmohamed, Danny Driess, Yilun Du, Debidatta Dwibedi, Michael Elabd, Claudio Fantacci, Cody Fong, Erik Frey, Chuyuan Fu, Marissa Giustina, Keerthana Gopalakrishnan, Laura Graesser, Leonard Hasenclever, Nicolas Heess, Brandon Hernaez, Alexander Herzog, R.~Alex Hofer, Jan Humplik, Atil Iscen, Mithun~George Jacob, Deepali Jain, Ryan Julian, Dmitry Kalashnikov, M.~Emre Karagozler, Stefani Karp, Chase Kew, Jerad Kirkland, Sean Kirmani, Yuheng Kuang, Thomas Lampe, Antoine Laurens, Isabel Leal, Alex~X. Lee,
  Tsang-Wei~Edward Lee, Jacky Liang, Yixin Lin, Sharath Maddineni, Anirudha Majumdar, Assaf~Hurwitz Michaely, Robert Moreno, Michael Neunert, Francesco Nori, Carolina Parada, Emilio Parisotto, Peter Pastor, Acorn Pooley, Kanishka Rao, Krista Reymann, Dorsa Sadigh, Stefano Saliceti, Pannag Sanketi, Pierre Sermanet, Dhruv Shah, Mohit Sharma, Kathryn Shea, Charles Shu, Vikas Sindhwani, Sumeet Singh, Radu Soricut, Jost~Tobias Springenberg, Rachel Sterneck, Razvan Surdulescu, Jie Tan, Jonathan Tompson, Vincent Vanhoucke, Jake Varley, Grace Vesom, Giulia Vezzani, Oriol Vinyals, Ayzaan Wahid, Stefan Welker, Paul Wohlhart, Fei Xia, Ted Xiao, Annie Xie, Jinyu Xie, Peng Xu, Sichun Xu, Ying Xu, Zhuo Xu, Yuxiang Yang, Rui Yao, Sergey Yaroshenko, Wenhao Yu, Wentao Yuan, Jingwei Zhang, Tingnan Zhang, Allan Zhou, and Yuxiang Zhou.
\newblock {Gemini Robotics: Bringing AI into the Physical World}, 2025.
\newblock URL \url{https://arxiv.org/abs/2503.20020}.
\newblock arXiv preprint arXiv:2503.20020.

\bibitem[{GenRobot}(2026)]{jianzhi10k2026}
{GenRobot}.
\newblock 10kh-realomin-opendata: Jianzhi 10k umi data source.
\newblock Hugging Face dataset, 2026.
\newblock URL \url{https://huggingface.co/datasets/genrobot2025/10Kh-RealOmin-OpenData}.

\bibitem[{GigaBrain Team} et~al.(2025){GigaBrain Team}, Ye, Wang, Ni, Huang, Zhao, Li, Li, Zhu, Feng, Li, Deng, Ouyang, Qin, Chen, Wang, Wang, Li, Li, Ding, Xu, Ye, Zhou, Dong, Wang, Liu, and Zhu]{gigabrain0}
{GigaBrain Team}, Angen Ye, Boyuan Wang, Chaojun Ni, Guan Huang, Guosheng Zhao, Haoyun Li, Jie Li, Jiagang Zhu, Lv~Feng, Peng Li, Qiuping Deng, Runqi Ouyang, Wenkang Qin, Xinze Chen, Xiaofeng Wang, Yang Wang, Yifan Li, Yilong Li, Yiran Ding, Yuan Xu, Yun Ye, Yukun Zhou, Zhehao Dong, Zhenan Wang, Zhichao Liu, and Zheng Zhu.
\newblock {GigaBrain-0: A World Model-Powered Vision-Language-Action Model}, 2025.
\newblock URL \url{https://arxiv.org/abs/2510.19430}.
\newblock arXiv preprint arXiv:2510.19430.

\bibitem[{GigaBrain Team} et~al.(2026){GigaBrain Team}, Wang, Li, Ni, Huang, Zhao, Li, Li, Lv, Liu, Feng, Yu, Li, Deng, Liu, Zhou, Chen, Wang, Wang, Li, Nie, Li, Zhou, Ye, Liu, and Zhu]{gigabrain05m}
{GigaBrain Team}, Boyuan Wang, Bohan Li, Chaojun Ni, Guan Huang, Guosheng Zhao, Hao Li, Jie Li, Jindi Lv, Jingyu Liu, Lv~Feng, Mingming Yu, Peng Li, Qiuping Deng, Tianze Liu, Xinyu Zhou, Xinze Chen, Xiaofeng Wang, Yang Wang, Yifan Li, Yifei Nie, Yilong Li, Yukun Zhou, Yun Ye, Zhichao Liu, and Zheng Zhu.
\newblock {GigaBrain-0.5M*: a VLA That Learns From World Model-Based Reinforcement Learning}, 2026.
\newblock URL \url{https://arxiv.org/abs/2602.12099}.
\newblock arXiv preprint arXiv:2602.12099.

\bibitem[{GLM-5-Team} et~al.(2026){GLM-5-Team}, Zeng, Lv, Hou, Du, Zheng, Chen, Yin, Ge, Huang, et~al.]{glm5team2026glm5}
{GLM-5-Team}, Aohan Zeng, Xin Lv, Zhenyu Hou, Zhengxiao Du, Qinkai Zheng, Bin Chen, Da~Yin, Chendi Ge, Chenghua Huang, et~al.
\newblock {GLM-5}: From vibe coding to agentic engineering, 2026.
\newblock URL \url{https://arxiv.org/abs/2602.15763}.

\bibitem[Goyal et~al.(2017)Goyal, Khot, Summers-Stay, Batra, and Parikh]{vqav2}
Yash Goyal, Tejas Khot, Douglas Summers-Stay, Dhruv Batra, and Devi Parikh.
\newblock Making the {V} in {VQA} matter: Elevating the role of image understanding in visual question answering.
\newblock In \emph{Proceedings of the IEEE Conference on Computer Vision and Pattern Recognition}, pages 6904--6913, 2017.
\newblock URL \url{https://openaccess.thecvf.com/content_cvpr_2017/html/Goyal_Making_the_v_CVPR_2017_paper.html}.

\bibitem[Gu et~al.(2023)Gu, Xiang, Li, Ling, Liu, Mu, Tang, Tao, Wei, Yao, Yuan, Xie, Huang, Chen, and Su]{gu2023maniskill2}
Jiayuan Gu, Fanbo Xiang, Xuanlin Li, Zhan Ling, Xiqiang Liu, Tongzhou Mu, Yihe Tang, Stone Tao, Xinyue Wei, Yunchao Yao, Xiaodi Yuan, Pengwei Xie, Zhiao Huang, Rui Chen, and Hao Su.
\newblock Maniskill2: A unified benchmark for generalizable manipulation skills.
\newblock In \emph{International Conference on Learning Representations}, 2023.

\bibitem[Guo et~al.(2026)Guo, Li, Li, Chen, Sun, Su, Wang, Zhang, Li, and Liu]{x-wam}
Jun Guo, Qiwei Li, Peiyan Li, Zilong Chen, Nan Sun, Yifei Su, Heyun Wang, Yuan Zhang, Xinghang Li, and Huaping Liu.
\newblock Unified 4d world action modeling from video priors with asynchronous denoising, 2026.
\newblock URL \url{https://arxiv.org/abs/2604.26694}.

\bibitem[Han et~al.(2025)Han, Qiu, Liao, Huang, Gao, Yan, and Liu]{han2025robocerebra}
Songhao Han, Boxiang Qiu, Yue Liao, Siyuan Huang, Chen Gao, Shuicheng Yan, and Si~Liu.
\newblock Robocerebra: A large-scale benchmark for long-horizon robotic manipulation evaluation.
\newblock \emph{arXiv preprint arXiv:2506.06677}, 2025.

\bibitem[Hao et~al.(2026)Hao, Zhou, Huang, Hou, Tang, Zhang, Li, Lu, Ren, Meng, Zhang, Wu, Lu, Dang, Guan, Wu, Hou, Li, Xia, Zhou, Zheng, Yue, Gu, Tian, Shen, Cui, Zhang, Xu, Wang, Sun, Zhu, Jiang, Guo, Gong, Zhang, Ding, Ma, Chen, Cai, Xiang, Qu, Luo, Ye, and Chen]{mimo}
Xiaoshuai Hao, Lei Zhou, Zhijian Huang, Zhiwen Hou, Yingbo Tang, Lingfeng Zhang, Guang Li, Zheng Lu, Shuhuai Ren, Xianhui Meng, Yuchen Zhang, Jing Wu, Jinghui Lu, Chenxu Dang, Jiayi Guan, Jianhua Wu, Zhiyi Hou, Hanbing Li, Shumeng Xia, Mingliang Zhou, Yinan Zheng, Zihao Yue, Shuhao Gu, Hao Tian, Yuannan Shen, Jianwei Cui, Wen Zhang, Shaoqing Xu, Bing Wang, Haiyang Sun, Zeyu Zhu, Yuncheng Jiang, Zibin Guo, Chuhong Gong, Chaofan Zhang, Wenbo Ding, Kun Ma, Guang Chen, Rui Cai, Diyun Xiang, Heng Qu, Fuli Luo, Hangjun Ye, and Long Chen.
\newblock Mimo-embodied: X-embodied foundation model technical report, 2026.
\newblock URL \url{https://arxiv.org/abs/2511.16518}.

\bibitem[Hoque et~al.(2026)Hoque, Huang, Yoon, Sivapurapu, and Zhang]{egodex}
Ryan Hoque, Peide Huang, David~J. Yoon, Mouli Sivapurapu, and Jian Zhang.
\newblock {EgoDex: Learning Dexterous Manipulation from Large-Scale Egocentric Video}.
\newblock In \emph{The Fourteenth International Conference on Learning Representations}, 2026.
\newblock \doi{10.48550/arXiv.2505.11709}.
\newblock URL \url{https://openreview.net/forum?id=FFxkFMU89E}.

\bibitem[Hou et~al.(2025)Hou, Wu, Liu, Che, Wu, Liao, Li, He, Feng, Jin, et~al.]{hou2025robomind2}
Chengkai Hou, Kun Wu, Jiaming Liu, Zhengping Che, Di~Wu, Fei Liao, Guangrun Li, Jingyang He, Qiuxuan Feng, Zhao Jin, et~al.
\newblock {RoboMIND 2.0}: A multimodal, bimanual mobile manipulation dataset for generalizable embodied intelligence.
\newblock \emph{arXiv preprint arXiv:2512.24653}, 2025.
\newblock URL \url{https://arxiv.org/abs/2512.24653}.

\bibitem[Intelligence et~al.(2025)Intelligence, Amin, Aniceto, Balakrishna, Black, Conley, Connors, Darpinian, Dhabalia, DiCarlo, Driess, Equi, Esmail, Fang, Finn, Glossop, Godden, Goryachev, Groom, Hancock, Hausman, Hussein, Ichter, Jakubczak, Jen, Jones, Katz, Ke, Kuchi, Lamb, LeBlanc, Levine, Li-Bell, Lu, Mano, Mothukuri, Nair, Pertsch, Ren, Sharma, Shi, Smith, Springenberg, Stachowicz, Stoeckle, Swerdlow, Tanner, Torne, Vuong, Walling, Wang, Williams, Yoo, Yu, Zhilinsky, and Zhou]{pi06}
Physical Intelligence, Ali Amin, Raichelle Aniceto, Ashwin Balakrishna, Kevin Black, Ken Conley, Grace Connors, James Darpinian, Karan Dhabalia, Jared DiCarlo, Danny Driess, Michael Equi, Adnan Esmail, Yunhao Fang, Chelsea Finn, Catherine Glossop, Thomas Godden, Ivan Goryachev, Lachy Groom, Hunter Hancock, Karol Hausman, Gashon Hussein, Brian Ichter, Szymon Jakubczak, Rowan Jen, Tim Jones, Ben Katz, Liyiming Ke, Chandra Kuchi, Marinda Lamb, Devin LeBlanc, Sergey Levine, Adrian Li-Bell, Yao Lu, Vishnu Mano, Mohith Mothukuri, Suraj Nair, Karl Pertsch, Allen~Z. Ren, Charvi Sharma, Lucy~Xiaoyang Shi, Laura Smith, Jost~Tobias Springenberg, Kyle Stachowicz, Will Stoeckle, Alex Swerdlow, James Tanner, Marcel Torne, Quan Vuong, Anna Walling, Haohuan Wang, Blake Williams, Sukwon Yoo, Lili Yu, Ury Zhilinsky, and Zhiyuan Zhou.
\newblock $\pi^{*}_{0.6}$: a vla that learns from experience, 2025.
\newblock URL \url{https://arxiv.org/abs/2511.14759}.

\bibitem[Jang et~al.(2025)Jang, Ye, Lin, Xiang, Bjorck, Fang, Hu, Huang, Kundalia, Lin, Magne, Mandlekar, Narayan, Tan, Wang, Wang, Wang, Xu, Zeng, Zheng, Zheng, Liu, Zettlemoyer, Fox, Kautz, Reed, Zhu, and Fan]{dreamgen}
Joel Jang, Seonghyeon Ye, Zongyu Lin, Jiannan Xiang, Johan Bjorck, Yu~Fang, Fengyuan Hu, Spencer Huang, Kaushil Kundalia, Yen-Chen Lin, Loic Magne, Ajay Mandlekar, Avnish Narayan, You~Liang Tan, Guanzhi Wang, Jing Wang, Qi~Wang, Yinzhen Xu, Xiaohui Zeng, Kaiyuan Zheng, Ruijie Zheng, Ming-Yu Liu, Luke Zettlemoyer, Dieter Fox, Jan Kautz, Scott Reed, Yuke Zhu, and Linxi Fan.
\newblock {DreamGen: Unlocking Generalization in Robot Learning through Video World Models}, 2025.
\newblock URL \url{https://arxiv.org/abs/2505.12705}.
\newblock arXiv preprint arXiv:2505.12705.

\bibitem[Kim et~al.(2024)Kim, Pertsch, Karamcheti, Xiao, Balakrishna, Nair, Rafailov, Foster, Lam, Sanketi, Vuong, Kollar, Burchfiel, Tedrake, Sadigh, Levine, Liang, and Finn]{openvla}
Moo~Jin Kim, Karl Pertsch, Siddharth Karamcheti, Ted Xiao, Ashwin Balakrishna, Suraj Nair, Rafael Rafailov, Ethan Foster, Grace Lam, Pannag Sanketi, Quan Vuong, Thomas Kollar, Benjamin Burchfiel, Russ Tedrake, Dorsa Sadigh, Sergey Levine, Percy Liang, and Chelsea Finn.
\newblock {OpenVLA: An Open-Source Vision-Language-Action Model}, 2024.
\newblock URL \url{https://arxiv.org/abs/2406.09246}.
\newblock arXiv preprint arXiv:2406.09246.

\bibitem[Kim et~al.(2025)Kim, Finn, and Liang]{openvla_oft}
Moo~Jin Kim, Chelsea Finn, and Percy Liang.
\newblock {Fine-Tuning Vision-Language-Action Models: Optimizing Speed and Success}, 2025.
\newblock URL \url{https://arxiv.org/abs/2502.19645}.
\newblock arXiv preprint arXiv:2502.19645.

\bibitem[Kim et~al.(2026)Kim, Gao, Lin, Lin, Ge, Lam, Liang, Song, Liu, Finn, and Gu]{cosmos_policy}
Moo~Jin Kim, Yihuai Gao, Tsung-Yi Lin, Yen-Chen Lin, Yunhao Ge, Grace Lam, Percy Liang, Shuran Song, Ming-Yu Liu, Chelsea Finn, and Jinwei Gu.
\newblock {Cosmos Policy: Fine-Tuning Video Models for Visuomotor Control and Planning}, 2026.
\newblock URL \url{https://arxiv.org/abs/2601.16163}.
\newblock arXiv preprint arXiv:2601.16163.

\bibitem[Kolve et~al.(2017)Kolve, Mottaghi, Han, VanderBilt, Weihs, Herrasti, Deitke, Ehsani, Gordon, Zhu, Kembhavi, Gupta, and Farhadi]{ai2thor}
Eric Kolve, Roozbeh Mottaghi, Winson Han, Eli VanderBilt, Luca Weihs, Alvaro Herrasti, Matt Deitke, Kiana Ehsani, Daniel Gordon, Yuke Zhu, Aniruddha Kembhavi, Abhinav Gupta, and Ali Farhadi.
\newblock {{AI2-THOR}: An Interactive 3D Environment for Visual AI}, 2017.
\newblock URL \url{https://arxiv.org/abs/1712.05474}.

\bibitem[Kuznetsova et~al.(2020)Kuznetsova, Rom, Alldrin, Uijlings, Krasin, Pont-Tuset, Kamali, Popov, Malloci, Kolesnikov, Duerig, and Ferrari]{openimages}
Alina Kuznetsova, Hassan Rom, Neil Alldrin, Jasper Uijlings, Ivan Krasin, Jordi Pont-Tuset, Shahab Kamali, Stefan Popov, Matteo Malloci, Alexander Kolesnikov, Tom Duerig, and Vittorio Ferrari.
\newblock The open images dataset {V4}: Unified image classification, object detection, and visual relationship detection at scale.
\newblock \emph{International Journal of Computer Vision}, 128\penalty0 (7):\penalty0 1956--1981, 2020.
\newblock \doi{10.1007/s11263-020-01316-z}.
\newblock URL \url{https://doi.org/10.1007/s11263-020-01316-z}.

\bibitem[Lee et~al.(2024)Lee, Wang, Etukuru, Kim, Shafiullah, and Pinto]{vqbet}
Seungjae Lee, Yibin Wang, Haritheja Etukuru, H.~Jin Kim, Nur Muhammad~Mahi Shafiullah, and Lerrel Pinto.
\newblock {Behavior Generation with Latent Actions}, 2024.
\newblock URL \url{https://arxiv.org/abs/2403.03181}.
\newblock arXiv preprint arXiv:2403.03181.

\bibitem[Li et~al.(2026)Li, Wang, Ding, Yang, Chen, Tian, Hu, Wang, Lin, Zhao, Liu, and Pang]{robointer_vqa}
Hao Li, Ziqin Wang, Zi-han Ding, Shuai Yang, Yilun Chen, Yang Tian, Xiaolin Hu, Tai Wang, Dahua Lin, Feng Zhao, Si~Liu, and Jiangmiao Pang.
\newblock {RoboInter}: A holistic intermediate representation suite towards robotic manipulation, 2026.
\newblock URL \url{https://arxiv.org/abs/2602.09973}.
\newblock Accepted at ICLR 2026.

\bibitem[Li et~al.(2025)Li, Zuo, Yu, Zhang, Yang, Zhang, Zhu, Zhang, Chen, Cui, Wang, Luo, Fan, Sun, Zeng, Pang, Zhang, Wang, Mu, Zhou, and Ding]{simplevla_rl}
Haozhan Li, Yuxin Zuo, Jiale Yu, Yuhao Zhang, Zhaohui Yang, Kaiyan Zhang, Xuekai Zhu, Yuchen Zhang, Tianxing Chen, Ganqu Cui, Dehui Wang, Dingxiang Luo, Yuchen Fan, Youbang Sun, Jia Zeng, Jiangmiao Pang, Shanghang Zhang, Yu~Wang, Yao Mu, Bowen Zhou, and Ning Ding.
\newblock {SimpleVLA-RL: Scaling VLA Training via Reinforcement Learning}, 2025.
\newblock URL \url{https://arxiv.org/abs/2509.09674}.
\newblock arXiv preprint arXiv:2509.09674.

\bibitem[Lin et~al.(2014)Lin, Maire, Belongie, Hays, Perona, Ramanan, Doll{\'a}r, and Zitnick]{coco}
Tsung-Yi Lin, Michael Maire, Serge Belongie, James Hays, Pietro Perona, Deva Ramanan, Piotr Doll{\'a}r, and C.~Lawrence Zitnick.
\newblock Microsoft {COCO}: Common objects in context.
\newblock In \emph{Computer Vision -- ECCV 2014}, pages 740--755. Springer, 2014.
\newblock \doi{10.1007/978-3-319-10602-1_48}.
\newblock URL \url{https://arxiv.org/abs/1405.0312}.

\bibitem[Lipman et~al.(2022)Lipman, Chen, Ben-Hamu, Nickel, and Le]{flowmatching}
Yaron Lipman, Ricky T.~Q. Chen, Heli Ben-Hamu, Maximilian Nickel, and Matt Le.
\newblock {Flow Matching for Generative Modeling}, 2022.
\newblock URL \url{https://arxiv.org/abs/2210.02747}.
\newblock arXiv preprint arXiv:2210.02747.

\bibitem[Liu et~al.(2024)Liu, Li, Li, and Lee]{llava665k}
Haotian Liu, Chunyuan Li, Yuheng Li, and Yong~Jae Lee.
\newblock Improved baselines with visual instruction tuning.
\newblock In \emph{Proceedings of the IEEE/CVF Conference on Computer Vision and Pattern Recognition}, pages 26296--26306, 2024.
\newblock \doi{10.1109/CVPR52733.2024.02484}.
\newblock URL \url{https://openaccess.thecvf.com/content/CVPR2024/html/Liu_Improved_Baselines_with_Visual_Instruction_Tuning_CVPR_2024_paper.html}.

\bibitem[Liu et~al.(2026)Liu, Bai, Ci, Ma, and Shou]{world_vla_loop}
Xiaokang Liu, Zechen Bai, Hai Ci, Kevin~Yuchen Ma, and Mike~Zheng Shou.
\newblock World-vla-loop: Closed-loop learning of video world model and vla policy, 2026.
\newblock URL \url{https://arxiv.org/abs/2602.06508}.

\bibitem[Llontop and Vadrevu(2026)]{groot_n17}
Edith Llontop and Kalyan Vadrevu.
\newblock {{NVIDIA Isaac GR00T N1.7}: Open Reasoning VLA Model for Humanoid Robots}.
\newblock Hugging Face Blog, April 2026.
\newblock URL \url{https://huggingface.co/blog/nvidia/gr00t-n1-7}.

\bibitem[Lu et~al.(2025)Lu, Guo, Zhang, Zhou, Jiang, Gao, Tang, and Wang]{vla_rl}
Guanxing Lu, Wenkai Guo, Chubin Zhang, Yuheng Zhou, Haonan Jiang, Zifeng Gao, Yansong Tang, and Ziwei Wang.
\newblock {VLA-RL: Towards Masterful and General Robotic Manipulation with Scalable Reinforcement Learning}, 2025.
\newblock URL \url{https://arxiv.org/abs/2505.18719}.
\newblock arXiv preprint arXiv:2505.18719.

\bibitem[Lv et~al.(2026)Lv, Li, Li, Kong, Wang, Yi, Nie, Wang, Zhu, Ni, Deng, Li, Lv, and Huang]{viva2026}
Jindi Lv, Hao Li, Jie Li, Fankun Kong, Yang Wang, Pengfei Yi, Yifei Nie, Xiaofeng Wang, Zheng Zhu, Chaojun Ni, Qiuping Deng, Hengtao Li, Jiancheng Lv, and Guan Huang.
\newblock Viva: A video-generative value model for robot reinforcement learning.
\newblock \url{https://arxiv.org/abs/2604.08168}, 2026.
\newblock Accessed: 2026-08-16.

\bibitem[Matthews et~al.(2025)Matthews, Beukman, Lu, and Foerster]{matthews2024kinetix}
Michael Matthews, Michael Beukman, Chris Lu, and Jakob Foerster.
\newblock Kinetix: Investigating the training of general agents through open-ended physics-based control tasks.
\newblock In \emph{The Thirteenth International Conference on Learning Representations}, 2025.
\newblock URL \url{https://arxiv.org/abs/2410.23208}.

\bibitem[Miao et~al.(2026)Miao, Feng, Wu, Lin, He, Li, and Long]{vla_jepa}
Shangchen Miao, Ningya Feng, Jialong Wu, Ye~Lin, Xu~He, Dong Li, and Mingsheng Long.
\newblock {JEPA-VLA: Video Predictive Embedding is Needed for VLA Models}, 2026.
\newblock URL \url{https://arxiv.org/abs/2602.11832}.
\newblock arXiv preprint arXiv:2602.11832.

\bibitem[{NVIDIA} et~al.(2025){NVIDIA}, :, Bjorck, Casta{\~n}eda, Cherniadev, Da, Ding, Fan, Fang, Fox, Hu, Huang, Jang, Jiang, Kautz, Kundalia, Lao, Li, Lin, Lin, Liu, Llontop, Magne, Mandlekar, Narayan, Nasiriany, Reed, Tan, Wang, Wang, Wang, Wang, Xiang, Xie, Xu, Xu, Ye, Yu, Zhang, Zhang, Zhao, Zheng, and Zhu]{groot_n1}
{NVIDIA}, :, Johan Bjorck, Fernando Casta{\~n}eda, Nikita Cherniadev, Xingye Da, Runyu Ding, Linxi~"Jim" Fan, Yu~Fang, Dieter Fox, Fengyuan Hu, Spencer Huang, Joel Jang, Zhenyu Jiang, Jan Kautz, Kaushil Kundalia, Lawrence Lao, Zhiqi Li, Zongyu Lin, Kevin Lin, Guilin Liu, Edith Llontop, Loic Magne, Ajay Mandlekar, Avnish Narayan, Soroush Nasiriany, Scott Reed, You~Liang Tan, Guanzhi Wang, Zu~Wang, Jing Wang, Qi~Wang, Jiannan Xiang, Yuqi Xie, Yinzhen Xu, Zhenjia Xu, Seonghyeon Ye, Zhiding Yu, Ao~Zhang, Hao Zhang, Yizhou Zhao, Ruijie Zheng, and Yuke Zhu.
\newblock {GR00T N1: An Open Foundation Model for Generalist Humanoid Robots}, 2025.
\newblock URL \url{https://arxiv.org/abs/2503.14734}.
\newblock arXiv preprint arXiv:2503.14734.

\bibitem[{NVIDIA} et~al.(2026){NVIDIA}, :, Aditi, Agarwal, Ali, Allen, Antolini, Aubame, Azzolini, Bai, Bala, Balaji, Bapst, Basant, Beladiya, Bhat, Bhat, Blick, Brighella, Cai, Cai, Cameracci, Cao, Cao, Carlson, Casanova, Chang, Chang, Chao, Chattopadhyay, Chaudhari, Chen, Chen, Chen, Chen, Chen, Chen, Chen, Cheng, Cheng, Chia, Choi, Chung, Cong, Cui, Dadela, Dadhich, Dai, Daw, Degirmenci, Monte, Denomme, Dharur, Lucca, Ding, Ding, Ding, Dong, Drumheller, Du, Dzhumamuratova, Efitorov, Eghbalzadeh, Eigbe, Hanafi, Eslami, Falk, Fan, Fan, Fasale, Fefilatyev, Feng, Ferroni, Fidler, Fu, Fugro, Gaikwad, Galda, Gao, Gao, Ge, Ghosh, Goel, Goel, Gokul, Govindaraju, Gu, Guerrero, Guo, Gupta, Gururani, Hadfield, Han, Handa, Hao, Harrim, Hassani, Hayes-Roth, He, Helvig, Hogg, Huang, Huang, Huang, Huang, Huffman, Hutchins, Indupuru, Ivanovic, Jain, Jang, Ji, Jian, Jiang, Jin, Joshi, Joshi, Joshi, Ju, Jung, Kang, Kassekert, Kautz, Khetan, Kiczka, Kierat, Kim, Kim, Kim, Kong, Kong, Kong, Kornuta, Krivov, Kuang, Kumar, Kuo,
  Kurian, Kutak, Lafleche, Lahkar, Laymoun, Lee, Lee, Leone, Li, Li, Li, Li, Li, Li, Li, Li, Li, Li, Li, Li, Liang, Liao, Lin, Lin, Liu, Liu, Liu, Lu, Lu, Luo, Luo, Luo, Lyu, Ma, Ma, Ma, Majchrowski, Marcoux, Martin, Miao, Mirzaei, Misra, Mo, Mohsin, Moon, Morkisz, Motiian, Motkov, Nah, Narang, Narayanan, Ngazimbi, Ouyang, Pachori, Page, Pang, Park, Patekar, Patwary, Pavone, Pham, Ping, Pouya, Prabhumoye, Praveen, Qu, Rabeti, Ramezanali, Reeb, Ren, Rumley, Rymer, Saito, Seol, Shao, Shekdar, Shen, Shi, Shi, Shi, Shih, Shoeybi, Sieniawski, Song, Sotelo, Sotoodeh, Srinivasa, Srinivasakumar, Stefaniak, Steiger, Sun, Tang, Tang, Tang, Tang, Tavakkoli, Ting, Tomala, Tseng, Varghese, Vasilev, Volk, Wagwani, Waleffe, Wang, Wang, Wang, Wang, Wang, Wang, Wang, Wang, Wang, Watve, Wehr, Wei, Weng, Wu, Wu, Xia, Xiao, Xiao, Xie, Xu, Xu, Xu, Xu, Xu, Xu, Yang, Yang, Yang, Yang, Yang, Yang, You, Yu, Yuan, Yuen, Zeng, Zeren, Zha, Zhang, Zhang, Zhang, Zhang, Zhang, Zhang, Zhang, Zhang, Zhao, Zhao, Zhautouskaya, Zhou, Zhou, Zhu,
  Zhu, Zhylko, and Zolkowski]{cosmos3}
{NVIDIA}, :, Aditi, Niket Agarwal, Arslan Ali, Jon Allen, Martin Antolini, Adeline Aubame, Alisson Azzolini, Junjie Bai, Maciej Bala, Yogesh Balaji, Josh Bapst, Aarti Basant, Mukesh Beladiya, Mohammad~Qazim Bhat, Zaid~Pervaiz Bhat, Dan Blick, Vanni Brighella, Han Cai, Tiffany Cai, Eric Cameracci, Jiaxin Cao, Yulong Cao, Mark Carlson, Carlos Casanova, Ting-Yun Chang, Yan Chang, Yu-Wei Chao, Prithvijit Chattopadhyay, Roshan Chaudhari, Chieh-Yun Chen, Junyu Chen, Ke~Chen, Qizhi Chen, Wenkai Chen, Xiaotong Chen, Yu~Chen, An-Chieh Cheng, Click Cheng, Xiu Chia, Jeana Choi, Chaeyeon Chung, Wenyan Cong, Yin Cui, Magdalena Dadela, Nalin Dadhich, Wenliang Dai, Joyjit Daw, Alperen Degirmenci, Rodrigo Vieira~Del Monte, Robert Denomme, Sameer Dharur, Marco~Di Lucca, Ke~Ding, Wenhao Ding, Yifan Ding, Yuzhu Dong, Nicole Drumheller, Yilun Du, Aigul Dzhumamuratova, Aleksandr Efitorov, Hamid Eghbalzadeh, Naomi Eigbe, Imad~El Hanafi, Hassan Eslami, Benedikt Falk, Jiaojiao Fan, Jim Fan, Amol Fasale, Sergiy Fefilatyev, Liang
  Feng, Francesco Ferroni, Sanja Fidler, Xiao Fu, Vikram Fugro, Prashant Gaikwad, TJ~Galda, Katelyn Gao, Yihuai Gao, Wenhang Ge, Sreyan Ghosh, Arushi Goel, Vivek Goel, Akash Gokul, Rama Govindaraju, Jinwei Gu, Miguel Guerrero, Elfie Guo, Aryaman Gupta, Siddharth Gururani, Hugo Hadfield, Song Han, Ankur Handa, Zekun Hao, Mohammad Harrim, Ali Hassani, Nathan Hayes-Roth, Yufan He, Chris Helvig, Cyrus Hogg, Madison Huang, Michael Huang, Sophia Huang, Yufan Huang, Jacob Huffman, DeLesley Hutchins, Suneel Indupuru, Boris Ivanovic, Arihant Jain, Joel Jang, Ryan Ji, Yanan Jian, Dongfu Jiang, Jingyi Jin, Atharva Joshi, Nikhilesh Joshi, Pranjali Joshi, Andy Ju, Jaehun Jung, Weiwei Kang, Scott Kassekert, Jan Kautz, Ashna Khetan, Julia Kiczka, Slawek Kierat, Gwanghyun Kim, Kuno Kim, Sunny Kim, Kezhi Kong, Xin Kong, Zhifeng Kong, Tomasz Kornuta, Egor Krivov, Hui Kuang, Saurav Kumar, Chia-Wen Kuo, George Kurian, Wojciech Kutak, JF~Lafleche, Himangshu Lahkar, Omar Laymoun, Jayjun Lee, Sanggil Lee, Gabriele Leone, Boyi Li,
  Freya Li, Jiajun Li, Jinfeng Li, Ling Li, Pengcheng Li, Shangru Li, Tingle Li, Xiaolong Li, Xuan Li, Zhaoshuo Li, Zhiqi Li, Hao Liang, Maosheng Liao, Chen-Hsuan Lin, Tsung-Yi Lin, Ming-Yu Liu, Sifei Liu, Zihan Liu, Hai~Loc Lu, Xiangyu Lu, Alice Luo, Ruipu Luo, Wenjie Luo, Jiangran Lyu, Martin~Ding Ma, Nic Ma, Qianli Ma, Dawid Majchrowski, Louis Marcoux, Miguel Martin, Qing Miao, Ashkan Mirzaei, Shreyas Misra, Kaichun Mo, Durra Mohsin, Hyejin Moon, Pawel Morkisz, Saeid Motiian, Kirill Motkov, Seungjun Nah, Yashraj Narang, Deepak Narayanan, Thabang Ngazimbi, Julian Ouyang, Shubham Pachori, David Page, Yatian Pang, Sehwi Park, Mahesh Patekar, Mostofa Patwary, Marco Pavone, Trung Pham, Wei Ping, Soha Pouya, Shrimai Prabhumoye, Varun Praveen, Delin Qu, Hesam Rabeti, Morteza Ramezanali, Marilyn Reeb, Xuanchi Ren, Kristen Rumley, Wojciech Rymer, Jun Saito, Yeongho Seol, John Shao, Piyush Shekdar, Tianwei Shen, Humphrey Shi, Min Shi, Stella Shi, Kevin Shih, Mohammad Shoeybi, Mateusz Sieniawski, Shuran Song,
  Alexander Sotelo, Amir Sotoodeh, Sunil Srinivasa, Vignesh Srinivasakumar, Bartosz Stefaniak, Rahul~Heinrich Steiger, Shangkun Sun, Jiaxiang Tang, Shitao Tang, Yangyang Tang, Yue Tang, Tolou Tavakkoli, Kayley Ting, Krzysztof Tomala, Wei-Cheng Tseng, Jibin Varghese, Sergei Vasilev, Thomas Volk, Raju Wagwani, Roger Waleffe, Andrew~Z. Wang, Boxiang Wang, Haoxiang Wang, Qiao Wang, Shihao Wang, Shijie Wang, Ting-Chun Wang, Yan Wang, Yu~Wang, Rohit Watve, David Wehr, Fangyin Wei, Xinshuo Weng, Jay~Zhangjie Wu, Kedi Wu, Hongchi Xia, Summer Xiao, Tianjun Xiao, Kevin Xie, Daguang Xu, Jiashu Xu, Mengyao Xu, Ruqing Xu, Xingqian Xu, Yao Xu, Dinghao Yang, Dong Yang, Hans Yang, Xiaodong Yang, Xuning Yang, Yichu Yang, Yurong You, Zhiding Yu, Hao Yuan, Simon Yuen, Xiaohui Zeng, Pengcuo Zeren, Cindy Zha, Haotian Zhang, Jenny Zhang, Jing Zhang, Liangkai Zhang, Paris Zhang, Shun Zhang, Xuanmeng Zhang, Zhizheng Zhang, Ann Zhao, Yilin Zhao, Yuliya Zhautouskaya, Charles Zhou, Fengzhe Zhou, Shilin Zhu, Yuke Zhu, Dima Zhylko, and
  Artur Zolkowski.
\newblock {Cosmos 3: Omnimodal World Models for Physical AI}, 2026.
\newblock URL \url{https://arxiv.org/abs/2606.02800}.
\newblock arXiv preprint arXiv:2606.02800.

\bibitem[{Open X-Embodiment Collaboration}(2024)]{oxe}
{Open X-Embodiment Collaboration}.
\newblock {Open X-Embodiment: Robotic Learning Datasets and RT-X Models}.
\newblock In \emph{2024 IEEE International Conference on Robotics and Automation (ICRA)}, pages 6892--6903, May 2024.
\newblock \doi{10.1109/ICRA57147.2024.10611477}.
\newblock URL \url{https://ieeexplore.ieee.org/document/10611477}.

\bibitem[Peng et~al.(2019)Peng, Kumar, Zhang, and Levine]{awr}
Xue~Bin Peng, Aviral Kumar, Grace Zhang, and Sergey Levine.
\newblock {Advantage-Weighted Regression: Simple and Scalable Off-Policy Reinforcement Learning}, 2019.
\newblock URL \url{https://arxiv.org/abs/1910.00177}.
\newblock arXiv preprint arXiv:1910.00177.

\bibitem[Pertsch et~al.(2025)Pertsch, Stachowicz, Ichter, Driess, Nair, Vuong, Mees, Finn, and Levine]{fast}
Karl Pertsch, Kyle Stachowicz, Brian Ichter, Danny Driess, Suraj Nair, Quan Vuong, Oier Mees, Chelsea Finn, and Sergey Levine.
\newblock {FAST: Efficient Action Tokenization for Vision-Language-Action Models}, 2025.
\newblock URL \url{https://arxiv.org/abs/2501.09747}.
\newblock arXiv preprint arXiv:2501.09747.

\bibitem[{Physical Intelligence} et~al.(2025{\natexlab{a}}){Physical Intelligence}, Amin, Aniceto, Balakrishna, Black, Conley, Connors, Darpinian, Dhabalia, DiCarlo, Driess, Equi, Esmail, Fang, Finn, Glossop, Godden, Goryachev, Groom, Hancock, Hausman, Hussein, Ichter, Jakubczak, Jen, Jones, Katz, Ke, Kuchi, Lamb, LeBlanc, Levine, Li-Bell, Lu, Mano, Mothukuri, Nair, Pertsch, Ren, Sharma, Shi, Smith, Springenberg, Stachowicz, Stoeckle, Swerdlow, Tanner, Torne, Vuong, Walling, Wang, Williams, Yoo, Yu, Zhilinsky, and Zhou]{recap}
{Physical Intelligence}, Ali Amin, Raichelle Aniceto, Ashwin Balakrishna, Kevin Black, Ken Conley, Grace Connors, James Darpinian, Karan Dhabalia, Jared DiCarlo, Danny Driess, Michael Equi, Adnan Esmail, Yunhao Fang, Chelsea Finn, Catherine Glossop, Thomas Godden, Ivan Goryachev, Lachy Groom, Hunter Hancock, Karol Hausman, Gashon Hussein, Brian Ichter, Szymon Jakubczak, Rowan Jen, Tim Jones, Ben Katz, Liyiming Ke, Chandra Kuchi, Marinda Lamb, Devin LeBlanc, Sergey Levine, Adrian Li-Bell, Yao Lu, Vishnu Mano, Mohith Mothukuri, Suraj Nair, Karl Pertsch, Allen~Z. Ren, Charvi Sharma, Lucy~Xiaoyang Shi, Laura Smith, Jost~Tobias Springenberg, Kyle Stachowicz, Will Stoeckle, Alex Swerdlow, James Tanner, Marcel Torne, Quan Vuong, Anna Walling, Haohuan Wang, Blake Williams, Sukwon Yoo, Lili Yu, Ury Zhilinsky, and Zhiyuan Zhou.
\newblock {$\pi^{*}_{0.6}$: a VLA That Learns From Experience}, 2025{\natexlab{a}}.
\newblock URL \url{https://arxiv.org/abs/2511.14759}.
\newblock arXiv preprint arXiv:2511.14759.

\bibitem[{Physical Intelligence} et~al.(2025{\natexlab{b}}){Physical Intelligence}, Black, Brown, Darpinian, Dhabalia, Driess, Esmail, Equi, Finn, Fusai, Galliker, Ghosh, Groom, Hausman, Ichter, Jakubczak, Jones, Ke, LeBlanc, Levine, Li-Bell, Mothukuri, Nair, Pertsch, Ren, Shi, Smith, Springenberg, Stachowicz, Tanner, Vuong, Walke, Walling, Wang, Yu, and Zhilinsky]{pi05}
{Physical Intelligence}, Kevin Black, Noah Brown, James Darpinian, Karan Dhabalia, Danny Driess, Adnan Esmail, Michael Equi, Chelsea Finn, Niccolo Fusai, Manuel~Y. Galliker, Dibya Ghosh, Lachy Groom, Karol Hausman, Brian Ichter, Szymon Jakubczak, Tim Jones, Liyiming Ke, Devin LeBlanc, Sergey Levine, Adrian Li-Bell, Mohith Mothukuri, Suraj Nair, Karl Pertsch, Allen~Z. Ren, Lucy~Xiaoyang Shi, Laura Smith, Jost~Tobias Springenberg, Kyle Stachowicz, James Tanner, Quan Vuong, Homer Walke, Anna Walling, Haohuan Wang, Lili Yu, and Ury Zhilinsky.
\newblock {$\pi_{0.5}$: a Vision-Language-Action Model with Open-World Generalization}, 2025{\natexlab{b}}.
\newblock URL \url{https://arxiv.org/abs/2504.16054}.
\newblock arXiv preprint arXiv:2504.16054.

\bibitem[{Physical Intelligence} et~al.(2026){Physical Intelligence}, Ai, Amin, Aniceto, Balakrishna, Balke, Black, Bokinsky, Cao, Charbonnier, Choudhary, Collins, Conley, Connors, Darpinian, Dhabalia, Dhaka, DiCarlo, Driess, Equi, Esmail, Fang, Finn, Glossop, Godden, Goryachev, Groom, Habeeb, Hancock, Hausman, Hussein, Hwang, Ichter, Jacobsen, Jakubczak, Jen, Jones, Kammerer, Katz, Ke, Khadikov, Kuchi, Lamb, LeBlanc, LeCount, Levine, Li, Li-Bell, Lialin, Liang, Lim, Lu, Luo, Mano, Marwaha, Mongush, Murphy, Nair, Patterson, Pertsch, Ren, Schelske, Sharma, Shi, Shi, Smith, Springenberg, Stachowicz, Stoeckle, Tang, Tanner, Tekeste, Torne, Vedder, Vuong, Walling, Wang, Wang, Wang, Whalen, Whitmore, Williams, Xu, Yoo, Yu, Zhang, Zhang, and Zhilinsky]{pi07}
{Physical Intelligence}, Bo~Ai, Ali Amin, Raichelle Aniceto, Ashwin Balakrishna, Greg Balke, Kevin Black, George Bokinsky, Shihao Cao, Thomas Charbonnier, Vedant Choudhary, Foster Collins, Ken Conley, Grace Connors, James Darpinian, Karan Dhabalia, Maitrayee Dhaka, Jared DiCarlo, Danny Driess, Michael Equi, Adnan Esmail, Yunhao Fang, Chelsea Finn, Catherine Glossop, Thomas Godden, Ivan Goryachev, Lachlan Groom, Haroun Habeeb, Hunter Hancock, Karol Hausman, Gashon Hussein, Victor Hwang, Brian Ichter, Connor Jacobsen, Szymon Jakubczak, Rowan Jen, Tim Jones, Gregg Kammerer, Ben Katz, Liyiming Ke, Mairbek Khadikov, Chandra Kuchi, Marinda Lamb, Devin LeBlanc, Brendon LeCount, Sergey Levine, Xinyu Li, Adrian Li-Bell, Vladislav Lialin, Zhonglin Liang, Wallace Lim, Yao Lu, Enyu Luo, Vishnu Mano, Nandan Marwaha, Aikys Mongush, Liam Murphy, Suraj Nair, Tyler Patterson, Karl Pertsch, Allen~Z. Ren, Gavin Schelske, Charvi Sharma, Baifeng Shi, Lucy~Xiaoyang Shi, Laura Smith, Jost~Tobias Springenberg, Kyle Stachowicz, Will
  Stoeckle, Jiaming Tang, Jimmy Tanner, Shalom Tekeste, Marcel Torne, Kyle Vedder, Quan Vuong, Anna Walling, Haohuan Wang, Jason Wang, XuDong Wang, Chris Whalen, Samuel Whitmore, Blake Williams, Charles Xu, Sukwon Yoo, Lili Yu, Wuming Zhang, Zhuoyang Zhang, and Ury Zhilinsky.
\newblock {$\pi_{0.7}$: a Steerable Generalist Robotic Foundation Model with Emergent Capabilities}, 2026.
\newblock URL \url{https://arxiv.org/abs/2604.15483}.
\newblock arXiv preprint arXiv:2604.15483.

\bibitem[Punamiya et~al.(2026)Punamiya, Kareer, Liu, Citron, Qiu, Cai, Gavryushin, Chen, Liconti, Zhu, Aphiwetsa, Li, Cheluva, Kuppili, Liu, Patel, Gao, Chung, Co, Zbizika, Liu, Xu, Xiong, Chen, Oliani, Xuan, Yang, Wang, Fort, Newcombe, Gao, Chong, Matsuda, Doriwala, Pollefeys, Katzschmann, Wang, Song, Hoffman, and Xu]{egoverse}
Ryan Punamiya, Simar Kareer, Zeyi Liu, Josh Citron, Ri-Zhao Qiu, Xiongyi Cai, Alexey Gavryushin, Jiaqi Chen, Davide Liconti, Lawrence~Y. Zhu, Patcharapong Aphiwetsa, Baoyu Li, Aniketh Cheluva, Pranav Kuppili, Yangcen Liu, Dhruv Patel, Aidan Gao, Hye-Young Chung, Ryan Co, Renee Zbizika, Jeff Liu, Xiaomeng Xu, Haoyu Xiong, Geng Chen, Sebastiano Oliani, Wenkai Xuan, Chenyu Yang, Xi~Wang, James Fort, Richard Newcombe, Josh Gao, Jason Chong, Garrett Matsuda, Aseem Doriwala, Marc Pollefeys, Robert Katzschmann, Xiaolong Wang, Shuran Song, Judy Hoffman, and Danfei Xu.
\newblock {EgoVerse: An Egocentric Human Dataset for Robot Learning from Around the World}, 2026.
\newblock URL \url{https://arxiv.org/abs/2604.07607}.
\newblock arXiv preprint arXiv:2604.07607.

\bibitem[{Qwen Team}(2026{\natexlab{a}})]{qwen3.5}
{Qwen Team}.
\newblock {Qwen3.5}: Towards native multimodal agents, February 2026{\natexlab{a}}.
\newblock URL \url{https://qwen.ai/blog?id=qwen3.5}.

\bibitem[{Qwen Team}(2026{\natexlab{b}})]{qwen3.6-27b}
{Qwen Team}.
\newblock {Qwen3.6-27B}: Flagship-level coding in a {27B} dense model, April 2026{\natexlab{b}}.
\newblock URL \url{https://qwen.ai/blog?id=qwen3.6-27b}.

\bibitem[Schuhmann et~al.(2021)Schuhmann, Vencu, Beaumont, Kaczmarczyk, Mullis, Katta, Coombes, Jitsev, and Komatsuzaki]{laion400m}
Christoph Schuhmann, Richard Vencu, Romain Beaumont, Robert Kaczmarczyk, Clayton Mullis, Aarush Katta, Theo Coombes, Jenia Jitsev, and Aran Komatsuzaki.
\newblock {LAION-400M}: Open dataset of {CLIP}-filtered 400 million image--text pairs.
\newblock In \emph{NeurIPS Workshop on Data-Centric AI}, 2021.
\newblock URL \url{https://arxiv.org/abs/2111.02114}.

\bibitem[Shen et~al.(2026)Shen, Gu, Qin, Wu, Wu, Tan, Sun, Wang, Wu, An, Cai, Feng, and Yang]{danqing}
Hengyu Shen, Tiancheng Gu, Bin Qin, Lan Wu, Yuling Wu, Shuo Tan, Zelong Sun, Jun Wang, Nan Wu, Xiang An, Weidong Cai, Ziyong Feng, and Kaicheng Yang.
\newblock {DanQing}: An up-to-date large-scale chinese vision--language pre-training dataset, 2026.
\newblock URL \url{https://arxiv.org/abs/2601.10305}.

\bibitem[Shu et~al.(2026)Shu, Lin, Wei, and Wang]{feat2go}
Junyang Shu, Zhiwei Lin, Bingqing Wei, and Yongtao Wang.
\newblock Feat2go: Visual feature-grounded value estimation for embodied reinforcement learning, 2026.
\newblock URL \url{https://arxiv.org/abs/2605.30795}.

\bibitem[{Spirit AI Team}(2026)]{spiritv15}
{Spirit AI Team}.
\newblock {Spirit-v1.5: Clean Data Is the Enemy of Great Robot Foundation Models}.
\newblock Spirit AI Blog, 2026.
\newblock URL \url{https://www.spirit-ai.com/en/blog/spirit-v1-5}.

\bibitem[Steiner et~al.(2024)Steiner, Pinto, Tschannen, Keysers, Wang, Bitton, Gritsenko, Minderer, Sherbondy, Long, Qin, Ingle, Bugliarello, Kazemzadeh, Mesnard, Alabdulmohsin, Beyer, and Zhai]{paligemma2}
Andreas Steiner, Andr{\'e}~Susano Pinto, Michael Tschannen, Daniel Keysers, Xiao Wang, Yonatan Bitton, Alexey Gritsenko, Matthias Minderer, Anthony Sherbondy, Shangbang Long, Siyang Qin, Reeve Ingle, Emanuele Bugliarello, Sahar Kazemzadeh, Thomas Mesnard, Ibrahim Alabdulmohsin, Lucas Beyer, and Xiaohua Zhai.
\newblock {PaliGemma 2: A Family of Versatile VLMs for Transfer}, 2024.
\newblock URL \url{https://arxiv.org/abs/2412.03555}.
\newblock arXiv preprint arXiv:2412.03555.

\bibitem[Suhr et~al.(2019)Suhr, Zhou, Zhang, Zhang, Bai, and Artzi]{nlvr2}
Alane Suhr, Stephanie Zhou, Ally Zhang, Iris Zhang, Huajun Bai, and Yoav Artzi.
\newblock A corpus for reasoning about natural language grounded in photographs.
\newblock In \emph{Proceedings of the 57th Annual Meeting of the Association for Computational Linguistics}, pages 6418--6428. Association for Computational Linguistics, 2019.
\newblock \doi{10.18653/v1/P19-1644}.
\newblock URL \url{https://aclanthology.org/P19-1644/}.

\bibitem[Tang et~al.(2025)Tang, Zhang, Zhang, Zhao, and Hao]{roboafford}
Yingbo Tang, Lingfeng Zhang, Shuyi Zhang, Yinuo Zhao, and Xiaoshuai Hao.
\newblock {RoboAfford}: A dataset and benchmark for enhancing object and spatial affordance learning in robot manipulation.
\newblock In \emph{Proceedings of the 33rd ACM International Conference on Multimedia}, pages 12706--12713. Association for Computing Machinery, 2025.
\newblock \doi{10.1145/3746027.3758209}.
\newblock URL \url{https://doi.org/10.1145/3746027.3758209}.

\bibitem[Team et~al.(2026{\natexlab{a}})Team, Abd, Aggarwal, Algayres, Andreev, Bachem, Ballantyne, Brick, C{\u{a}}rbune, Casbon, et~al.]{gemma4}
Gemma Team, Sherif~El Abd, Vaibhav Aggarwal, Robin Algayres, Alek Andreev, Olivier Bachem, Ian Ballantyne, Cormac Brick, Victor C{\u{a}}rbune, Michelle Casbon, et~al.
\newblock Gemma 4 technical report.
\newblock \emph{arXiv preprint arXiv:2607.02770}, 2026{\natexlab{a}}.

\bibitem[Team et~al.(2026{\natexlab{b}})Team, Ma, Wang, Li, Ni, Li, Huang, Zhao, Li, Li, Liu, Lu, Deng, Yu, Xu, Zhou, Xu, Chen, Wang, Tian, Wang, Chang, Zhou, Ye, Wu, Wu, and Zhu]{gigaworld1}
GigaWorld Team, Angyuan Ma, Boyuan Wang, Bohan Li, Chaojun Ni, Guo Li, Guan Huang, Guosheng Zhao, Hao Li, Hengtao Li, Jingyu Liu, Jiwen Lu, Qiuping Deng, Tingdong Yu, Xuancheng Xu, Xinyu Zhou, Xiuwei Xu, Xinze Chen, Xiaofeng Wang, Xiaoyu Tian, Yang Wang, Yifan Chang, Yukun Zhou, Yun Ye, Zhenyu Wu, Zhanqian Wu, and Zheng Zhu.
\newblock Gigaworld-1: A roadmap to build world models for robot policy evaluation, 2026{\natexlab{b}}.
\newblock URL \url{https://arxiv.org/abs/2607.02642}.

\bibitem[Tong et~al.(2024)Tong, Brown, Wu, Woo, Middepogu, Akula, Yang, Yang, Iyer, Pan, Wang, Fergus, LeCun, and Xie]{cambrian}
Shengbang Tong, Ellis Brown, Penghao Wu, Sanghyun Woo, Manoj Middepogu, Sai~Charitha Akula, Jihan Yang, Shusheng Yang, Adithya Iyer, Xichen Pan, Austin Wang, Rob Fergus, Yann LeCun, and Saining Xie.
\newblock {Cambrian-1}: A fully open, vision-centric exploration of multimodal {LLMs}.
\newblock In \emph{Advances in Neural Information Processing Systems}, volume~37, pages 87310--87356, 2024.
\newblock \doi{10.52202/079017-2771}.
\newblock URL \url{https://proceedings.neurips.cc/paper_files/paper/2024/hash/9ee3a664ccfeabc0da16ac6f1f1cfe59-Abstract-Conference.html}.

\bibitem[Wu et~al.(2025)Wu, Liu, Xie, Wang, Li, Yang, Li, Zhu, Wu, Liu, et~al.]{wu2025robocoin}
Shihan Wu, Xuecheng Liu, Shaoxuan Xie, Pengwei Wang, Xinghang Li, Bowen Yang, Zhe Li, Kai Zhu, Hongyu Wu, Yiheng Liu, et~al.
\newblock {RoboCOIN}: An open-sourced bimanual robotic data collection for integrated manipulation.
\newblock \emph{arXiv preprint arXiv:2511.17441}, 2025.
\newblock URL \url{https://arxiv.org/abs/2511.17441}.

\bibitem[Wu et~al.(2026)Wu, Wang, Lu, Sun, Liu, Wang, Yan, Wang, Ma, Wang, Liu, Yang, Zhou, Zhang, Zhou, Su, Xue, Tan, Zhang, Zhang, Liao, Zhu, Shen, and Zheng]{lingbot_vla2}
Wei Wu, Fangjing Wang, Fan Lu, He~Sun, Shi Liu, Yunnan Wang, Yibin Yan, Yong Wang, Shuailei Ma, Xinyang Wang, Yibin Liu, Shuai Yang, Tianxiang Zhou, Kejia Zhang, Lei Zhou, Cheng Su, Nan Xue, Bin Tan, Han Zhang, Youchao Zhang, Fei Liao, Xing Zhu, Yujun Shen, and Kecheng Zheng.
\newblock {From Foundation to Application: Improving VLA Models in Practice}, 2026.
\newblock URL \url{https://arxiv.org/abs/2607.06403}.
\newblock arXiv preprint arXiv:2607.06403.

\bibitem[{Xiaomi Robotics Team} et~al.(2026){Xiaomi Robotics Team}, Guo, Jin, Li, Li, Li, Liu, Peng, Qin, Su, Sun, Sun, Suo, Wang, Wang, Wu, Xia, Zhang, Zhao, Chen, Chen, He, Li, Li, Li, Qu, Song, Xiang, Xie, Xu, Ye, Ye, Zhao, and Zhou]{xiaomi_robotics_1}
{Xiaomi Robotics Team}, Jun Guo, Piaopiao Jin, Jason Li, Peiyan Li, Yingyan Li, Futeng Liu, Wanli Peng, Optimus Qin, Yifei Su, Nan Sun, Qiao Sun, Runze Suo, Heyun Wang, Yunhong Wang, Rujie Wu, Caoyu Xia, Lina Zhang, Jack Zhao, Guoliang Chen, Wenlong Chen, Xinze He, Bin Li, Qing Li, Zhuorong Li, Heng Qu, Wenxuan Song, Diyun Xiang, Yifan Xie, Peiran Xu, Hangjun Ye, Wen Ye, Han Zhao, and Quanyun Zhou.
\newblock {Xiaomi-Robotics-1: Scaling Vision-Language-Action Models with over 100K Hours of Real-World Trajectories}, 2026.
\newblock URL \url{https://arxiv.org/abs/2607.15330}.
\newblock arXiv preprint arXiv:2607.15330.

\bibitem[Xu et~al.(2026)Xu, Springenberg, Equi, Amin, Esmail, Levine, and Ke]{rltoken}
Charles Xu, Jost~Tobias Springenberg, Michael Equi, Ali Amin, Adnan Esmail, Sergey Levine, and Liyiming Ke.
\newblock {RL Token: Bootstrapping Online RL with Vision-Language-Action Models}, 2026.
\newblock URL \url{https://arxiv.org/abs/2604.23073}.
\newblock arXiv preprint arXiv:2604.23073.

\bibitem[Yang et~al.(2026)Yang, Guo, Lu, Zhaxizhuoma, Zhang, Wang, Xiao, Yan, Chen, Ding, Yu, Bai, and Li]{umi_vqa}
Siyuan Yang, Linzheng Guo, Ouyang Lu, Zhaxizhuoma, Daoran Zhang, Xinmiao Wang, Ting Xiao, Fangzheng Yan, Zhijun Chen, Yan Ding, Chao Yu, Chenjia Bai, and Xuelong Li.
\newblock {VISTA}: Vision-grounded and physics-validated adaptation of {UMI} data for {VLA} training, 2026.
\newblock URL \url{https://arxiv.org/abs/2606.04708}.

\bibitem[Ye et~al.(2026{\natexlab{a}})Ye, Wang, Ni, Huang, Zhao, Li, Li, Li, Lv, Liu, Cao, Li, Deng, Mei, Wang, Chen, Zhou, Wang, Chang, Li, Zhou, Ye, Liu, and Zhu]{gigaworld_policy}
Angen Ye, Boyuan Wang, Chaojun Ni, Guan Huang, Guosheng Zhao, Hao Li, Hengtao Li, Jie Li, Jindi Lv, Jingyu Liu, Min Cao, Peng Li, Qiuping Deng, Wenjun Mei, Xiaofeng Wang, Xinze Chen, Xinyu Zhou, Yang Wang, Yifan Chang, Yifan Li, Yukun Zhou, Yun Ye, Zhichao Liu, and Zheng Zhu.
\newblock {GigaWorld-Policy: An Efficient Action-Centered World-Action Model}, 2026{\natexlab{a}}.
\newblock URL \url{https://arxiv.org/abs/2603.17240}.
\newblock arXiv preprint arXiv:2603.17240.

\bibitem[Ye et~al.(2026{\natexlab{b}})Ye, Ge, Zheng, Gao, Yu, Kurian, Indupuru, Tan, Zhu, Xiang, Malik, Lee, Liang, Ranawaka, Gu, Xu, Wang, Hu, Narayan, Bjorck, Wang, Kim, Niu, Zheng, Xie, Wu, Wang, Julian, Xu, Du, Chebotar, Reed, Kautz, Zhu, Fan, and Jang]{dreamzero}
Seonghyeon Ye, Yunhao Ge, Kaiyuan Zheng, Shenyuan Gao, Sihyun Yu, George Kurian, Suneel Indupuru, You~Liang Tan, Chuning Zhu, Jiannan Xiang, Ayaan Malik, Kyungmin Lee, William Liang, Nadun Ranawaka, Jiasheng Gu, Yinzhen Xu, Guanzhi Wang, Fengyuan Hu, Avnish Narayan, Johan Bjorck, Jing Wang, Gwanghyun Kim, Dantong Niu, Ruijie Zheng, Yuqi Xie, Jimmy Wu, Qi~Wang, Ryan Julian, Danfei Xu, Yilun Du, Yevgen Chebotar, Scott Reed, Jan Kautz, Yuke Zhu, Linxi~"Jim" Fan, and Joel Jang.
\newblock {World Action Models are Zero-shot Policies}, 2026{\natexlab{b}}.
\newblock URL \url{https://arxiv.org/abs/2602.15922}.
\newblock arXiv preprint arXiv:2602.15922.

\bibitem[Young et~al.(2014)Young, Lai, Hodosh, and Hockenmaier]{flickr30k}
Peter Young, Alice Lai, Micah Hodosh, and Julia Hockenmaier.
\newblock From image descriptions to visual denotations: New similarity metrics for semantic inference over event descriptions.
\newblock \emph{Transactions of the Association for Computational Linguistics}, 2:\penalty0 67--78, 2014.
\newblock \doi{10.1162/tacl_a_00166}.
\newblock URL \url{https://aclanthology.org/Q14-1006/}.

\bibitem[Yu et~al.(2024)Yu, Sun, Zhang, Cui, Zhang, Cao, Wang, and Liu]{capsfusion}
Qiying Yu, Quan Sun, Xiaosong Zhang, Yufeng Cui, Fan Zhang, Yue Cao, Xinlong Wang, and Jingjing Liu.
\newblock {CapsFusion}: Rethinking image--text data at scale.
\newblock In \emph{Proceedings of the IEEE/CVF Conference on Computer Vision and Pattern Recognition}, pages 14022--14032, 2024.
\newblock URL \url{https://openaccess.thecvf.com/content/CVPR2024/html/Yu_CapsFusion_Rethinking_Image-Text_Data_at_Scale_CVPR_2024_paper.html}.

\bibitem[Yu et~al.(2026)Yu, Zhang, Liu, Liu, Kang, Li, Shi, Ma, Yang, Pan, Chen, Liu, Sun, Guo, Zhang, Zhou, Xu, Chen, Huang, Wang, Kuzi, Zhai, Su, Gan, Liang, Wang, and Wang]{walloss05}
Ryan Yu, Pushi Zhang, Starrick Liu, Brae Liu, Miracle Kang, Shalfun Li, Lights Shi, Ellie Ma, Ping Yang, Chris Pan, Jerry Chen, Dongxiu Liu, Rain Sun, Miles Guo, Byron Zhang, Hugo Zhou, Zach Xu, Vincent Chen, Harrison Huang, James Wang, Dance Kuzi, Andy Zhai, Hang Su, Roy Gan, Lucy Liang, Hao Wang, and Qian Wang.
\newblock {Wall-OSS-0.5 Technical Report}, 2026.
\newblock URL \url{https://arxiv.org/abs/2605.30877}.
\newblock arXiv preprint arXiv:2605.30877.

\bibitem[Yuan et~al.(2026)Yuan, Liang, Chen, Wang, Li, Lin, Huang, Lei, Zhang, Zhang, Zhang, Fan, Zhou, Peng, Lv, Chen, Yang, Huang, Lin, Liu, Zhou, Wu, and Chen]{qwen_robotmanip}
Haoqi Yuan, Zhixuan Liang, Anzhe Chen, Ye~Wang, Haoyang Li, Pei Lin, Yiyang Huang, Zixing Lei, Tong Zhang, Jiazhao Zhang, Jie Zhang, Jingyang Fan, Gengze Zhou, Qihang Peng, Chenxu Lv, Xiaoyue Chen, An~Yang, Fei Huang, Junyang Lin, Dayiheng Liu, Jingren Zhou, Chenfei Wu, and Xiong-Hui Chen.
\newblock {Qwen-RobotManip Technical Report: Alignment Unlocks Scale for Robotic Manipulation Foundation Models}, 2026.
\newblock URL \url{https://arxiv.org/abs/2606.17846}.
\newblock arXiv preprint arXiv:2606.17846.

\bibitem[Yuan et~al.(2025)Yuan, Duan, Blukis, Pumacay, Krishna, Murali, Mousavian, and Fox]{robopoint}
Wentao Yuan, Jiafei Duan, Valts Blukis, Wilbert Pumacay, Ranjay Krishna, Adithyavairavan Murali, Arsalan Mousavian, and Dieter Fox.
\newblock {RoboPoint}: A vision--language model for spatial affordance prediction in robotics.
\newblock In \emph{Proceedings of the 8th Conference on Robot Learning}, volume 270 of \emph{Proceedings of Machine Learning Research}, pages 4005--4020. PMLR, 2025.
\newblock URL \url{https://proceedings.mlr.press/v270/yuan25c.html}.

\bibitem[Zhang et~al.(2026)Zhang, Xiang, Lin, Huang, Wang, Zhong, Dong, Wu, Rao, Zhang, He, Chen, Huang, Chen, Su, Yu, Wang, Zhu, Teng, Guo, Zhang, Liu, Wang, Lu, Hu, and Zhang]{hyvla05}
He~Zhang, Lingzhu Xiang, Haitao Lin, Zeyu Huang, Minghui Wang, Dingyan Zhong, Yubo Dong, Yihao Wu, Yongming Rao, Dongsheng Zhang, Wanjia He, Ling Chen, Kai Huang, Jiahao Chen, Sichang Su, Xumin Yu, Ziyi Wang, Chengwei Zhu, Xiao Teng, Yuchun Guo, Yufeng Zhang, Yuandong Liu, Rui Wang, Zisheng Lu, Han Hu, and Zhengyou Zhang.
\newblock {Hy-Embodied-0.5-VLA: From Vision-Language-Action Models to a Real-World Robot Learning Stack}, 2026.
\newblock URL \url{https://arxiv.org/abs/2606.14409}.
\newblock arXiv preprint arXiv:2606.14409.

\bibitem[Zheng et~al.(2025)Zheng, Li, Wang, Liu, Kang, Feng, Zheng, Zou, Chen, Zeng, Zhang, Pang, Liu, Wang, and Zhan]{x-vla}
Jinliang Zheng, Jianxiong Li, Zhihao Wang, Dongxiu Liu, Xirui Kang, Yuchun Feng, Yinan Zheng, Jiayin Zou, Yilun Chen, Jia Zeng, Ya-Qin Zhang, Jiangmiao Pang, Jingjing Liu, Tai Wang, and Xianyuan Zhan.
\newblock X-vla: Soft-prompted transformer as scalable cross-embodiment vision-language-action model, 2025.
\newblock URL \url{https://arxiv.org/abs/2510.10274}.

\bibitem[Zheng et~al.(2026)Zheng, Peng, Li, Zheng, Li, Jin, Wei, Zhang, Zheng, Cao, Gu, Zou, Li, Wu, Yang, Liu, Li, Si, Zhu, Fu, Wang, Yao, Zhao, Chen, and Ding]{wiyh}
Yupeng Zheng, Jichao Peng, Weize Li, Yuhang Zheng, Xiang Li, Yujie Jin, Julong Wei, Guanhua Zhang, Ruiling Zheng, Ming Cao, Songen Gu, Zhenhong Zou, Kaige Li, Ke~Wu, Mingmin Yang, Jiahao Liu, Pengfei Li, Hengjie Si, Feiyu Zhu, Wang Fu, Likun Wang, Ruiwen Yao, Jieru Zhao, Yilun Chen, and Wenchao Ding.
\newblock World in your hands: A large-scale and open-source ecosystem for learning human-centric manipulation in the wild, 2026.
\newblock URL \url{https://arxiv.org/abs/2512.24310}.

\bibitem[Zhong et~al.(2026)Zhong, Liu, Wei, Xiong, Yao, Liu, and Ren]{acot-vla}
Linqing Zhong, Yi~Liu, Yifei Wei, Ziyu Xiong, Maoqing Yao, Si~Liu, and Guanghui Ren.
\newblock Acot-vla: Action chain-of-thought for vision-language-action models, 2026.
\newblock URL \url{https://arxiv.org/abs/2601.11404}.

\bibitem[Zhou et~al.(2025)Zhou, An, Chi, Han, Rong, Zhang, Wang, Wang, Huang, Sheng, and Zhang]{refspatial}
Enshen Zhou, Jingkun An, Cheng Chi, Yi~Han, Shanyu Rong, Chi Zhang, Pengwei Wang, Zhongyuan Wang, Tiejun Huang, Lu~Sheng, and Shanghang Zhang.
\newblock {RoboRefer}: Towards spatial referring with reasoning in vision--language models for robotics.
\newblock In \emph{Advances in Neural Information Processing Systems}, volume~38, 2025.
\newblock URL \url{https://proceedings.neurips.cc/paper_files/paper/2025/hash/29416b66c2149872b9d1415a3fd2c5e0-Abstract-Conference.html}.

\bibitem[Zhou et~al.(2019)Zhou, Barnes, Lu, Yang, and Li]{6drot}
Yi~Zhou, Connelly Barnes, Jingwan Lu, Jimei Yang, and Hao Li.
\newblock On the continuity of rotation representations in neural networks.
\newblock In \emph{2019 IEEE/CVF Conference on Computer Vision and Pattern Recognition (CVPR)}, pages 5738--5746, 2019.
\newblock \doi{10.1109/CVPR.2019.00589}.

\bibitem[Zhu et~al.(2025)Zhu, Yu, Feng, Burchfiel, Shah, and Gupta]{uwm}
Chuning Zhu, Raymond Yu, Siyuan Feng, Benjamin Burchfiel, Paarth Shah, and Abhishek Gupta.
\newblock {Unified World Models: Coupling Video and Action Diffusion for Pretraining on Large Robotic Datasets}.
\newblock In \emph{Proceedings of Robotics: Science and Systems}, Los Angeles, CA, USA, June 2025.
\newblock \doi{10.15607/RSS.2025.XXI.015}.
\newblock URL \url{https://www.roboticsproceedings.org/rss21/p015.html}.

\end{thebibliography}
